\documentclass{article}

\usepackage[main, final]{neurips_2026}
\usepackage{threeparttable}

\usepackage{array}
\usepackage{booktabs}
\usepackage{tabularx}
\usepackage{array}
\usepackage{hyperref}
\usepackage[table]{xcolor} %

\newcolumntype{L}[1]{>{\raggedright\arraybackslash}p{#1}}
\newcolumntype{C}[1]{>{\centering\arraybackslash}p{#1}}
\newcolumntype{R}[1]{>{\raggedleft\arraybackslash}p{#1}}
\usepackage[utf8]{inputenc} %
\usepackage[T1]{fontenc}    %
\usepackage{hyperref}       %
\usepackage{url}            %
\usepackage{booktabs}       %
\usepackage{amsmath}        %
\usepackage{amsfonts}       %
\usepackage{nicefrac}       %
\usepackage{microtype}      %
\usepackage{xcolor}         %
\usepackage{graphicx}       %
\usepackage{wrapfig}        %
\usepackage{subcaption}     %
\usepackage{multirow}
\usepackage{comment}        %
\usepackage{fvextra}       
\usepackage[utf8]{inputenc}
\usepackage{booktabs}
\usepackage{tabularx}
\usepackage{enumitem}
\makeatletter
\let\hyper@linkstart\@gobbletwo
\let\hyper@linkend\@empty
\makeatother

\title{DeepArrhythmia: Segment-Contextualized ECG Arrhythmia Classification via Selective Evidence Acquisition}

\author{
  Jiahui Li, Ruili Fang, Zishuai Liu, WenZhan Song, Jin Lu, Fei Dou \\
  University of Georgia \\
  \texttt{\{jl57095,ruili.fang,zishuai.liu,wsong,Jin.Lu,Fei.Dou\}@uga.edu}
}

\begin{document}

\maketitle

\begin{abstract}

Beat-level Electrocardiography (ECG) arrhythmia detection aims to assign an arrhythmia class to each beat in a recording, yet many existing systems treat beats as isolated local instances. This is limiting because beat labels often depend on multi-beat rhythm context, including timing, compensatory pauses, and beat-to-beat morphological consistency. We present \textbf{DeepArrhythmia}, a tool-grounded multimodal framework for segment-contextualized beat-level ECG arrhythmia classification. Given a multi-beat ECG segment, DeepArrhythmia combines the raw ECG signal and a rendered waveform image, localizes R peaks to identify beat instances, and produces structured beat-level predictions. The framework decouples physiological measurement from evidence integration using specialized tools for beat localization, numerical rhythm--morphology extraction, and morphology-focused textual analysis. DeepArrhythmia uses segment-level confidence to route between minimal and rich evidence states, since richer physiological evidence is not uniformly useful. This agentic design integrates rhythm context, explicit physiological grounding, and selective evidence acquisition for decision making.

\end{abstract}

\section{Introduction}

Cardiac arrhythmias are common rhythm disorders associated with cardiovascular disease \citep{koppad2021arrhythmia} and are routinely detected and assessed using electrocardiography (ECG). As a non-invasive recording of cardiac electrical activity, ECG remains central to clinical arrhythmia evaluation, where recordings have traditionally been reviewed by experts to identify and label abnormal beats \citep{hong2020opportunities}. This clinical workflow has motivated extensive research on automated beat-level ECG arrhythmia classification, whose basic goal is to assign an arrhythmia class to each beat in an ECG recording ~\citep{ebrahimi2020review}. Despite substantial progress, many existing beat-level systems still treat each beat primarily as a local classification instance, using isolated beat crops or short beat-centered windows as input.

\begin{figure}[ht]
  \centering
  \includegraphics[width=0.75\linewidth]{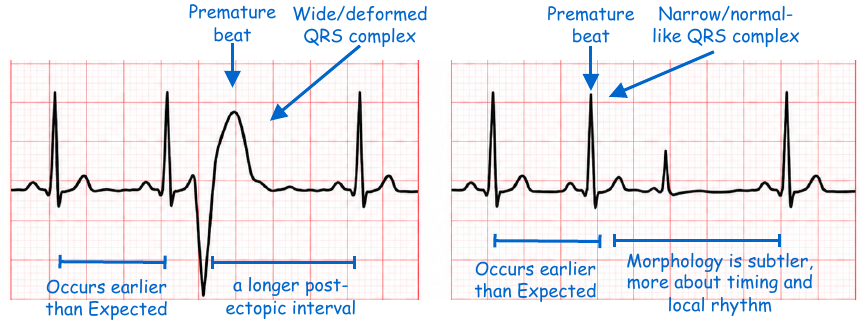}
\caption{VEB/SVEB discrimination requires morphology in context. }
  \label{fig:context-comparison}
\end{figure}

This local treatment is limited because beat-level arrhythmia classification is beat-instance prediction, but not purely beat-local. Each prediction corresponds to an individual beat instance, typically anchored by its R-peak location. The evidence needed to classify a beat often extends beyond the target complex to the surrounding rhythm context. As illustrated in Figure~\ref{fig:context-comparison}, ventricular ectopic beats (VEBs) and supraventricular ectopic beats (SVEBs) can both occur prematurely and produce a post-ectopic pause, so timing alone may be ambiguous. Their distinction often depends on contextual morphology: VEBs typically show a wide and aberrant QRS complex, whereas SVEBs tend to preserve a QRS morphology closer to neighboring normal beats. These examples motivate segment-contextualized beat-level classification, where each beat receives its own label while the decision can draw on morphology, rhythm timing, and beat-to-beat consistency within a multi-beat ECG segment.

This context dependence exposes two limitations in existing automated beat-level systems. First, isolated-beat and short-window methods preserve local morphology but underuse rhythm context, such as preceding and following R--R intervals, compensatory pauses, and cross-beat morphological consistency. Second, longer-context end-to-end models may process more signal context, but their predictions are often not explicitly grounded in physiological anchors or intermediate measurements. As a result, it is difficult to verify whether a decision is supported by recognizable ECG cues such as R--R dynamics, QRS morphology, P-wave behavior, or compensatory pauses. A natural way to address this limitation is to decouple physiological measurement from evidence integration: specialized ECG tools can provide precise beat localization and structured rhythm--morphology measurements, while a multimodal model can integrate heterogeneous evidence for beat-level decision making. However, richer physiological evidence should not be acquired indiscriminately. For easy segments, additional measurements may add computation without improving accuracy; for noisy or weakly informative signals, auxiliary evidence can even mislead the model. Thus, the key question is not only how to incorporate physiological evidence, but when such evidence should be acquired.

We propose \textbf{DeepArrhythmia}, a tool-grounded multimodal framework for segment-contextualized beat-level ECG arrhythmia classification. Given a multi-beat ECG segment, DeepArrhythmia uses both the raw ECG signal and a rendered ECG waveform image as complementary inputs. A Peak Detector first localizes R peaks to identify beat instances and provide temporal anchors for evidence alignment and structured output. The central multimodal agent then produces beat-level predictions under a minimal evidence state using the signal, image, and detected beat anchors. When the segment-level confidence indicates that the current evidence may be insufficient, DeepArrhythmia selectively invokes a Feature Extractor and a Morphology Analyzer to obtain explicit numerical rhythm--morphology measurements and textual morphology evidence before producing refined beat-level predictions. 
This design combines multi-beat rhythm context, structured physiological grounding, and confidence-guided selective evidence acquisition within a unified agentic framework.
Building on the above, we make the following contributions:

\begin{enumerate}[leftmargin=*,itemsep=0.25em, topsep=0.25em, parsep=0pt, partopsep=0pt]
    \item We reformulate beat-level arrhythmia classification as segment-contextualized, peak-aligned prediction. Rather than classifying isolated beat crops, DeepArrhythmia predicts labels for R-peak-aligned beats within multi-beat ECG segments, allowing each decision to combine local morphology with surrounding rhythm context. %
    \item We introduce a tool-grounded multimodal ECG agent that separates physiological measurement from evidence integration. Specialized tools expose detected peaks, numerical rhythm--morphology measurements, and morphology-focused textual descriptions as structured intermediate evidence that is used both for beat-level prediction and for interpretation and error analysis.
    \item We propose confidence-guided selective evidence acquisition.  Motivated by the non-uniform marginal value of physiological evidence, DeepArrhythmia uses confidence as an evidence gate to balance under-evidenced decisions on difficult rhythm-dependent segments against over-evidenced decisions where weak auxiliary measurements add cost or noise when applied indiscriminately. The acquisition policy is learned through threshold-induced routed supervised fine-tuning, so the agent learns \emph{when} richer evidence is worth acquiring. %

\end{enumerate}
We do not claim novelty for the individual components in isolation; the contribution is the framework that couples these three elements for beat-level ECG classification.

\section{Related Work}

\textbf{Beat-Level Arrhythmia Classification and Rhythm Context}
Automated beat-level ECG arrhythmia classification has been studied extensively using both feature-based and end-to-end learning approaches. Feature-based methods typically extract handcrafted rhythm and morphology descriptors, such as RR intervals, R-peak amplitudes, and beat-shape features, and then apply conventional machine learning classifiers \citep{limam2017atrial,mondejar2019heartbeat}. In contrast, end-to-end approaches directly process raw ECG signals or transformed representations, such as time--frequency images, using deep neural networks \citep{huang2019ecg,rajkumar2019arrhythmia,hannun2019cardiologist,zhao2019deep}. These approaches achieve strong predictive performance, but two limitations remain common. Many operate on isolated beat crops or short windows, underusing surrounding rhythm context. Some performance may rely on patient-specific or intra-patient evaluation, limiting evidence of subject-disjoint generalization. In contrast, we study beat-level arrhythmia classification under subject-disjoint splits using 10-second segments that preserve rhythm.

\textbf{Tool-Grounded Multimodal ECG Reasoning and Structured Evidence}
Recent ECG-specific multimodal large language models (MLLMs) integrate ECG signals, rendered ECG images, and textual reasoning to support ECG understanding and improve interpretability. GEM~\citep{lan2025gem} aligns time-series and image representations for grounded ECG interpretation, while PULSE~\citep{liu2024teach} and ECG-R1~\citep{jin2026ecg} pursue ECG-oriented vision--language reasoning. Although MLLMs are effective at integrating heterogeneous evidence, they remain limited in fine-grained numerical tasks, including precise event localization and explicit physiological measurement~\citep{arbuzov2025beyond,fons2024evaluating,li2025peak}. ECG-Agent explores tool calling for multi-turn ECG interaction~\citep{chung2026ecg}; however, it primarily presents tool outputs to improve interpretability rather than incorporating them as structured intermediate evidence for downstream decision-making. In contrast, DeepArrhythmia separates physiological measurement from evidence integration, enabling arrhythmia classification that is both more accurate and more interpretable.

\textbf{Selective Evidence Acquisition and Adaptive Tool Use}
Recent agentic frameworks~\cite{yao2023react,jin2025search,chung2026ecg} enable models to selectively invoke the tools most appropriate for downstream tasks. However, these approaches often depend heavily on the model's prior knowledge and on carefully designed prompts. In expert domains such as arrhythmia classification, such reliance can limit reliable tool use, particularly when the triggering conditions are implicit, as in variations in the complexity of ECG time series. DeepArrhythmia instead explicitly leverages the relationship between model confidence and predictive accuracy, using confidence as an evidence gate for selective tool invocation and thereby balancing predictive performance against computational cost.

\section{Method}
\label{sec:method}

\begin{figure}[ht]
  \centering
    \includegraphics[width=\linewidth]{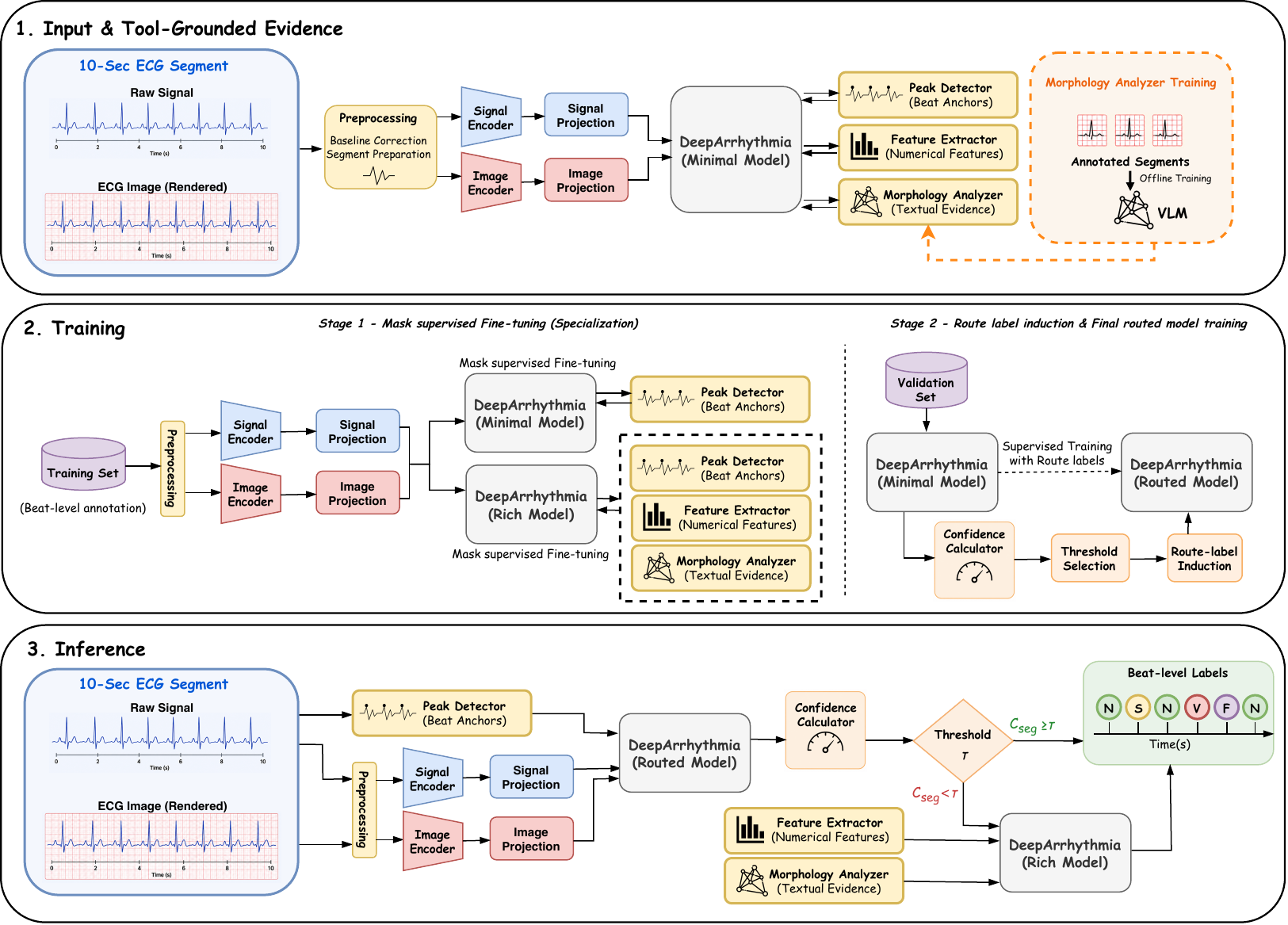}%
\caption{\textbf{Overview of the DeepArrhythmia framework.}
\textbf{(1)} Multimodal ECG inputs with three tool-grounded evidence
sources. \textbf{(2)} Two-stage mask-SFT yields Minimal, Rich, and
Routed specialists. \textbf{(3)} At inference, confidence
$C_{\mathrm{seg}}$ vs.\ threshold $\tau$ routes between minimal and
rich evidence.}
  \label{fig:deeparrhythmia_framework}
\end{figure}

\textbf{Problem Formulation and Evidence States}
Let \(x=(x_{\mathrm{ts}},x_{\mathrm{img}})\) denote a time-normalized ECG context segment of fixed duration \(T=10\) seconds. For a dataset \(d\) with sampling frequency \(f_d\), the signal view is represented on the native sampling grid as \(x_{\mathrm{ts}}\in\mathbb{R}^{C\times L_d}\), with \(L_d=\lfloor T f_d\rfloor\), where \(C\) is the number of ECG input channels or leads. Thus, \(L_d\) may vary across datasets, but every segment spans the same physical \(T\)-second context. The rendered waveform view \(x_{\mathrm{img}}\in\mathbb{R}^{H\times W\times 3}\) is generated from the same \(T\) interval.
Each segment \(x\) contains a variable number of beat instances \(B_x=\{b_{x,k}\}_{k=1}^{K_x}\), where \(K_x\) depends on the subject's heart rate and rhythm within the \(T\) window. The goal is to assign an arrhythmia label \(y_{x,k}\in\mathcal{Y}\) to each beat instance \(b_{x,k}\), where \(\mathcal{Y}=\{N,S,V,F\}\) denotes normal, supraventricular ectopic, ventricular ectopic, and fusion beats under the AAMI mapping. The model outputs a structured beat-level sequence \(\hat{Y}(x)=\{[a_{x,k}:\hat{y}_{x,k}]\}_{k=1}^{K_x}\), where \(a_{x,k}\) is the temporal anchor used to identify beat \(b_{x,k}\) in the serialized output. In practice, these anchors are instantiated as R-peak sample positions by the Peak Detector described below. The task is therefore beat-level classification conditioned on segment-level context, rather than sample-wise labeling or segment-level classification. The overview of the DeepArrhythmia framework is shown in Figure~\ref{fig:deeparrhythmia_framework}.

\subsection{Tool-Grounded Evidence Interfaces and Multimodal Integration}

DeepArrhythmia separates physiological measurement from multimodal evidence integration. Specialized ECG tools expose structured physiological evidence, while the central multimodal agent consumes this evidence together with signal and image tokens to produce peak-aligned beat predictions. The central agent receives ECG signal tokens, rendered ECG image tokens, and serialized tool outputs. Signal and image views are encoded separately and projected into the language-model embedding space. Tool interactions are represented with explicit \texttt{<Call>} and \texttt{<Output>} spans. The final answer is emitted as structured beat-level text, e.g., \([a_1:N]\;[a_2:S]\;\cdots\;[a_K:V]\).

\textbf{Beat-anchor interface.}
The \textit{Peak Detector} instantiates the abstract beat anchors \(A_x\) by localizing candidate R peaks from \(x_{\mathrm{ts}}\). These anchors determine where beat-level labels are emitted, and they also define the alignment for numerical feature extraction and morphology analysis.

\textbf{Numerical physiology interface.}
Given \(x_{\mathrm{ts}}\) and \(A_x\), the \textit{Feature Extractor} produces per-beat rhythm and morphology measurements, \(Z_{\mathrm{feat}}(x)=T_{\mathrm{feat}}(x_{\mathrm{ts}},A_x)\), including RR intervals, normalized RR intervals, R-peak amplitude, higher-order statistics, and numerical morphology descriptors. These measurements provide explicit numerical grounding for rhythm and morphology cues.

\textbf{Textual morphology interface.}
The \textit{Morphology Analyzer} produces morphology-aware textual evidence, \(Z_{\mathrm{morph}}(x)=T_{\mathrm{morph}}(x_{\mathrm{img}},A_x,Z_{\mathrm{feat}})\), from beat-centered ECG image crops and auxiliary context. It is trained by split-safe teacher distillation on training subjects only and receives no ground-truth label at inference time. Its output is used as supportive morphology-aware evidence rather than as a fully independent explanation. The same intermediate outputs are both consumed by the central agent for prediction and retained as auditable evidence for interpretation and error analysis. Thus, the central agent acts as an evidence integrator and decision maker, rather than a standalone ECG signal processor.

\subsection{Confidence-Guided Selective Evidence Acquisition}

DeepArrhythmia operationalizes selective evidence acquisition as a routing decision between two physiological evidence states. After beat anchors have been obtained, the model first predicts under a minimal peak-guided evidence state and then uses the \textit{Confidence Calculator} to decide whether the current evidence appears sufficient or whether optional richer evidence should be acquired. The Confidence Calculator is the fourth tool interface in DeepArrhythmia, but unlike the Peak Detector, Feature Extractor, and Morphology Analyzer, it does not produce new physiological measurements. Instead, it acts as an evidence-gating tool that controls the transition between evidence states.

We define the minimal evidence state contains the synchronized signal view, rendered waveform view, and beat anchors: \(E_{\min}(x)=\{x_{\mathrm{ts}},x_{\mathrm{img}},A_x\}\), where \(A_x=\{a_{x,k}\}_{k=1}^{K_x}\) denotes the beat anchors returned by the Peak Detector. The rich evidence state augments this minimal state with structured numerical rhythm--morphology measurements and textual morphology evidence: \(E_{\mathrm{rich}}(x)=E_{\min}(x)\cup\{Z_{\mathrm{feat}}(x),Z_{\mathrm{morph}}(x)\}\). Thus, the routing problem is not whether to classify the segment, but whether to classify it using only \(E_{\min}\) or to acquire \(E_{\mathrm{rich}}\) before the final prediction.

The model first generates an initial structured prediction under \(E_{\min}\). For each beat anchor \(a_{x,k}\), we obtain a class posterior \(p_{x,k}\in[0,1]^{|\mathcal{Y}|}\) from the language-model logits at the corresponding label-generation position in the structured output. Concretely, we apply a softmax over the canonical class tokens \(\mathcal{Y}=\{N,S,V,F\}\), and the initial predicted label is \(\hat{y}^{\min}_{x,k}=\arg\max_{y\in\mathcal{Y}}p_{x,k,y}\). The Confidence Calculator converts these beat-level posteriors into a segment-level evidence-sufficiency score. The beat-wise confidence is \(c_{x,k}=\max_{y\in\mathcal{Y}}p_{x,k,y}\), and the segment-level confidence is computed by averaging over the \(K_x\) beats in the segment: \(C_{\mathrm{seg}}(x)=\frac{1}{K_x}\sum_{k=1}^{K_x}c_{x,k}\). We route once per segment rather than once per beat, because the optional Feature Extractor and Morphology Analyzer operate on segment-level rhythm and morphology context. Appendix~\ref{app:Routing Scheme Analysis} compares alternative segment-level confidence statistics, and Appendix~\ref{app:threshold-sensitivity} analyzes threshold sensitivity, calibration, and routing rates.%

Given a dataset-specific threshold \(\tau_d\), DeepArrhythmia follows the policy
\[
\pi(x)=
\begin{cases}
E_{\min}(x), & C_{\mathrm{seg}}(x)\ge \tau_d,\\
E_{\mathrm{rich}}(x), & C_{\mathrm{seg}}(x)< \tau_d .
\end{cases}
\]
If \(C_{\mathrm{seg}}(x)\ge\tau_d\), the model returns the initial minimal-evidence prediction \(\hat{Y}_{\min}(x)=\{[a_{x,k}:\hat{y}^{\min}_{x,k}]\}_{k=1}^{K_x}\). If \(C_{\mathrm{seg}}(x)<\tau_d\), the model invokes the optional evidence-producing tools, obtains \(Z_{\mathrm{feat}}\) and \(Z_{\mathrm{morph}}\), and produces a refined prediction under \(E_{\mathrm{rich}}\): \(\hat{Y}_{\mathrm{rich}}(x)=\{[a_{x,k}:\hat{y}^{\text{rich}}_{x,k}]\}_{k=1}^{K_x}\).

This policy treats confidence as a practical proxy for evidence sufficiency, not as a direct estimator of the true utility of additional tools. The design is intended to balance two failure modes: under-evidence, where difficult rhythm-dependent segments are classified from insufficient physiological context, and over-evidence, where optional measurements are applied uniformly even when they may be redundant, costly, or noisy. Appendix~\ref{app:Example of Complete Pipeline for DeepArrhythmia} provides a complete example of this inference pipeline.

\subsection{Two-stage Training and Implementation}
\label{sec:training_strategy}

We train DeepArrhythmia with two stages of masked supervised fine-tuning over serialized tool-use trajectories. Each trajectory contains generated tool-call tokens, externally returned tool-output spans, and a final structured beat-level answer. 
For each dataset, we first construct \textbf{\textit{subject-disjoint training and test partitions}}. The test subjects are never used for tool distillation, threshold selection, or supervised fine-tuning. We further split the training subjects into a Stage~1 optimization split \(D_1\) and a held-out Stage~2 route-induction split \(D_2\) using a 9:1 ratio. The split \(D_2\) is held out from Stage~1 training and is used only to select the confidence threshold and construct routed SFT trajectories. The selected threshold is fixed before test-time evaluation.
We apply shift-based segment re-anchoring on the training split to mitigate class imbalance. This augmentation preserves the fixed 10-second context horizon while increasing exposure to minority-class beats at different positions within the segment. 
Details are provided in Appendix~\ref{app:shift_augmentation}.

\textbf{Masked SFT objective.}
Let \(\tau^*=(\tau^*_1,\ldots,\tau^*_N)\) denote a target token trajectory and let \(m_n\in\{0,1\}\) indicate whether token \(\tau^*_n\) is supervised. The masked SFT loss is
$
\mathcal{L}_{\mathrm{SFT}}(\theta)
=
-\sum_{(x,\tau^*)}
\sum_{n=1}^{N}
m_n
\log p_{\theta}(\tau^*_n \mid x,\tau^*_{<n}).
$
We set \(m_n=1\) for generated tool-call tokens and final-answer tokens, and \(m_n=0\) for externally returned tool-output spans and other context tokens. Thus, the model is trained to decide which tools to call and to generate the final structured beat-level prediction, while deterministic tool outputs are consumed rather than imitated.

\textbf{Stage 1: evidence-state specialization.}
Stage~1 trains two branch-specific agents corresponding to the two evidence states. The minimal-evidence agent \(M_{\min}\) is trained on short trajectories that call the Peak Detector, receive beat anchors \(A_x\), and then emit the final structured prediction under
$
E_{\min}(x)
$
The rich-evidence agent \(M_{\mathrm{rich}}\) is trained on trajectories that call the Peak Detector, Feature Extractor, and Morphology Analyzer before producing the final prediction under
$
E_{\mathrm{rich}}(x)
$
This stage teaches the model the output syntax, the alignment between beat anchors and class labels, and how to predict under both minimal and rich evidence budgets before adaptive routing is introduced.

\textbf{Stage 2: confidence-gated routed SFT.}
Stage~2 trains the routed agent \(M_{\mathrm{route}}\) to follow the confidence-gated acquisition policy. We first run \(M_{\min}\) on the held-out route-induction split \(D_2\). For each segment, the Confidence Calculator computes the segment-level score from the minimal-evidence class posteriors. We also run \(M_{\mathrm{rich}}\) on the same split to obtain the prediction that would result from acquiring full rich evidence.
For dataset \(d\), we sweep candidate thresholds and select
$
\tau_d
=
\arg\max_{\tau}
\mathrm{MicroF1}
\left(
\hat{Y}_{\tau};D_2
\right),
$
where $\hat{Y}_{\tau}(x)$ is either $\hat{Y}_{\min}(x)$ or $\hat{Y}_{\mathrm{rich}}(x)$.
The selected threshold induces deterministic route labels on \(D_2\): segments above the threshold stop after minimal evidence, whereas segments below the threshold acquire the Feature Extractor and Morphology Analyzer outputs before final prediction. These route labels are threshold-induced evidence-acquisition decisions, not normal-versus-abnormal labels.

We then construct routed trajectories for \(M_{\mathrm{route}}\). Each trajectory calls the Peak Detector, produces an initial minimal-evidence prediction, queries the Confidence Calculator, and receives the resulting confidence score. If \(C_{\mathrm{seg}}(x)\ge\tau_d\), the trajectory terminates with the minimal prediction. If \(C_{\mathrm{seg}}(x)<\tau_d\), the trajectory continues by invoking the Feature Extractor and Morphology Analyzer and then emits the rich-evidence prediction. We initialize \(M_{\mathrm{route}}\) from \(M_{\min}\) to preserve the minimal-evidence prediction behavior and then fine-tune it on these routed trajectories. At inference time, the same threshold \(\tau_d\) is fixed and applied without access to test labels.

\textbf{Implementation.}
We instantiate Qwen3.5-4B as Multimodal LLM backbone and use pretrained ECG-CoCa encoder for signal modality~\citep{zhao2025ecgchat}. The ECG-CoCa encoder is kept frozen during DeepArrhythmia training, so optimization focuses on multimodal evidence integration, tool-use generation, and the evidence-acquisition policy.  Details are in Appendix~\ref{app:implementaiton}.

\section{Experiments}
\label{experiment}

\textbf{Datasets and baselines.}
We evaluate model performance on four electrocardiogram (ECG) datasets: MIT-BIH Arrhythmia \citep{moody2001impact}, MIT-BIH Supraventricular \citep{greenwald1990improved}, INCART \citep{tihonenko2007st}, and VitalDB Arrhythmia \citep{eun2026vitaldb}.
We compare DeepArrhythmia against four baseline families. The first is a conventional machine-learning baseline, SVM~\citep{mondejar2019heartbeat}. The second is a set of deep-learning classifiers: 2D CNN~\citep{limam2017atrial}, STFT 2D CNN~\citep{huang2019ecg}, 1D ResNet~\citep{hannun2019cardiologist}, Transformer~\citep{hu2022transformer}, xECG~\citep{lunelli2025benchecg}, and TCN~\citep{ingolfsson2021ecg}. The third is a set of general-purpose multimodal language models: Gemini 3.1~\citep{team2023gemini}, GPT-5.4~\citep{singh2025openai}, Qwen3.5-4B~\citep{bai2023qwen}, and Gemma-3-27B~\citep{gemma2025gemma3}. The fourth is a set of ECG-oriented vision-language models: PULSE~\citep{liu2024teach}, GEM~\citep{lan2025gem}, and ECG-R1~\citep{jin2026ecg}. For language-model-based baselines, we append the outputs of the Peak Detector and Feature Extractor to the input prompt to ensure a fair comparison under matched tool-derived evidence. The Morphology Analyzer is excluded because it is trained with dataset-specific supervision, which would compromise the fairness of this comparison.

\textbf{Non-agentic baseline ablation.}
We evaluate matched non-agentic fusion baselines to isolate the contribution of each modality. Specifically, we replace the decoder-only backbone with a fusion classifier that consumes the same input ingredients but performs no tool calling or sequential decision making. The model encodes each 10-second ECG segment with the ECG-CoCa tower and uses precomputed Qwen3.5-4B embeddings for the rendered ECG image and text evidence. For each detected R peak, it pools a beat-local ECG representation, concatenates it with the global ECG representation, image embedding, structured peak-derived numerical features, and optional morphology/prompt text embedding, and then applies a supervised fusion head to predict a label for each beat.

\textbf{Experiment setting.}
For data partitioning, we adopt the DS1/DS2 protocol \citep{mondejar2019heartbeat}, which separates training and test sets by subject identity with balanced numbers of IDs. This split prevents subject overlap between training and testing and reduces information leakage.
Following prior ECG annotation practice, we exclude class \(Q\), which denotes
unknown or unclassifiable beats. We therefore report results on the clinically
meaningful beat-label set \(\mathcal{Y} = \{N, S, V, F\}\), where \(N\), \(S\),
\(V\), and \(F\) denote normal, supraventricular ectopic, ventricular ectopic,
and fusion beats, respectively.
All four datasets are highly imbalanced under the coarse AAMI mapping; full beat-level class statistics are provided in Appendix~\ref{app:dataset_class_distribution}. To assess whether the observed gains can be attributed to data augmentation alone, Appendix~\ref{app:baseline_augmentation} reports a complementary evaluation of the augmentation strategy on the TCN deep-learning baseline.
A repository containing the implementation, configurations, and reproduction instructions is available at 
\url{https://github.com/jimmylihui/DeepArrhythmia_repo}.

\textbf{DeepArrhythmia variants}
We compare the minimal- and rich-specialization models with the threshold-induced routed model. For additional analysis, we also include a simple routing baseline. This baseline is trained on trajectories derived from the Stage~2 split, where each sample is assigned to the rich-evidence trajectory if the rich-specialization model yields a more accurate prediction than the minimal-specialization model, and to the minimal-evidence trajectory otherwise.

\subsection{Main Results: Context, Evidence, and Acquisition Policy}

\begin{table*}[t]
  \centering
  \caption{Performance comparison across datasets using Macro-F1 and Micro-F1. Best results are shown in bold, and second-best results are underlined. For machine learning and deep learning models, we report the mean over three random seeds. For VLMs, we use a decoding temperature of 0. For datasets with absent classes in the test split, Macro-F1 is averaged over classes present in the ground truth.}
  \label{tab:result-performance}
  \footnotesize
  \resizebox{\textwidth}{!}{
  \begin{tabular}{lcccccccc}
    \toprule
    & \multicolumn{2}{c}{MIT-BIH} & \multicolumn{2}{c}{MIT-BIH-SUP} & \multicolumn{2}{c}{INCART} & \multicolumn{2}{c}{VitalDB} \\
    \cmidrule(r){2-3} \cmidrule(r){4-5} \cmidrule(r){6-7} \cmidrule(r){8-9}
    \textbf{Model} & \textbf{Ma} & \textbf{Mi} & \textbf{Ma} & \textbf{Mi} & \textbf{Ma} & \textbf{Mi} & \textbf{Ma} & \textbf{Mi} \\
    \midrule

    \rowcolor{gray!15} \multicolumn{9}{l}{\textit{DeepArrhythmia}} \\
    Rich-specialization model & \textbf{0.593} & \underline{0.950} & \underline{0.614} & \underline{0.952} & \textbf{0.710} & \underline{0.981} & 0.858 & 0.911 \\

    Simple routing baseline & 0.566 & 0.943 & 0.613 & \textbf{0.953} & \underline{0.692} & 0.979 & 0.861 & 0.914 \\

    Threshold-induced routed model & \underline{0.588} & \textbf{0.951} & \textbf{0.615} & \textbf{0.953} & 0.689 & 0.979 & \textbf{0.885} & \underline{0.922} \\

    Minimal-specialization model & 0.534 & 0.917 & 0.562 & 0.938 & 0.620 & 0.963 & \underline{0.877} & \textbf{0.923} \\
    \midrule

    \rowcolor{gray!15} \multicolumn{9}{l}{\textit{Matched non-agentic fusion baseline}} \\

    Signal + image + peaks & 0.456 & 0.856 & 0.361 & 0.739 & 0.536 & 0.910 & 0.631 & 0.792 \\
    Signal + image + peaks + features & \textbf{0.593} & 0.928 & 0.516 & 0.912 & 0.637 & 0.966 & 0.770 & 0.873 \\
    Signal + image + peaks + features + morphology & 0.567 & 0.937 & 0.536 & 0.926 & 0.641 & 0.967 & 0.821 & 0.900 \\

    \rowcolor{gray!15} \multicolumn{9}{l}{\textit{Machine learning model}} \\
    SVM~\citep{mondejar2019heartbeat} & 0.523 & 0.934 & 0.575 & 0.938 & 0.681 & \textbf{0.985} & 0.600 & 0.747 \\
    \midrule

    \rowcolor{gray!15} \multicolumn{9}{l}{\textit{Deep learning models}} \\
    2D CNN~\citep{limam2017atrial} & 0.423 & 0.839 & 0.486 & 0.829 & 0.587 & 0.938 & 0.618 & 0.780 \\
    STFT CNN~\citep{huang2019ecg} & 0.366 & 0.853 & 0.395 & 0.832 & 0.462 & 0.930 & 0.373 & 0.682 \\
    1D ResNet~\citep{hannun2019cardiologist} & 0.362 & 0.760 & 0.467 & 0.814 & 0.501 & 0.869 & 0.654 & 0.786 \\
    Transformer~\citep{hu2022transformer} & 0.363 & 0.907 & 0.427 & 0.913 & 0.513 & 0.962 & 0.746 & 0.850 \\
    xECG~\citep{lunelli2025benchecg} & 0.440 & 0.893 & 0.512 & 0.901 & 0.523 & 0.934 & 0.698 & 0.820 \\
    TCN~\citep{ingolfsson2021ecg} & 0.444 & 0.704 & 0.451 & 0.783 & 0.444 & 0.831 & 0.676 & 0.811 \\
    \midrule

    \rowcolor{gray!15} \multicolumn{9}{l}{\textit{Closed-source VLMs}} \\
    Gemini 3.1~\citep{team2023gemini} & 0.410 & 0.878 & 0.522 & 0.849 & 0.653 & 0.924 & 0.389 & 0.693 \\
    GPT-5.4~\citep{singh2025openai} & 0.352 & 0.835 & 0.307 & 0.727 & 0.428 & 0.920 & 0.252 & 0.664 \\
    \midrule

    \rowcolor{gray!15} \multicolumn{9}{l}{\textit{Open-source VLMs}} \\
    Qwen3.5-4B~\citep{bai2023qwen} & 0.414 & 0.785 & 0.443 & 0.865 & 0.517 & 0.861 & 0.373 & 0.709 \\
    Gemma-3-27B~\citep{gemma2025gemma3} & 0.378 & 0.880 & 0.277 & 0.817 & 0.367 & 0.878 & 0.209 & 0.608 \\
    \midrule

    \rowcolor{gray!15} \multicolumn{9}{l}{\textit{ECG VLMs}} \\
    PULSE~\citep{liu2024teach} & 0.211 & 0.646 & 0.226 & 0.660 & 0.216 & 0.670 & 0.242 & 0.568 \\
    GEM~\citep{lan2025gem} & 0.205 & 0.459 & 0.194 & 0.468 & 0.273 & 0.465 & 0.285 & 0.398 \\
    ECG-R1~\citep{jin2026ecg} & 0.199 & 0.446 & 0.116 & 0.237 & 0.270 & 0.532 & 0.243 & 0.515 \\
    \bottomrule
  \end{tabular}
  }
\end{table*}

Table~\ref{tab:result-performance} shows that DeepArrhythmia variants achieve the best or tied-best performance across datasets and metrics. These comparisons highlight three main advantages of the framework.

\textbf{Contextual information improves performance.}
DeepArrhythmia explicitly leverages segment-level context and consistently outperforms the conventional machine-learning, deep-learning, and VLM baselines. The value of contextual information is further supported by Appendix~\ref{app:context_length}, where increasing the available context length leads to consistent improvements in performance. This interpretation is also reinforced by the morphology-analyzer statistics and qualitative examples in Appendix~\ref{app:Interpretation Analysis} and Appendix~\ref{app:interpretation_examples}, which show that rhythm-related cues and cross-beat morphological consistency frequently contribute to model decisions.

\textbf{Context must be processed appropriately.}
These results indicate that contextual information is beneficial only when it is represented and modeled appropriately. On the representation side, the matched non-agentic fusion baselines improve substantially when structured features are added, and the morphology modality yields further gains on three of the four datasets, even though both are ultimately derived from the same raw ECG signal. On the modeling side, however, access to additional evidence alone is insufficient: SVM and VLM baselines can consume contextual information, yet they remain well below the strongest DeepArrhythmia variants. Even under closely matched inputs, DeepArrhythmia outperforms the fusion baselines (on MIT-BIH, Macro/Micro-F1 0.593/0.950 vs.\ 0.567/0.937 for the non-agentic fusion baseline with the full evidence set including morphology), and even without the Morphology Analyzer it exceeds the SVM and Gemini 3.1 (Macro-F1 0.5694 vs.\ 0.523 and 0.410; Table~\ref{tab:ablation-study}), suggesting that model architecture plays a central role in translating heterogeneous evidence into accurate beat-level predictions. This interpretation is further supported by Appendix~\ref{app:Faithfulness Evaluation}, where perturbing the evidence for a target beat changes the prediction for that beat while leaving neighboring beats largely unaffected, indicating a targeted link between evidence and classification. Model capacity is likewise important: Appendix~\ref{app:model_scale} shows that performance improves with backbone size (0.8B to 4B).

\textbf{Not all additional evidence is equally useful.}
Additional evidence is most helpful for difficult segments, but it does not provide uniform benefit for every sample. Appendix~\ref{app:confidence_distribution} shows that richer evidence improves performance primarily on low-confidence segments, whereas high-confidence segments benefit much less. Moreover, low-quality evidence can be detrimental. As discussed in Appendix~\ref{Appendix:VitalDB Result Analysis}, the weaker utility of VitalDB features suggests lower signal quality or noisier labels, and this is consistent with the observation that the rich-specialization model underperforms the minimal-specialization model on VitalDB. This dataset-dependent behavior motivates selective evidence acquisition rather than always invoking every tool. By combining confidence thresholds with routing, the threshold-induced routed model consistently matches or exceeds the simple routing baseline and achieves the best or tied-best performance among DeepArrhythmia variants while reducing reliance on uninformative evidence.

Beyond the main within-dataset benchmarks, Appendix~\ref{app:model_scale} shows that performance improves with backbone size. The gains are statistically reliable: 95\% bootstrap confidence intervals on MIT-BIH Macro-F1 do not overlap (DeepArrhythmia 0.593 [0.583, 0.603] vs.\ SVM 0.523 [0.516, 0.530]), and the ordering is preserved under balanced accuracy (Appendix~\ref{app:detail analysis}). The framework is also robust to the routing threshold, whose confidence is well calibrated (ECE 0.024--0.058; Appendix~\ref{app:threshold-sensitivity}), to peak-detector errors (Appendix~\ref{app:peak detector}), and to signal noise (Appendix~\ref{app: Impact of Noise}). We also identify two important limitations. Appendix~\ref{app:cross_dataset} demonstrates limited cross-dataset generalization, although DeepArrhythmia degrades less than the baselines (mean off-diagonal Micro-F1 0.698 vs.\ 0.457 for the 1D ResNet), and Appendix~\ref{app:detail analysis} shows that performance on extremely rare classes remains weak in absolute terms.

\subsection{Ablation and Adaptive Evidence Acquisition}

\begin{wraptable}[11]{r}{0.45\textwidth}
  \vspace{-\baselineskip}
  \centering
  \caption{Ablation on the MIT-BIH test split.}
  \label{tab:ablation-study}
  \resizebox{0.45\textwidth}{!}{%
  \setlength{\tabcolsep}{1pt}
  \begin{tabular}{@{}lcc@{}}
    \toprule
    Configuration & Macro-F1 & Micro-F1 \\
    \midrule
    Full DeepArrhythmia & 0.5929 & 0.9501 \\
    w/o shift augmentation & 0.4259 & 0.9120 \\
    \midrule
    \rowcolor{gray!15} \multicolumn{3}{l}{\textit{Modality and subsystem ablations}} \\
    \quad w/o ECG image & 0.5394 & 0.9426 \\
    \quad w/o Feature Extractor & 0.3527 & 0.9066 \\
    \quad w/o Morphology Analyzer & 0.5694 & 0.9433 \\
    \quad w/o Peak Detector & -- & -- \\
    \bottomrule
  \end{tabular}
  }
\end{wraptable}

\textbf{Ablation Study.}
Table~\ref{tab:ablation-study} summarizes the ablation results on MIT-BIH.
Shift augmentation is a critical component of DeepArrhythmia: removing it reduces Micro-F1 from 0.9501 to 0.9120, indicating that the model benefits substantially from improved data balance. The ablations further show that richer evidence sources---including the ECG image, Feature Extractor, and Morphology Analyzer---each contribute positively to performance, with the Feature Extractor having a particularly important role. Finally, the Peak Detector is indispensable because it provides the beat anchors required for downstream alignment and beat-level classification; without it, the model cannot localize candidate beats reliably enough to perform classification.

\textbf{Adaptive Evidence Acquisition.}
Table~\ref{tab:adaptive-tool-policy-comparison} compares the computational profiles of the rich-specialization model, minimal-specialization model, simple routing baseline, threshold-induced routed model, and rich-specialization non-agentic baseline. The minimal-specialization model is the most efficient variant, with an average latency of 2.86 seconds per sample across datasets. By contrast, the rich-specialization model is the most expensive because it always acquires the full set of evidence. The rich-specialization non-agentic baseline removes the decoder-only backbone, yet its computational cost remains high, with an average latency of 3.47 seconds per sample, indicating that inference cost is dominated largely by evidence acquisition rather than by autoregressive decoding alone. The simple routing baseline occupies an intermediate operating point between the rich- and minimal-specialization models; however, its latency varies only modestly across datasets, ranging from 3.24 to 3.48 seconds per sample, which suggests comparatively limited adaptation to dataset-specific evidence utility. In contrast, the threshold-induced routed model exhibits greater variation in computational cost, with latency ranging from 3.27 to 3.58 seconds per sample. This pattern is consistent with more selective routing: on datasets such as MIT-BIH, where richer evidence is more beneficial, the model allocates a larger proportion of samples to the rich-evidence trajectory, whereas on datasets such as VitalDB, where richer evidence is less informative, it routes more samples to the minimal-evidence trajectory to reduce computational cost.
Appendix~\ref{app:implementaiton} summarizes this device setup. All latency values are measured with three independent single-GPU A6000 workers, batch size 1, bfloat16 weights, and greedy decoding with up to 8192 new tokens.

\begin{table*}[t]
\centering
\caption{Computational cost and latency comparison of DeepArrhythmia variants across datasets.}
\label{tab:adaptive-tool-policy-comparison}
\footnotesize
\resizebox{\textwidth}{!}{%
\begin{tabular}{lcccccccccccc}
\toprule
& \multicolumn{3}{c}{MIT-BIH} & \multicolumn{3}{c}{MIT-BIH-SUP} & \multicolumn{3}{c}{INCART} & \multicolumn{3}{c}{VitalDB} \\
\cmidrule(lr){2-4} \cmidrule(lr){5-7} \cmidrule(lr){8-10} \cmidrule(lr){11-13}
Policy & A.\,Cal & T/s & Lat. & A.\,Cal & T/s & Lat. & A.\,Cal & T/s & Lat. & A.\,Cal & T/s & Lat. \\
\midrule
Rich-Spec.\ (Agentic)         & 3    & 648 & 3.89 & 3    & 646 & 3.85 & 3    & 652 & 3.94 & 3    & 666 & 4.02 \\
Rich-Spec.\ (Non-agentic)     & 3    & 573 & 3.40 & 3    & 570 & 3.39 & 3    & 580 & 3.50 & 3    & 595 & 3.59 \\
Simple Routing                & 1.78 & 365 & 3.24 & 1.99 & 409 & 3.28 & 2.24 & 474 & 3.53 & 1.92 & 413 & 3.48 \\
Threshold Routing$^{\mathrm{a}}$ & 3.41 & 510 & 3.58 & 3.20 & 459 & 3.40 & 3.07 & 436 & 3.45 & 2.52 & 319 & 3.27 \\
Minimal-Spec.                 & 1    & 183 & 2.83 & 1    & 178 & 2.73 & 1    & 186 & 2.88 & 1    & 197 & 3.01 \\
\bottomrule
\end{tabular}%
}
\par\smallskip
\begin{minipage}{\textwidth}
\footnotesize
\textit{A.\,Cal} = average tool calls per sample; \textit{T/s} = tokens per sample; \textit{Lat.} = seconds per sample; $^{\mathrm{a}}$ adds one tool call.
\end{minipage}
\end{table*}

\textbf{Expert evaluation of generated morphology analyses.}
We conducted a blinded expert evaluation of the generated morphology analyses to assess explanation quality and clinical plausibility; the full evaluation protocol is provided in Appendix~\ref{app:human_eval_protocol}. Across the three systems, 200 samples per system (600 total) were independently scored by three experts on seven criteria. We compared analyses generated by Gemini 3.1 Pro with beat-label conditioning, Gemini 3.1 Pro without beat-label conditioning, and the DeepArrhythmia Morphology Analyzer. As shown in Tables~\ref{tab:human_eval}, DeepArrhythmia received strong ratings for physiological correctness (4.92/5), use of segment context (4.39/5), clinical utility (4.22/5), and overall usefulness (4.19/5). Relative to Gemini 3.1 Pro without label conditioning, DeepArrhythmia is more competitive on physiological correctness and contextual grounding, but it remains below the label-conditioned Gemini 3.1 Pro teacher across all reported metrics. This pattern suggests that access to beat-label information enables the teacher model to produce more diagnostically explicit and evidentially complete analyses. Additional interpretation statistics and qualitative examples are provided in Appendix~\ref{app:Interpretation Analysis} and~\ref{app:interpretation_examples}, and Appendix~\ref{app:teacher_model_effect} examines the effect of alternative teacher models. Notably, the Morphology Analyzer is trained with label-conditioned teacher distillation but is used without label input at inference time.

\begin{table*}[ht]
\centering
\caption{Human expert evaluation of morphology analyses. Mean (STD) on a 5-point scale.}
\label{tab:human_eval}
\small
\setlength{\tabcolsep}{2.5pt}
\resizebox{\textwidth}{!}{%
\begin{tabular}{lcccc|cccc}
\toprule
& \multicolumn{4}{c|}{Clinical Metrics} & \multicolumn{4}{c}{Utility \& Safety Metrics} \\
\cmidrule(lr){2-5} \cmidrule(lr){6-9}
Model & Diag. Supp. & ECG Evid. & Physiol. Corr. & Seg. Context & Evid. Compl. & Uncert. \& Safety & Clin. Utility & Overall \\
\midrule
Gemini (w/ labels)  & 4.98 (0.14) & 5.00 (0.00) & 5.00 (0.00) & 5.00 (0.00) & 4.92 (0.31) & 4.88 (0.33) & 4.99 (0.10) & 4.97 (0.19) \\
Gemini (wo labels)  & 4.60 (1.01)          & 4.68 (0.84)          & 4.84 (0.58)          & 4.37 (0.84)          & 4.38 (0.94)          & 4.74 (0.73)          & 4.67 (0.89)          & 4.61 (0.86)          \\
DeepArrhythmia              & 3.88 (0.62)          & 4.25 (0.89)          & 4.92 (0.27)          & 4.39 (0.75)          & 3.51 (0.90)          & 4.14 (0.78)          & 4.22 (0.91)          & 4.19 (0.86)          \\
\bottomrule
\end{tabular}%
}
\end{table*}

\section{Conclusion and Limitations}
\label{sec:Conclusion and Limitation}

We presented \textbf{DeepArrhythmia}, a tool-grounded multimodal framework for segment-contextualized beat-level ECG arrhythmia classification from multi-beat ECG segments. The central idea is to treat each prediction as a beat-instance decision while allowing the evidence for that decision to come from surrounding rhythm context, structured physiological measurements, and morphology-focused textual evidence. DeepArrhythmia decouples physiological measurement from evidence integration through specialized tools for beat localization, numerical rhythm--morphology extraction, and morphology analysis, while using confidence-guided routing to decide when richer evidence should be acquired. Across four public arrhythmia datasets, the results show that multi-beat context and structured physiological grounding improve beat-level classification, and that adaptive evidence acquisition provides a practical alternative to always invoking all tools. Beyond predictive performance, the retained intermediate evidence offers a basis for interpretation, error analysis, and expert review. DeepArrhythmia targets offline and retrospective analysis and clinical decision support, such as Holter review and second reading of flagged segments, rather than hard real-time on-device alarming.

Several limitations remain. The confidence-guided routing policy may be vulnerable to domain shift, because confidence calibration can vary across devices, sampling rates, patient populations, and noise conditions. Future work should therefore replace fixed thresholds with more robust uncertainty-aware routing, such as label-free or conformal threshold selection, and validate the framework on multi-center cohorts. Performance on extremely rare classes remains limited, especially for fusion beats, and improving recognition of these classes, for example through generative synthesis of realistic fusion morphology, remains an important open challenge.

\begin{ack}
The authors declare no competing interests.
\end{ack}

\bibliographystyle{plainnat}
\bibliography{references}

\appendix

\clearpage
\onecolumn
\thispagestyle{plain}

\phantomsection
\addcontentsline{toc}{section}{Appendix Index}

\begin{center}
{\Large\bfseries DeepArrhythmia: Segment-Contextualized ECG
Arrhythmia Classification via Selective Evidence
Acquisition}
\end{center}

\vspace{2.2em}

\begingroup
\small
\bfseries
\renewcommand{\arraystretch}{1.55}
\setlength{\tabcolsep}{0pt}

\begin{tabularx}{\textwidth}{@{}Xr@{}}

\hyperref[app:implementaiton]{\textcolor{red}{A\quad Implementation Details}}
&
\textbf{\pageref{app:implementaiton}}
\\

\hyperref[app:method_details]{\textcolor{red}{B\quad Additional Method Details}}
&
\textbf{\pageref{app:method_details}}
\\

\hyperref[app:baseline_augmentation]{\textcolor{red}{C\quad Baseline Augmentation}}
&
\textbf{\pageref{app:baseline_augmentation}}
\\

\hyperref[Appendix:VitalDB Result Analysis]{\textcolor{red}{D\quad VitalDB Result Analysis}}
&
\textbf{\pageref{Appendix:VitalDB Result Analysis}}
\\

\hyperref[app:Routing Scheme Analysis]{\textcolor{red}{E\quad Routing Scheme Analysis}}
&
\textbf{\pageref{app:Routing Scheme Analysis}}
\\

\hyperref[app:threshold-sensitivity]{\textcolor{red}{F\quad Sensitivity to the Routing Threshold and Calibration}}
&
\textbf{\pageref{app:threshold-sensitivity}}
\\

\hyperref[app:dataset_class_distribution]{\textcolor{red}{G\quad Dataset Class Distribution}}
&
\textbf{\pageref{app:dataset_class_distribution}}
\\

\hyperref[app:impact of data spliting]{\textcolor{red}{H\quad Impact of Data Splitting}}
&
\textbf{\pageref{app:impact of data spliting}}
\\

\hyperref[app:confidence_distribution]{\textcolor{red}{I\quad Accuracy Distribution by Prediction Confidence}}
&
\textbf{\pageref{app:confidence_distribution}}
\\

\hyperref[app:cross_dataset]{\textcolor{red}{J\quad Cross-Dataset Performance}}
&
\textbf{\pageref{app:cross_dataset}}
\\

\hyperref[app:morph_kd]{\textcolor{red}{K\quad Knowledge Distillation for the Morphology Analyzer}}
&
\textbf{\pageref{app:morph_kd}}
\\

\hyperref[app:Example of Complete Pipeline for DeepArrhythmia]{\textcolor{red}{L\quad Example of the Complete Pipeline for DeepArrhythmia}}
&
\textbf{\pageref{app:Example of Complete Pipeline for DeepArrhythmia}}
\\

\hyperref[app:detail analysis]{\textcolor{red}{M\quad Detailed Result Analysis for DeepArrhythmia}}
&
\textbf{\pageref{app:detail analysis}}
\\

\hyperref[app:Faithfulness Evaluation]{\textcolor{red}{N\quad Faithfulness Evaluation}}
&
\textbf{\pageref{app:Faithfulness Evaluation}}
\\

\hyperref[app:context_length]{\textcolor{red}{O\quad Impact of Context Length}}
&
\textbf{\pageref{app:context_length}}
\\

\hyperref[app:peak detector]{\textcolor{red}{P\quad Impact of the Peak Detector}}
&
\textbf{\pageref{app:peak detector}}
\\

\hyperref[app: Impact of Noise]{\textcolor{red}{Q\quad Impact of Noise}}
&
\textbf{\pageref{app: Impact of Noise}}
\\

\hyperref[app:model_scale]{\textcolor{red}{R\quad Impact of Backbone Scale}}
&
\textbf{\pageref{app:model_scale}}
\\

\hyperref[app:Interpretation Analysis]{\textcolor{red}{S\quad Interpretation Analysis}}
&
\textbf{\pageref{app:Interpretation Analysis}}
\\

\hyperref[app:interpretation_examples]{\textcolor{red}{T\quad Interpretation Examples}}
&
\textbf{\pageref{app:interpretation_examples}}
\\

\hyperref[app:teacher_model_effect]{\textcolor{red}{U\quad Effect of the Teacher Model in the Morphology Analyzer}}
&
\textbf{\pageref{app:teacher_model_effect}}
\\

\hyperref[app:broader_impacts_limitations]{\textcolor{red}{V\quad Broader Impacts and Limitations}}
&
\textbf{\pageref{app:broader_impacts_limitations}}
\\

\hyperref[app:human_eval_protocol]{\textcolor{red}{W\quad Human Evaluation Protocol}}
&
\textbf{\pageref{app:human_eval_protocol}}
\\

\end{tabularx}
\endgroup
\normalsize
\clearpage

\section{Implementation Details}
\label{app:implementaiton}
\begin{table}[ht]
\centering
\caption{Training configuration used for supervised fine-tuning of DeepArrhythmia.}
\label{tab:training_configs}
\begin{tabular}{ll}
\toprule
\textbf{Parameter} & \textbf{Configuration / Value} \\ 
\midrule
Backbone model & Qwen3.5-4B \\
ECG encoder & Pretrained ECG-CoCa checkpoint \\
GPUs & 3 $\times$ NVIDIA A6000 GPUs \\
Memory & 49 GB\\
Precision & bfloat16 \\
Attention & FlashAttention-2 \\
Distributed training & DeepSpeed ZeRO-3 \\
Per-device batch size & 1 \\
Gradient accumulation & 4 \\
Effective global batch size & 12 \\
Learning rate & $2 \times 10^{-5}$ \\
Max sequence length & 8000 \\
\bottomrule
\end{tabular}
\end{table}

Table~\ref{tab:training_configs} summarizes the training configuration used for DeepArrhythmia. The total GPU hours required to train the model on MIT-BIH, MIT-BIH-SUP, INCART, and VitalDB were 61.30, 127.05, 57.72, and 179.98, respectively.

\section{Additional Method Details}
\label{app:method_details}

\subsection{Shift-based augmentation}
\label{app:shift_augmentation}
\begin{figure}[t]
  \centering
  
  \includegraphics[width=0.95\linewidth]{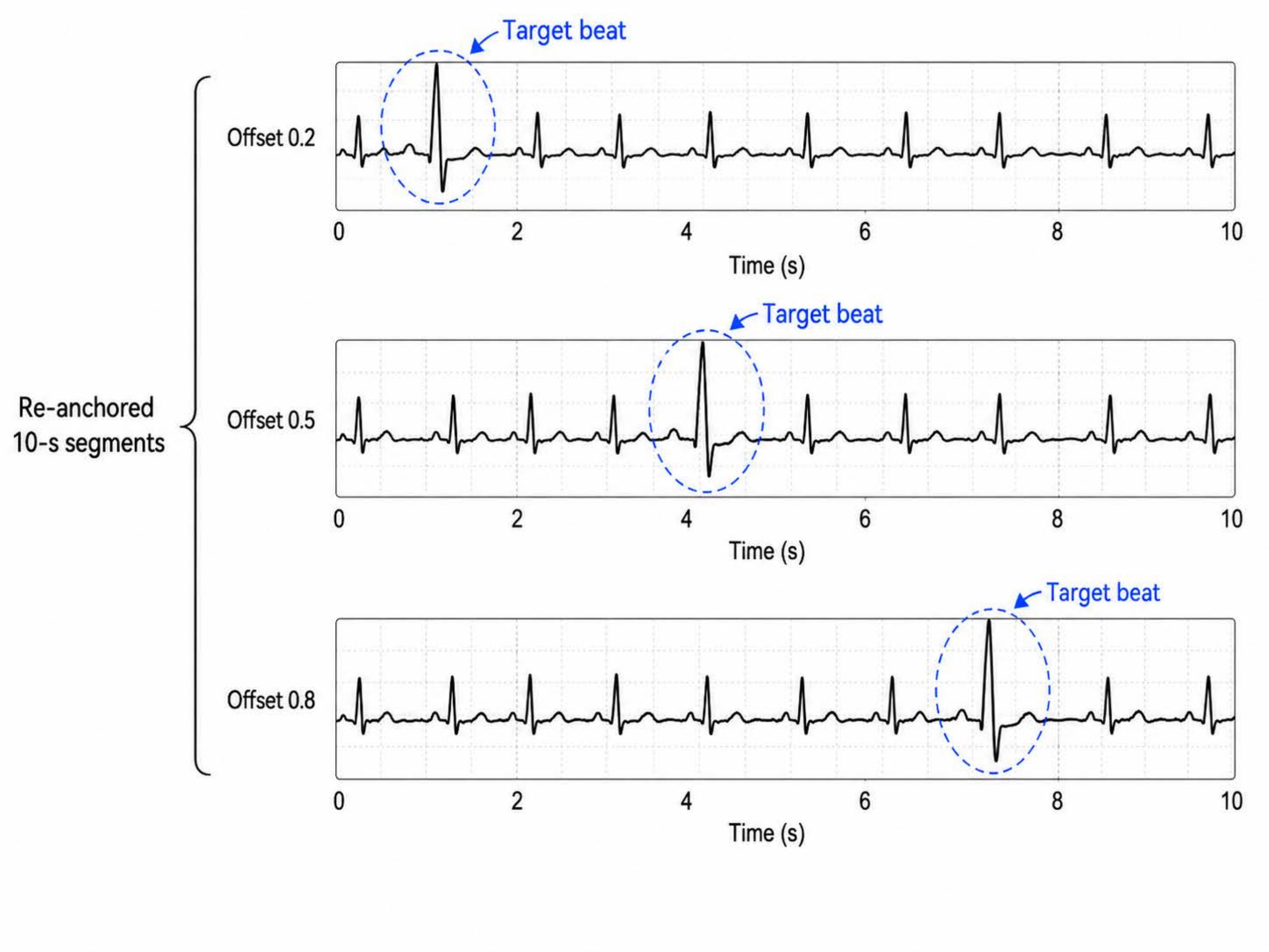}%
  
  \caption{Illustration of the shift-based augmentation strategy used to alleviate class imbalance. Rare classes are oversampled by placing their R-peaks at different positions within the segment, thereby increasing data diversity.}
  \label{fig:shift_augmentation}
\end{figure}

To mitigate class imbalance, we apply shift-based augmentation under the AAMI coarse class mapping as shown in Fig~\ref{fig:shift_augmentation}. We first compute class frequencies and define target relative abundances for non-normal classes with respect to the normal (N) class. For each non-N beat, we generate additional fixed-length (10 s) segments by re-anchoring the window so that the beat's R-peak appears at predefined fractional offsets along the segment axis. Rarer classes are assigned more distinct offsets to approach the target ratio, whereas classes already near the target are not oversampled. Candidate windows are clipped to valid record boundaries and deduplicated against the standard non-overlapping grid. During data loading, only augmentation files present on disk are ingested and merged with the base training split.

\subsection{Detailed tool definitions}
\label{app:tool_details}

\textbf{Peak Detector.}
To handle variable numbers of beats per segment, the first step is robust R-peak localization. We implement the peak detector using a 1D-UNet++ architecture \citep{zhou20211d}. Given an input segment $x_{ts}$, the detector outputs a point-wise probability sequence:
\begin{equation}
    p_t = f_{\theta}(x_{ts})_t, \qquad p_t \in [0,1], \quad t=1,\dots,L,
\end{equation}
where $f_{\theta}$ denotes the detector parameterized by $\theta$.

Training uses point-wise supervision at each time index. Let $y_t \in [0,1]$ denote the smoothed target label, obtained by applying a Gaussian kernel centered at each annotated peak to improve robustness. The optimization objective is point-wise binary cross-entropy:
\begin{equation}
    \mathcal{L}_{\mathrm{peak}} = -\frac{1}{L}\sum_{t=1}^{L}\left[y_t\log p_t + (1-y_t)\log(1-p_t)\right].
\end{equation}
At inference time, candidate peaks are obtained from the probability sequence using a classical peak-finding procedure \citep{pan2007real} with thresholding and non-maximum selection, yielding the final R-peak set $\mathcal{P}=\{\hat{t}_k\}_{k=1}^{K}$ for each segment.

\textbf{Feature Extractor.}
To complement learned representations with explicit physiological cues, we extract a set of handcrafted numerical features from each detected beat~\cite{hannun2019cardiologist}. Specifically, five feature groups are used: (i) R--R interval, (ii) normalized R--R interval, (iii) R-peak amplitude, (iv) higher-order statistics (HOS), and (v) numerical morphology descriptors.

For HOS, each beat is partitioned into five temporal sub-intervals, and both skewness and kurtosis are computed for each sub-interval, resulting in a 10-dimensional feature vector. Morphology descriptors are derived from Euclidean distances in the (sample, amplitude) space between the R-peak and four characteristic points selected from predefined local windows. These points are identified as follows: $\max(\mathrm{beat}[0,40])$, $\min(\mathrm{beat}[75,85])$, $\min(\mathrm{beat}[95,105])$, and $\max(\mathrm{beat}[150,180])$.

\textbf{Morphology Analyzer.}
To model beat morphology in an interpretable form, we employ a vision-language morphology analyzer that takes as input the segment ECG image together with beat-local auxiliary context. For a queried beat within the current segment, the student analyzer generates a textual morphology rationale:
\begin{equation}
    t_m = g_{\phi}\!\left(x_{\mathrm{img}}, c\right),
\end{equation}
where \(t_m\) denotes the generated morphology text, \(x_{\mathrm{img}}\) denotes the ECG image of the current segment, and \(c\) denotes the auxiliary context provided by the Feature Extractor for the queried beat.

We train $g_{\phi}$ with a knowledge-distillation framework using an online teacher LLM (e.g., Gemini 3.1). Given image input, ground-truth class label, and auxiliary context features (e.g., RR intervals and peak amplitude), the teacher produces reference explanations:
\begin{equation}
    \tilde{t}_{m} = g_{\psi}\!\left(x_{\mathrm{img}}, y, c\right),
\end{equation}
where $g_{\psi}$ denotes the teacher model, $y$ is the ground-truth label of the queried beat, and $c$ is the corresponding auxiliary context. The impact of teacher-model choice is analyzed in Appendix~\ref{app:teacher_model_effect}.

To prevent data leakage, teacher-generated rationales are produced exclusively for beats in the training partition; at inference time, the Morphology Analyzer operates without access to the ground-truth label.

The student is optimized with a token-level morphology distillation objective:
\begin{equation}
    \mathcal{L}_{\mathrm{morph}} = -\sum_{n=1}^{N} \log p_{\phi}\!\left(\tilde{t}_{m,n} \mid \tilde{t}_{m,<n}, x_{\mathrm{img}}, c\right),
\end{equation}
where $N$ is the target explanation length. The generated morphology text is passed to the central agent as structured evidence for final arrhythmia classification. In our experiments, Qwen3.5-4B serves as the base model for the student Morphology Analyzer.

\textbf{Confidence Calculator.}
The Confidence Calculator maps the current beat-level class posteriors to a single segment-level routing signal. For each detected beat, it takes the largest predicted class probability as that beat's confidence, and it then averages those beat-wise confidences across the segment to obtain $C_{\mathrm{seg}}$.
This scalar $C_{\mathrm{seg}}$ is the confidence statistic used to decide whether the additional evidence-producing tools should be invoked. Accordingly, routing is performed once per segment in the present study; per-beat routing was not explored here.
In the final agentic system, the Confidence Calculator is queried once at most per segment, immediately after peak detection and before any optional evidence call. 

\subsection{Detailed training objectives}
\label{app:training_objectives}
We introduce two modality placeholders, \texttt{<Signal>} and \texttt{<Image>}, to indicate insertion points for modality-specific tokens. Given time-series input $x_T$ and image input $x_I$, modality features are extracted and projected into the language-model embedding space using independent encoders and projectors:
\begin{align}
    e_T &= \mathrm{Encoder}_T(x_T; \theta_{E_T}), & z_T &= \mathrm{Proj}_T(e_T; \theta_{P_T}), \\
    e_I &= \mathrm{Encoder}_I(x_I; \theta_{E_I}), & z_I &= \mathrm{Proj}_I(e_I; \theta_{P_I}),
\end{align}
where $z_T \in \mathbb{R}^{N_T \times d}$ and $z_I \in \mathbb{R}^{N_I \times d}$ denote signal and image token sequences in a shared embedding space, and $\theta_M = \{\theta_{P_T}, \theta_{P_I}\}$ denotes projector parameters. The language model is conditioned on textual context $x_{\text{text}}$ together with $(z_T, z_I)$ by injecting $z_T$ at \texttt{<Signal>} and $z_I$ at \texttt{<Image>}.

We serialize each trajectory as a token sequence
\(\tau=(\tau_1,\ldots,\tau_T)\), containing optional \texttt{<Call>} spans,
externally returned \texttt{<Output>} spans, and the final peak-aligned textual
answer. The final answer is generated as ordinary language-model text, e.g.,
\texttt{[77:N] [370:N] [663:N] [947:N]}.

Let \(X=(x_{\mathrm{text}}, z_T, z_I)\) denote the multimodal conditioning
context, and let \(h_i\) denote the autoregressive language-model state before
generating token \(\tau_i\). The agent defines the standard next-token
probability \(p_{\theta}\!\left(\tau_i \mid X, \tau_{<i}\right)\) over the language-model vocabulary. 

Training uses masked next-token supervision over valid tool-action tokens and
final-answer tokens. The objective is written as the sum of tool-action and
class-answer token losses:
\begin{equation}
    \mathcal{L}_{\mathrm{agent}}
    =
    \mathcal{L}_{\mathrm{tool}}
    +
    \mathcal{L}_{\mathrm{cls}},
    \label{eq:agent_loss}
\end{equation}
with
\begin{equation}
    \mathcal{L}_{\mathrm{tool}}
    =
    -\sum_{i=1}^{T}
    m_i^{a}
    \log p_{\theta}\!\left(\tau_i^{*} \mid X, \tau_{<i}^{*}\right),
    \quad
    \mathcal{L}_{\mathrm{cls}}
    =
    -\sum_{i=1}^{T}
    m_i^{y}
    \log p_{\theta}\!\left(\tau_i^{*} \mid X, \tau_{<i}^{*}\right),
    \label{eq:agent_token_losses}
\end{equation}
where \(\tau^{*}=(\tau_1^{*},\ldots,\tau_T^{*})\) is the serialized
trajectory containing tool calls, tool outputs, and the final textual prediction.
The masks \(m_i^{a}\) and \(m_i^{y}\) select supervised tool-action tokens and
final-answer tokens, respectively. Externally provided tool-output tokens are
included as context and masked out from the loss.

\subsection{Threshold Selection}
\label{app:threshold}
We compute a segment-level confidence score for each example using the Confidence Calculator and choose a dataset-specific threshold that maximizes the micro F1 score for beat classification for Stage~2 SFT set. This threshold is fixed at inference time. The resulting dataset-specific thresholds are reported in Table~\ref{tab:balanced_f1_thresholds}.
\begin{table}[htbp]
\centering
\caption{Micro-F1 confidence thresholds for tool-use decision}
\label{tab:balanced_f1_thresholds}
\begin{tabular}{lcccc}
\hline
Dataset & MIT-BIH Arrhythmia & MIT-BIH Supraventricular & INCART & VitalDB \\
\hline
Threshold & 0.990529 & 0.980933 & 0.993532 & 0.98519 \\
\hline
\end{tabular}
\end{table}

\begin{figure*}[t]
  \centering
  \begin{subfigure}{0.48\textwidth}
    \centering
    \includegraphics[width=\linewidth]{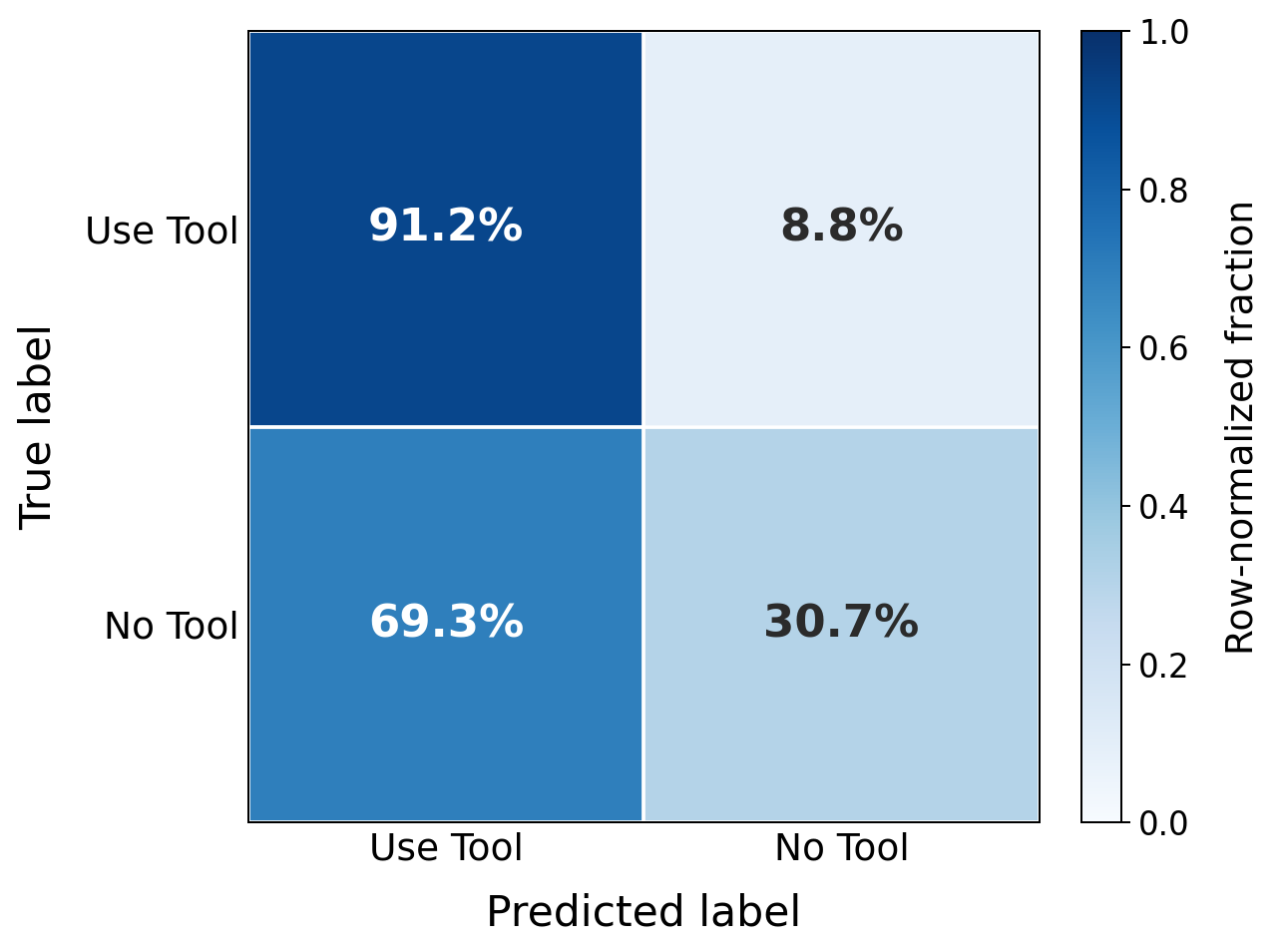}
    \caption{Confusion matrix on MIT-BIH Arrhythmia.}
    \label{fig:tool_use_mitbih}
  \end{subfigure}
  \hfill
  \begin{subfigure}{0.48\textwidth}
    \centering
    \includegraphics[width=\linewidth]{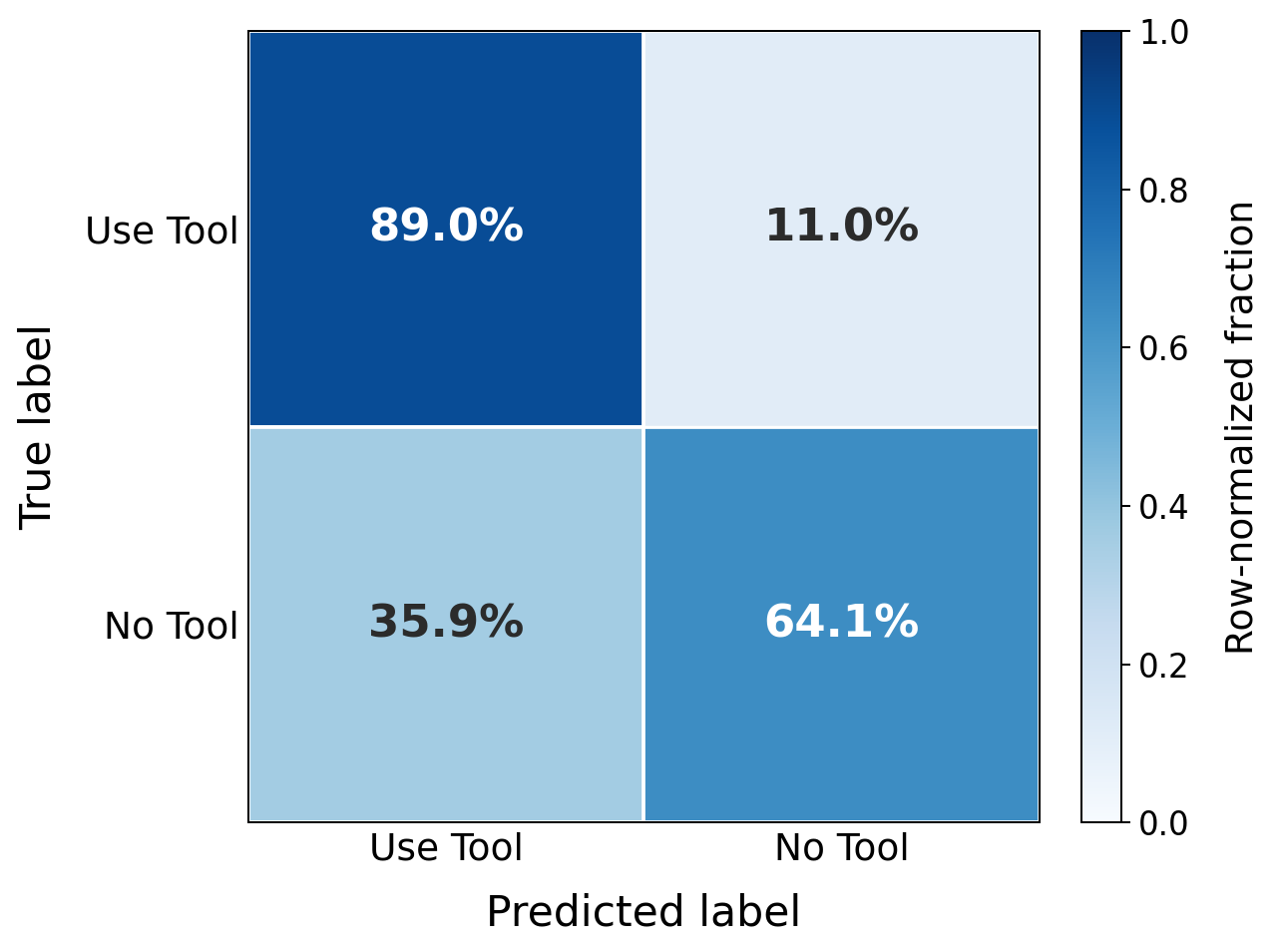}
    \caption{Confusion matrix on MIT-BIH Supraventricular.}
    \label{fig:tool_use_mitbih_super}
  \end{subfigure}

  \vspace{0.5em}

  \begin{subfigure}{0.48\textwidth}
    \centering
    \includegraphics[width=\linewidth]{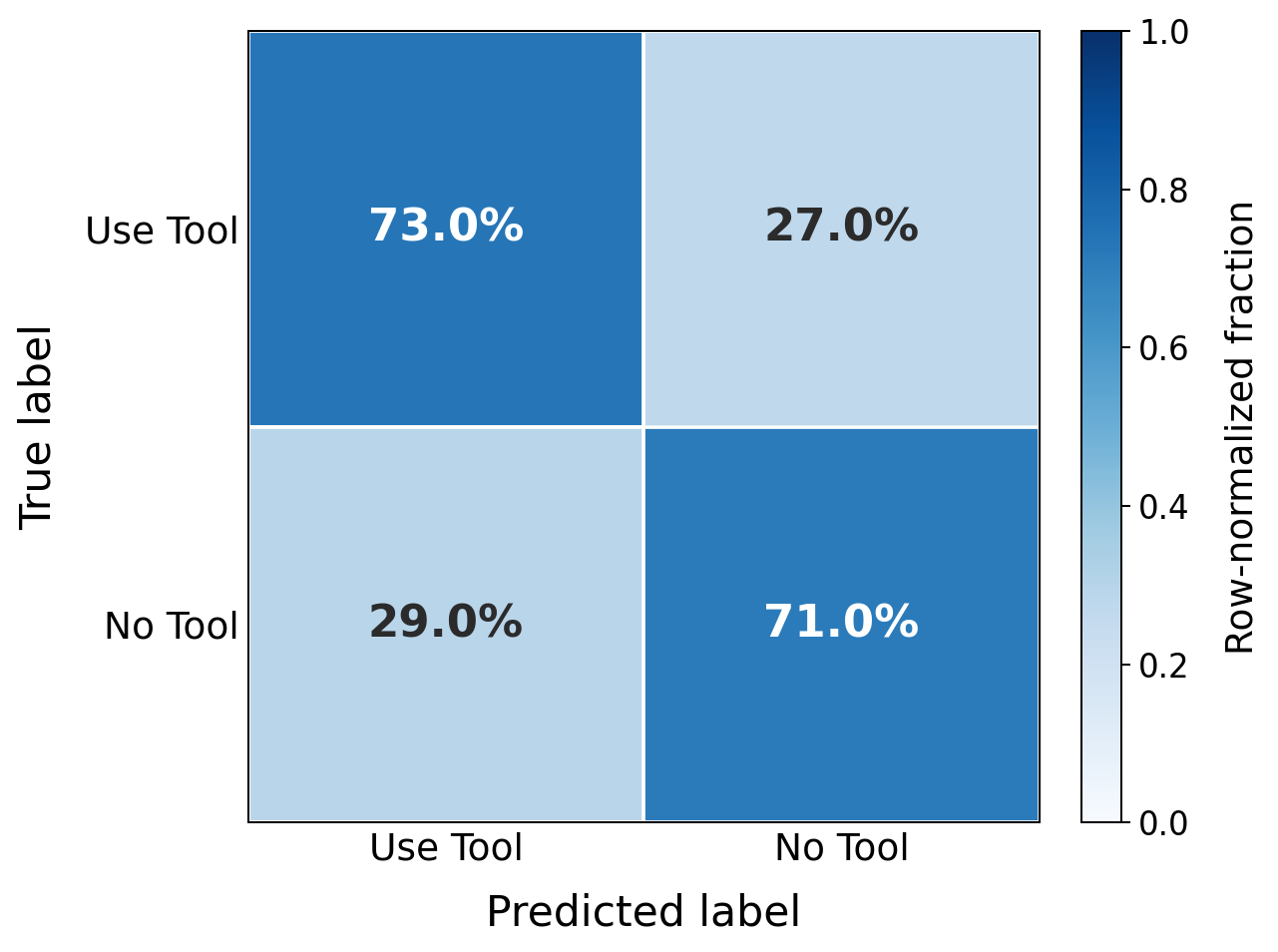}
    \caption{Confusion matrix on INCART.}
    \label{fig:tool_use_incart}
  \end{subfigure}
  \hfill
  \begin{subfigure}{0.48\textwidth}
    \centering
    \includegraphics[width=\linewidth]{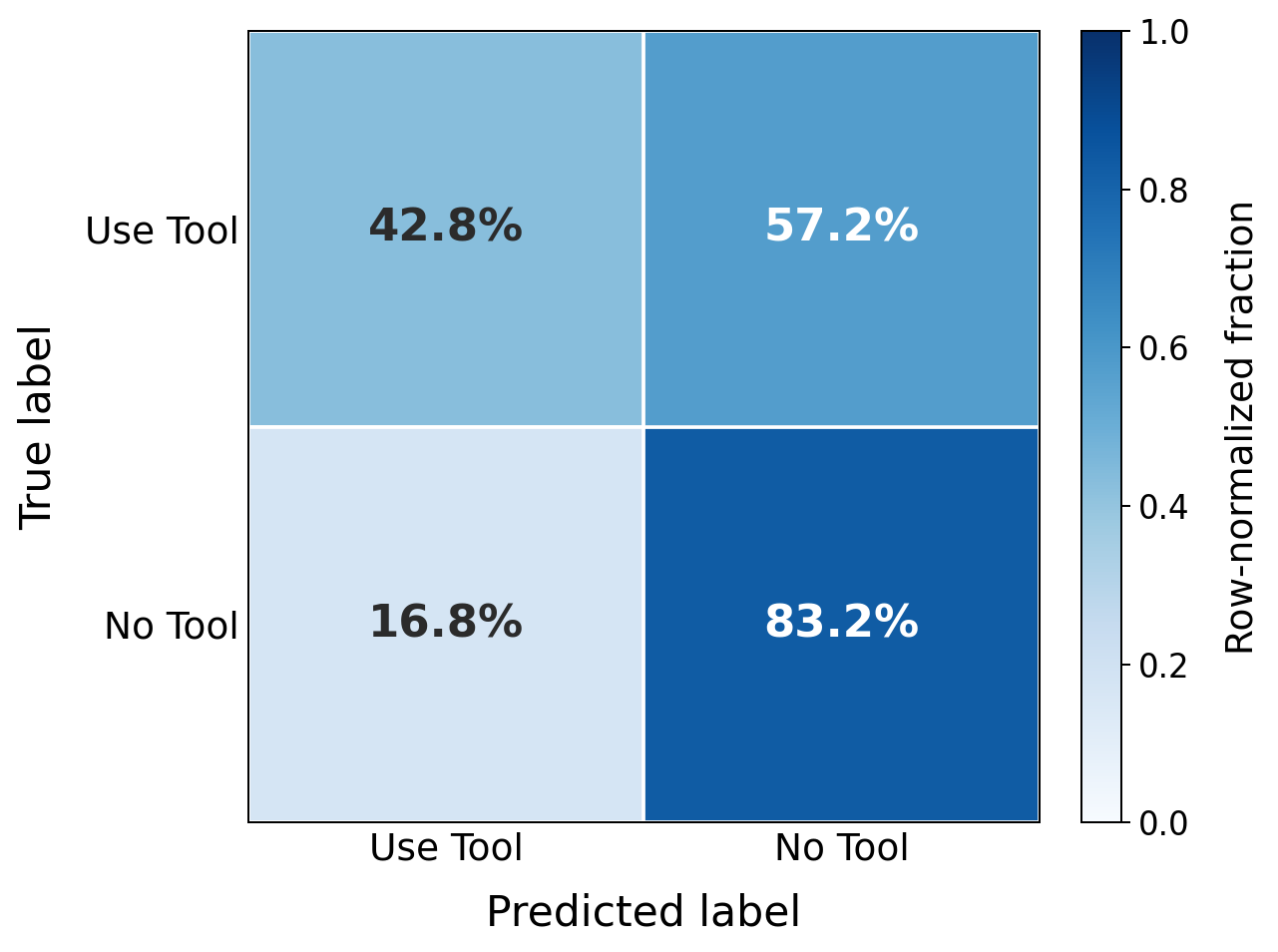}
    \caption{Confusion matrix on VitalDB.}
    \label{fig:tool_use_vitaldb}
  \end{subfigure}

  \caption{Confusion-matrix panels for tool use across four datasets.}
  \label{fig:tool use confusion matrix}
\end{figure*}

Figure~\ref{fig:tool use confusion matrix} presents the row-normalized confusion matrices for the tool-use decision across datasets. On MIT-BIH Arrhythmia, MIT-BIH superventricular, and Incart, the classifier exhibits a strong tendency to invoke tools.
In contrast, VitalDB exhibits a clear shift toward the no-tool prediction: 83.2\% of true no-tool cases are correctly classified, whereas only 42.8\% of true tool-use cases are identified. This pattern suggests a two-part limitation on VitalDB: first, the short-branch confidence signal used for routing is less effective at separating samples that should remain on the no-tool branch from those that should escalate to the richer branch; second, even after invocation, the optional evidence appears less useful on this dataset, as further discussed in Section~\ref{Appendix:VitalDB Result Analysis}.

\section{Baseline Augmentation}
\label{app:baseline_augmentation}

To assess whether standard imbalance-mitigation techniques can strengthen a conventional baseline, we evaluate the TCN model under three training settings: the original non-augmented baseline, class-weighted loss~\cite{xu2020class}, and class-balanced sampling augmentation~\cite{rendon2020data}. Table~\ref{tab:baseline_augmentation_tcn} summarizes the resulting Macro-F1 and Micro-F1 scores across the four datasets. Class-weighted loss yields the most consistent improvement, outperforming the non-augmented baseline on both metrics for all datasets. By contrast, class-balanced sampling produces mixed effects: it improves Macro-F1 on INCART and MIT-BIH-SUP, but reduces Micro-F1 on three of the four datasets. Despite these modest gains, all augmented TCN variants remain substantially below DeepArrhythmia. Overall, these results indicate that standard imbalance-aware training offers only limited improvements for the TCN baseline.

\begin{table*}[t]
  \centering
  \caption{TCN baseline performance under alternative imbalance-mitigation strategies. We report Macro-F1 and Micro-F1 on the standard test split of each dataset.}
  \label{tab:baseline_augmentation_tcn}
  \footnotesize
  \resizebox{\textwidth}{!}{%
    \begin{tabular}{lcccccccc}
      \toprule
      & \multicolumn{2}{c}{MIT-BIH} & \multicolumn{2}{c}{INCART} & \multicolumn{2}{c}{MIT-BIH-SUP} & \multicolumn{2}{c}{VitalDB} \\
      \cmidrule(r){2-3} \cmidrule(r){4-5} \cmidrule(r){6-7} \cmidrule(r){8-9}
      \textbf{Training setting} & \textbf{Ma} & \textbf{Mi} & \textbf{Ma} & \textbf{Mi} & \textbf{Ma} & \textbf{Mi} & \textbf{Ma} & \textbf{Mi} \\
      \midrule
      Non-augmented baseline & 0.4679 & 0.7048 & 0.4403 & 0.8249 & 0.4336 & 0.7796 & 0.5069 & 0.8114 \\
      Class-weighted loss & \textbf{0.4796} & \textbf{0.7370} & 0.4705 & \textbf{0.8500} & 0.4597 & \textbf{0.8257} & \textbf{0.5223} & \textbf{0.8149} \\
      Class-balanced sampling augmentation & 0.4446 & 0.6708 & \textbf{0.4712} & 0.8285 & \textbf{0.4680} & 0.8208 & 0.4895 & 0.7850 \\
      \bottomrule
    \end{tabular}%
  }
\end{table*}

\section{VitalDB Result Analysis}
\label{Appendix:VitalDB Result Analysis}

The VitalDB performance becomes worse after adding the auxiliary ECG-classification evidence. To analyze whether this evidence is intrinsically useful, we evaluate the discriminative power of the evidence features alone.

For each beat, we parse the auxiliary feature vector
\begin{equation}
\mathbf{x}_i =
[
\mathrm{RR}_i,
\mathrm{normRR}_i,
\mathrm{amp}_i,
\mathrm{HOS}_i,
\mathrm{myMorph}_i
]
\in \mathbb{R}^{23},
\end{equation}
from the evidence text. We then train a balanced logistic regression classifier using only $\mathbf{x}_i$, without ECG images or language-model predictions. 

\begin{table}[ht]
\centering
\caption{Feature-only classification performance.}
\label{tab:feature_only_separability}
\begin{tabular}{lccc}
\toprule
Dataset & Accuracy & Macro-F1 & Weighted-F1 \\
\midrule
MIT-BIH & 0.7080 & 0.5209 & 0.7895 \\
VitalDB & 0.4879 & 0.3851 & 0.5315 \\
\bottomrule
\end{tabular}
\end{table}

The results in Table~\ref{tab:feature_only_separability} show a clear dataset-level gap in feature-only separability. On MIT-BIH, the handcrafted evidence features alone achieve 0.7080 accuracy, 0.5209 Macro-F1, and 0.7895 Weighted-F1, indicating that they retain meaningful class information even without ECG images or language-model predictions. On VitalDB, the same feature-only classifier drops to 0.4879 accuracy, 0.3851 Macro-F1, and 0.5315 Weighted-F1, showing that these auxiliary features are substantially less discriminative on that dataset.

This gap helps explain the downstream behavior. When the auxiliary evidence is relatively informative, as on MIT-BIH, conditioning on it can support the final prediction. When the evidence is weak, as on VitalDB, adding it can inject noise rather than useful information, which may degrade final performance.

\section{Routing Scheme Analysis}
\label{app:Routing Scheme Analysis}
We further analyze the routing mechanism on the MIT-BIH Arrhythmia test set by comparing two segment-level confidence statistics for optional tool invocation: the mean beat-wise confidence within a segment and the minimum beat-wise confidence within a segment. The mean-based statistic reflects the overall certainty of the segment, whereas the minimum-based statistic emphasizes the most uncertain beat. Table~\ref{tab:routing_scheme_comparison} reports the confidence threshold for each routing rule, the resulting Macro-F1 and Micro-F1, and the number of segments assigned to the no-tool and tool-enabled branches.

Both routing rules achieve the same Macro-F1 of 0.588, indicating comparable class-balanced performance. However, the mean-confidence rule attains a slightly higher Micro-F1 (0.951 vs.~0.948) while routing slightly fewer segments to the tool-enabled branch (2,954 vs.~2,982) and slightly more segments to the no-tool branch (1,244 vs.~1,216). By contrast, the minimum-confidence rule is more aggressive, because a single low-confidence beat can send the entire segment to the tool-enabled branch. %

\begin{table}[t]
\centering
\caption{Comparison of segment-level routing rules on the MIT-BIH Arrhythmia test set. Confidence denotes the tuned threshold used by each routing rule. The No tool and Use tool columns report the number of evaluation segments assigned to the no-tool and tool-enabled branches, respectively.}
\label{tab:routing_scheme_comparison}
\small
\begin{tabular}{lccccc}
\toprule
Routing signal & Confidence & Macro-F1 & Micro-F1 & No tool & Use tool \\
\midrule
Mean confidence & 0.990529 & 0.588 & 0.951 & 1,244 & 2,954 \\
Minimum confidence & 0.990214 & 0.588 & 0.948 & 1,216 & 2,982 \\
\bottomrule
\end{tabular}
\end{table}

\section{Sensitivity to the Routing Threshold}
\label{app:threshold-sensitivity}

\begin{figure}[t]
\centering
\begin{subfigure}[t]{0.48\textwidth}
  \includegraphics[width=\textwidth]{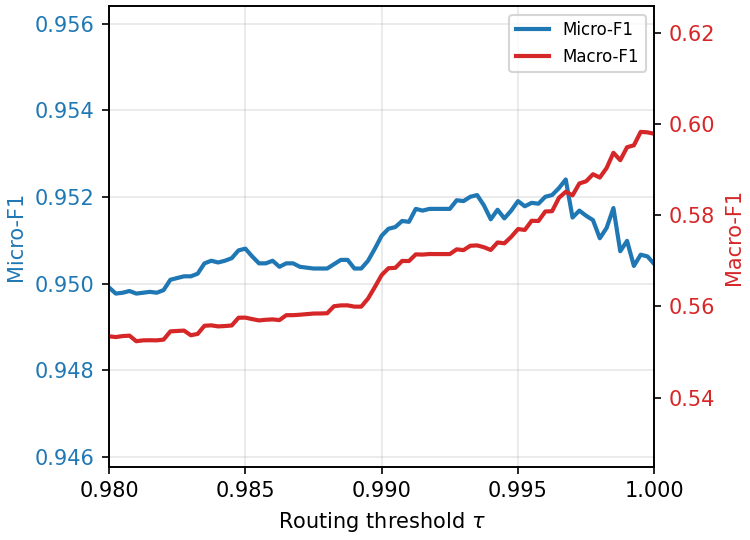}
  \caption{MIT-BIH}
\end{subfigure}\hfill
\begin{subfigure}[t]{0.48\textwidth}
  \includegraphics[width=\textwidth]{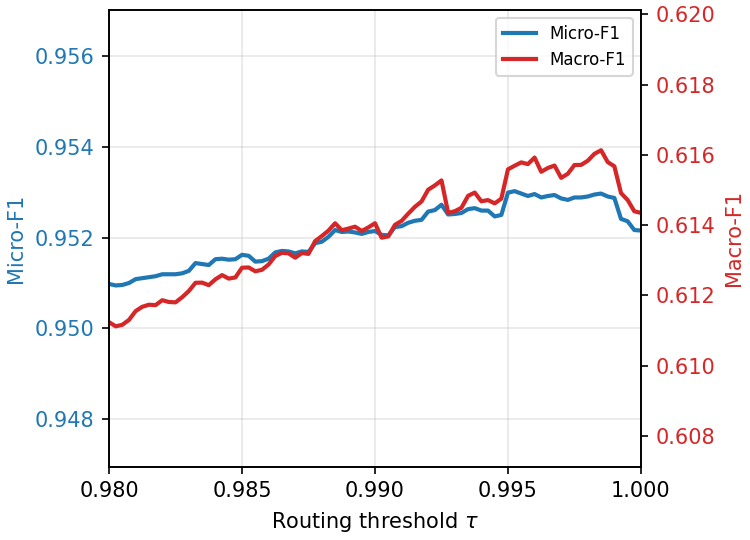}
  \caption{MIT-BIH Supraventricular}
\end{subfigure}
\vspace{0.6em}
\begin{subfigure}[t]{0.48\textwidth}
  \includegraphics[width=\textwidth]{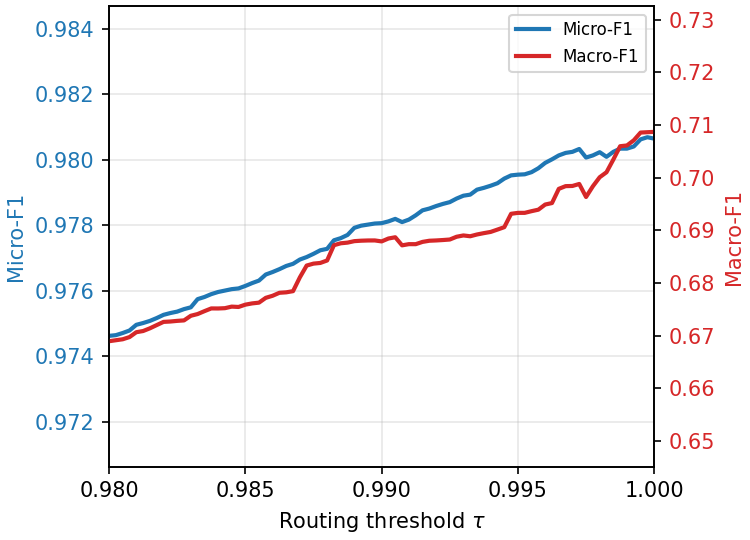}
  \caption{INCART}
\end{subfigure}\hfill
\begin{subfigure}[t]{0.48\textwidth}
  \includegraphics[width=\textwidth]{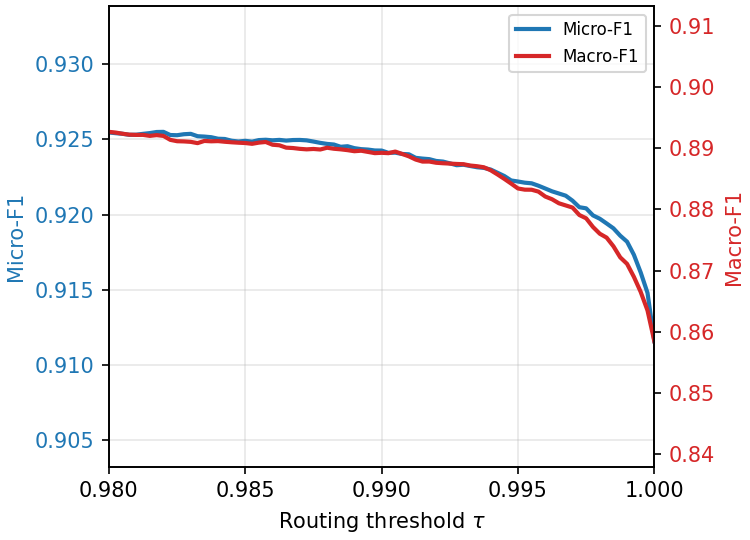}
  \caption{VitalDB}
\end{subfigure}
\caption{Beat-level Micro-F1 (left axis, blue) and Macro-F1 (right axis, red)
versus the routing threshold $\tau$ over the operative range $[0.98,1.0]$. Performance is flat in $\tau$ on
all datasets (Micro-F1 varies by $\leq 1.4$ points), and the effect of rich
evidence is dataset-dependent: monotonically beneficial on INCART, detrimental on
VitalDB, and maximized at an interior threshold on MIT-BIH and MIT-BIH
Supraventricular, where selective routing slightly surpasses always acquiring rich
evidence.}
\label{fig:threshold-sensitivity}
\end{figure}

Because the per-dataset threshold $\tau_d$ is selected on a held-out split, a
reasonable concern is that the routed model is brittle to the precise value of
$\tau_d$ and may not transfer across acquisition conditions. We evaluate this
directly by characterizing performance as a continuous function of the threshold.
For a given $\tau$, the policy of Section~\ref{sec:method} emits the
minimal-evidence prediction on segments whose confidence satisfies
$C_{\text{seg}}\!\geq\!\tau$ and the rich-evidence prediction otherwise; sweeping
$\tau$ thus interpolates the entire family of policies between the pure
minimal-evidence model ($\tau\!\to\!0$) and the pure rich-evidence model
($\tau\!\to\!1$). As
$C_{\text{seg}}$ concentrates near unity---the minimal-evidence model is
confidently correct on the dominant normal class---all non-trivial routing occurs
within $\tau\!\in\![0.98,1.0]$, which we adopt as the operative range in
Figure~\ref{fig:threshold-sensitivity}.

Performance is stable across this range on all four datasets. Micro-F1 spans only
$0.9498$--$0.9524$ (MIT-BIH), $0.9509$--$0.9530$ (MIT-BIH Supraventricular),
$0.9746$--$0.9807$ (INCART), and $0.9116$--$0.9255$ (VitalDB), a maximum variation
of $1.4$ points with no discontinuity. A mis-specified threshold therefore
degrades accuracy by at most a few tenths of a point rather than inducing a
collapse, indicating that $\tau_d$ need not be finely tuned to remain effective.

The value of acquiring rich evidence nonetheless varies markedly across
datasets, and the optimal operating point is not consistently at either extreme.
On INCART, rich evidence is monotonically beneficial and performance is maximized
as $\tau\!\to\!1$. On VitalDB the trend is reversed: both Micro-F1 and Macro-F1
peak near the minimal-evidence end ($0.9255$ and $0.8927$ at $\tau\!\approx\!0.98$)
and decline monotonically toward $\tau\!=\!1$, so indiscriminate acquisition is
actively harmful, consistent with the noisier VitalDB features analyzed in
Appendix~\ref{Appendix:VitalDB Result Analysis}. On MIT-BIH and MIT-BIH Supraventricular the Micro-F1
curve is non-monotonic and attains an interior optimum ($0.9524$ at
$\tau\!\approx\!0.997$ and $0.9530$ at $\tau\!\approx\!0.995$, respectively) that
marginally exceeds the always-rich endpoint ($0.9504$ and $0.9522$); withholding
rich evidence on the most confident segments is thus not merely cheaper but
slightly more accurate than acquiring it everywhere.

Taken together, these results show that the utility of physiological evidence is
heterogeneous---beneficial on some datasets, detrimental on others, and best
applied selectively on the rest---so no fixed acquisition policy is uniformly
optimal. Confidence-gated routing accommodates this heterogeneity automatically,
keeping the model on the favorable side of the accuracy--cost trade-off on every
dataset while remaining robust to the exact choice of threshold.

\paragraph{Calibration of the routing score.} The confidence of a beat is $c=\max_y p_y$ over the posterior on $\{\mathrm{N},\mathrm{S},\mathrm{V},\mathrm{F}\}$, and the routing score $C_{\text{seg}}$ is the mean beat confidence within a segment. On the minimal-evidence predictions of each subject-disjoint test split we bin all beats into $M=10$ equal-width confidence bins and report $\mathrm{ECE}=\sum_{m}\frac{|B_m|}{N}\,|\mathrm{acc}(B_m)-\mathrm{conf}(B_m)|$ and $\mathrm{MCE}=\max_{m:|B_m|>0}|\mathrm{acc}(B_m)-\mathrm{conf}(B_m)|$ (Table~\ref{tab:calibration}). ECE is low throughout ($0.024$--$0.058$): the reliability curve tracks the diagonal at the high-confidence end, where the large majority of beats lie. The larger MCE ($0.117$--$0.164$) is driven by sparsely populated mid-confidence bins with mild over-confidence, which is benign for routing because the policy only requires the score to \emph{order} segments by difficulty. Under cross-dataset shift calibration is not guaranteed to transfer, which is why $\tau_d$ is re-selected per dataset. The selected thresholds all fall in the narrow band $\tau_d\in[0.9809,0.9935]$ (Table~\ref{tab:calibration}), inside the swept range, so substituting another dataset's threshold changes Micro-F1 by at most the span of the sweep ($0.3$, $0.2$, $0.6$, and $1.4$ points on MIT-BIH, MIT-BIH Supraventricular, INCART, and VitalDB). A genuinely new site could shift the confidence distribution itself; label-free alternatives such as fixing $\tau$ at a confidence percentile of the incoming data are left to future work.

\paragraph{Alternative routers.} We compare the threshold router with a stochastic entropy router and a learned (logistic-regression) router. The simple threshold matches or exceeds both (Macro-F1 on MIT-BIH $0.593$ vs.\ $0.592$/$0.586$; INCART $0.629$ vs.\ $0.628$/$0.622$; VitalDB $0.892$ vs.\ $0.886$/$0.894$ for threshold vs.\ entropy/learned).

\paragraph{Routing rates.} Table~\ref{tab:routing_rates} reports how often the rich-evidence path is selected under each dataset's threshold ($C_{\text{seg}}<\tau_d$). Arrhythmia-containing segments trigger the rich path $2.2$--$5.3\times$ more often than normal-only segments. Rich routing concentrates in the lowest-confidence quartile, where the minimal-evidence model is least accurate (accuracy $0.81$--$0.91$ vs.\ $0.97$--$1.00$ elsewhere), and it is lowest on VitalDB, where rich evidence is least helpful. Because noise lowers confidence, degraded segments are routed to rich evidence more often, so the mechanism fails safe (Appendix~\ref{app: Impact of Noise}).

\begin{table}[t]
\centering
\caption{Calibration of the minimal-evidence confidence (10 equal-width bins) and the selected thresholds $\tau_d$.}
\label{tab:calibration}
\small
\begin{tabular}{lccc}
\toprule
Dataset & ECE & MCE & $\tau_d$ \\
\midrule
MIT-BIH & 0.058 & 0.121 & 0.9905 \\
MIT-BIH Supraventricular & 0.029 & 0.117 & 0.9809 \\
INCART & 0.024 & 0.155 & 0.9935 \\
VitalDB & 0.042 & 0.164 & 0.9852 \\
\bottomrule
\end{tabular}
\end{table}

\begin{table}[t]
\centering
\caption{Fraction of segments routed to the rich-evidence path, overall and by segment type.}
\label{tab:routing_rates}
\small
\begin{tabular}{lccc}
\toprule
Dataset & Rich (all) & Normal-only & Abnormal \\
\midrule
MIT-BIH & 47.4\% & 32.2\% & 72.0\% \\
MIT-BIH Supraventricular & 43.4\% & 17.3\% & 70.0\% \\
INCART & 31.3\% & 9.1\% & 48.1\% \\
VitalDB & 17.1\% & 8.2\% & 27.4\% \\
\bottomrule
\end{tabular}
\end{table}

\section{Dataset Class Distribution}
\label{app:dataset_class_distribution}

\begin{table}[t]
  \centering
  \caption{Beat-level class distribution across datasets (datasets shown as columns).}
  \label{tab:dataset_class_distribution}
  \small
  \setlength{\tabcolsep}{5pt}
  \begin{tabular}{lrrrr}
    \toprule
    Metric & MIT-BIH & INCART & MIT-BIH SV & VitalDB \\
    \midrule
    Records & 48 & 75 & 78 & 482 \\
    Total beats & 109,494 & 175,874 & 184,582 & 640,107 \\
    N (\%) & 90,631 (82.77) & 153,676 (87.38) & 162,339 (87.95) & 436,503 (68.19) \\
    S (\%) & 2,781 (2.54) & 1,960 (1.11) & 12,198 (6.61) & 183,052 (28.60) \\
    V (\%) & 7,236 (6.61) & 20,013 (11.38) & 9,943 (5.39) & 16,097 (2.51) \\
    F (\%) & 803 (0.73) & 219 (0.12) & 23 (0.01) & 0 (0.00) \\
    Q (\%) & 8,043 (7.35) & 6 ($<0.01$) & 79 (0.04) & 4,455 (0.70) \\
    \bottomrule
  \end{tabular}
\end{table}

Table~\ref{tab:dataset_class_distribution} shows that all four datasets are strongly imbalanced, but the degree and structure of imbalance vary substantially across datasets. Normal beats dominate every dataset, accounting for 68.19\% to 87.95\% of all annotations, which means that aggregate metrics can be heavily influenced by the majority class. MIT-BIH and INCART are both dominated by normal beats, but INCART contains a noticeably larger ventricular proportion (11.38\%), whereas MIT-BIH includes a relatively larger fraction of unclassifiable beats (Q, 7.35\%). MIT-BIH Superventricular remains normal-heavy overall, yet contains the highest proportion of supraventricular beats among the three conventional ECG benchmarks (6.61\%), making it useful for evaluating supraventricular discrimination. VitalDB exhibits the most distinct distribution: compared with the other datasets, it contains a much lower normal-beat proportion and a substantially higher supraventricular proportion (28.60\%), while ventricular beats remain comparatively rare. Fusion beats are extremely scarce in every dataset and entirely absent in VitalDB, which helps explain the instability of class-F evaluation. 

\section{Impact of data splitting}
\label{app:impact of data spliting}
\begin{figure*}[t]
  \centering
  \begin{subfigure}{0.48\textwidth}
    \centering
    \includegraphics[width=\linewidth]{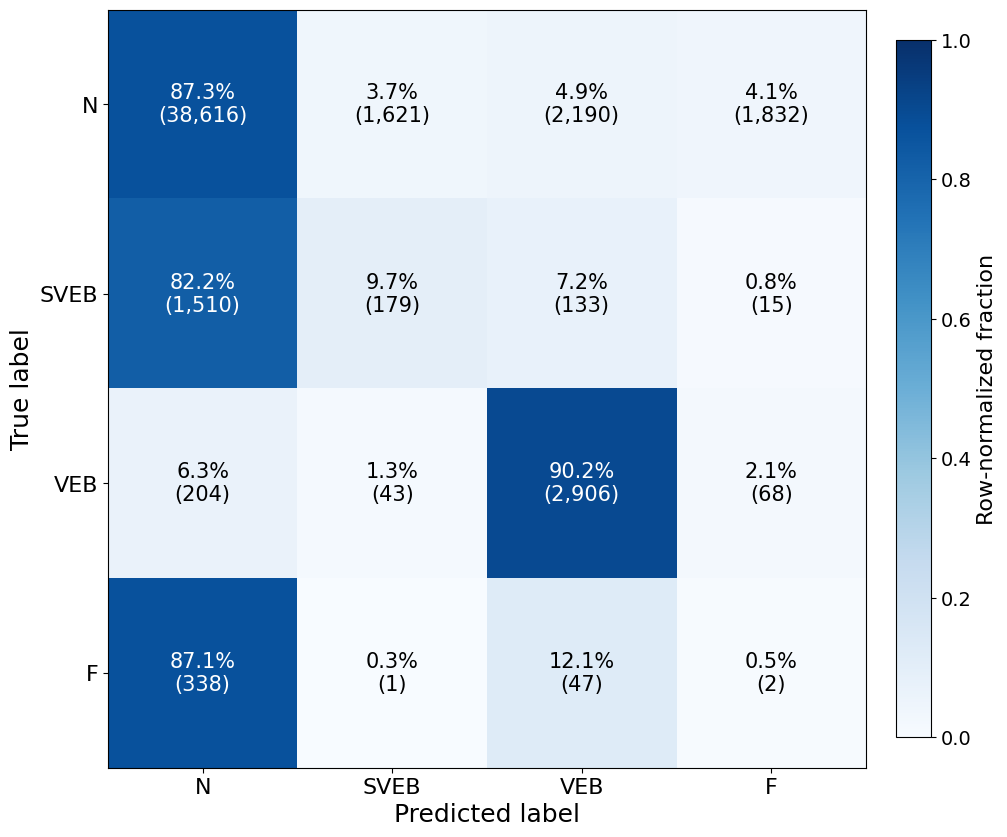}
    \caption{Subject-disjoint DS1/DS2 split on MIT-BIH.}
    \label{fig:mitbih-ds1-ds2}
  \end{subfigure}
  \hfill
  \begin{subfigure}{0.48\textwidth}
    \centering
    \includegraphics[width=\linewidth]{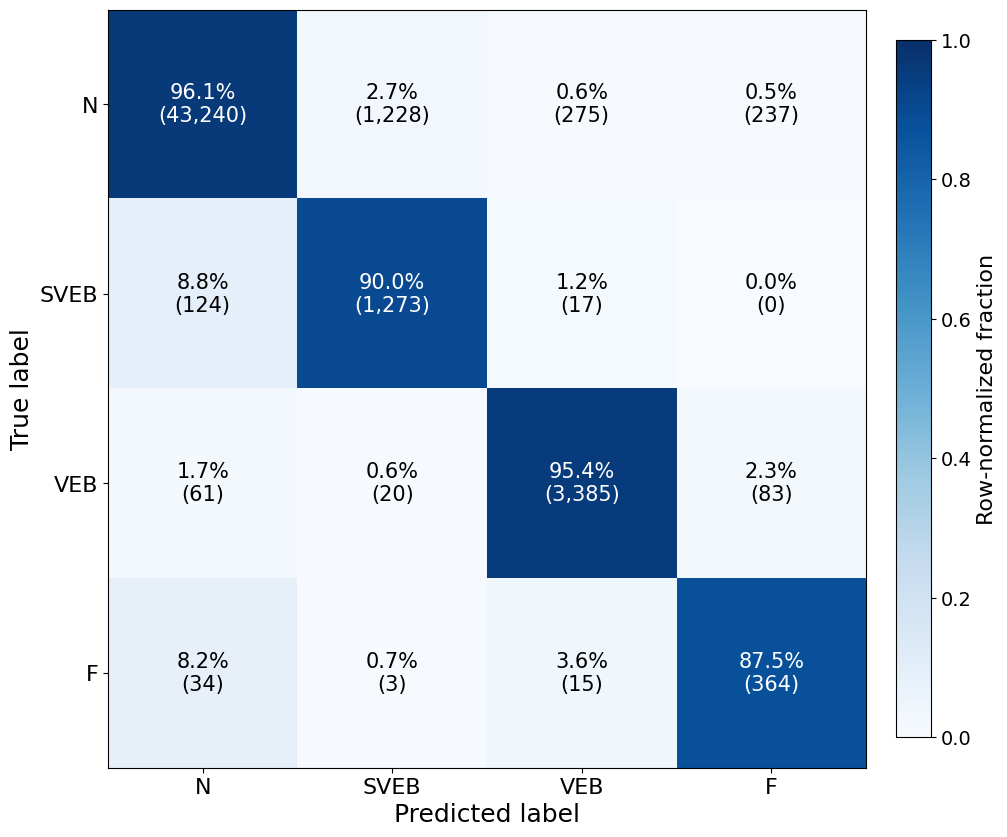}
    \caption{Random split on MIT-BIH.}
    \label{fig:mitbih-random-split}
  \end{subfigure}
  \caption{Comparison of MIT-BIH evaluation outcomes under DS1/DS2 and random split protocols.}
  \label{fig:split-visual-comparison-mitbih}
\end{figure*}

Prior ECG studies often report strong performance on MIT-BIH \citep{limam2017atrial, huang2019ecg}; however, reported results are highly sensitive to the data-partition protocol. In this work, we adopt the subject-disjoint DS1/DS2 split, which prevents patient overlap between training and test sets and therefore mitigates identity leakage. By contrast, random beat-level splitting can assign beats from the same subject to both sets, yielding an easier but less clinically realistic evaluation scenario. Figure~\ref{fig:split-visual-comparison-mitbih} presents the confusion matrices obtained under the two evaluation protocols for the 2D CNN baseline~\cite{limam2017atrial}. The random split produces markedly higher apparent performance, increasing SVEB recall from 9.7\% to 90.0\% and fusion-beat (F) recall from 0.5\% to 87.5\%. This large discrepancy indicates substantial leakage risk under random splitting and supports the use of subject-disjoint evaluation for reliable clinical assessment.

\section{Accuracy Distribution by Prediction Confidence}
\label{app:confidence_distribution}
\begin{figure*}[t]
  \centering
  \includegraphics[width=0.9\textwidth]{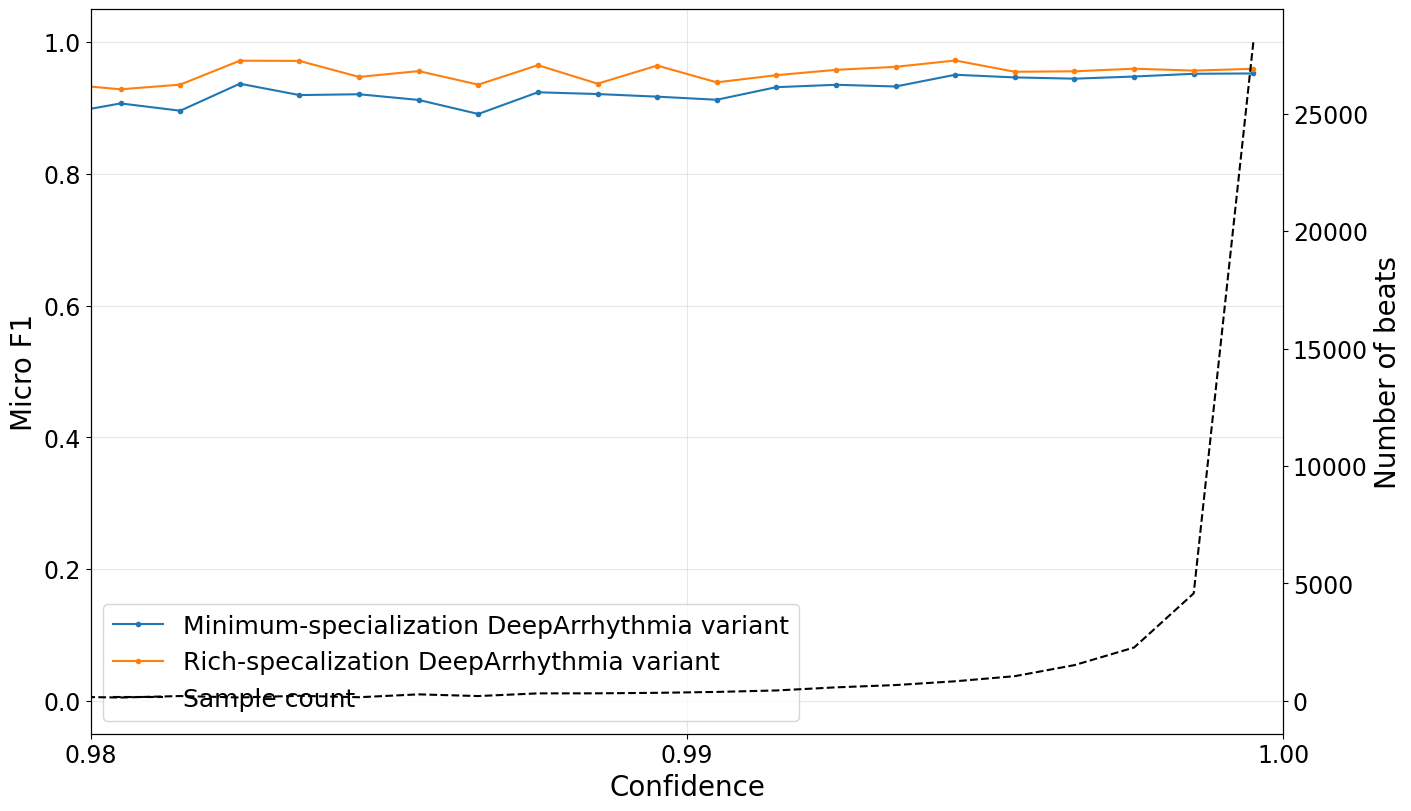}
  \caption{Relationship between prediction confidence and segment-level Micro-F1 on the MIT-BIH dataset. The x-axis shows the confidence score assigned to each segment by the minimal-specialization DeepArrhythmia variant, whereas the y-axis reports segment-level Micro-F1 for both the minimal- and rich-specialization DeepArrhythmia variants. We additionally visualize the number of samples at each confidence level to characterize the distribution of evaluation segments.}
  \label{fig:confidence-vs-f1}
\end{figure*}

Figure~\ref{fig:confidence-vs-f1} illustrates the relationship between segment-level prediction confidence and Micro-F1 on the MIT-BIH dataset.  Most segments cluster at very high confidence values (above 0.99), indicating that a substantial portion of the test set is relatively easy for the lightweight baseline. By contrast, the gains from routing are concentrated in the low-confidence regime, and the magnitude of improvement increases as confidence decreases. This pattern suggests that low-confidence segments correspond to genuinely difficult cases and that richer evidence improves accuracy precisely where the baseline is least reliable. Overall, these results support the premise that a confidence-gated routing strategy achieves a favorable accuracy--efficiency trade-off by selectively invoking additional evidence only for challenging inputs.

\section{Cross Dataset Performance}
\label{app:cross_dataset}

Table~\ref{tab:cross-dataset-performance} reports cross-dataset generalization, with rows denoting the training dataset and columns denoting the evaluation dataset. For computational efficiency, we evaluate each source--target pair on a 10\% sample (fixed seed) of the corresponding target test set for DeepArrhythmia with threshold-induced routing. Nothing from the target dataset is used for training, tuning, normalization, or threshold selection: DeepArrhythmia keeps its per-dataset checkpoint and routing policy, the 1D ResNet uses the source's normalization statistics, and the SVM uses its source-fitted scaler. In-domain performance is highest for all four target datasets, with matched train--test Micro-F1 values of 0.9521 on MITBIH, 0.9527 on Superventricular, 0.9791 on INCART, and 0.9023 on VitalDB. Cross-dataset transfer is nevertheless highly uneven. Models trained on Superventricular and INCART generalize comparatively well to MITBIH and to each other, whereas the VitalDB-trained model transfers poorly overall, most notably to Superventricular (0.0935). MITBIH-trained models transfer reasonably to Superventricular (0.9272) but degrade substantially on INCART and VitalDB (0.5540 and 0.5514, respectively). Overall, these results indicate substantial dataset shift and pronounced source--target asymmetry despite strong in-domain accuracy.

To assess whether this behavior is specific to our framework or an intrinsic
property of the task, we run the identical cross-dataset protocol on two
representative baselines: the raw-signal 1D ResNet (Table~\ref{tab:cross-dataset-resnet}) and the feature-based SVM \ (Table~\ref{tab:cross-dataset-svm}). Both baselines
show the same source--target asymmetry, confirming that the shift is driven by
dataset factors (sampling rate, lead configuration, acquisition device, and class
priors) rather than by any single model. Crucially, when trained on informative
source datasets such as Superventricular and INCART, DeepArrhythmia transfers to
unseen target datasets more effectively than either baseline. For the
Superventricular-trained model, DeepArrhythmia attains a Micro-F1 of 0.9351 on
MITBIH and 0.8711 on INCART, compared with 0.0743 and 0.4786 for the 1D ResNet
and 0.8827 and 0.8624 for the SVM. The INCART-trained model shows the same
advantage, reaching 0.9243 on MITBIH and 0.8942 on Superventricular, versus 0.4199
and 0.6787 for the ResNet and 0.8825 and 0.8886 for the SVM. Averaged over all
cross-dataset transfers originating from these two sources, DeepArrhythmia
achieves a mean Micro-F1 of 0.824, ahead of the SVM (0.811) and well ahead of the
1D ResNet (0.459). The SVM's figure is inflated by off-diagonal collapse to
majority-class prediction (Macro-F1 $0.649\rightarrow0.294$), whereas DeepArrhythmia keeps
detecting abnormal beats. Over all twelve off-diagonal transfers, DeepArrhythmia retains a
mean Micro-F1 of 0.698 versus 0.457 for the 1D ResNet. DeepArrhythmia retains abnormal-beat detection on most transfers; its worst case is confined to the VitalDB-trained direction, where it over-predicts a single abnormal class. These results indicate that, given a sufficiently informative
training distribution, DeepArrhythmia yields stronger cross-dataset
generalization than both a raw-signal deep network and a feature-based classifier
under matched conditions.

\begin{table*}[t]
\centering
\caption{Cross-dataset generalization performance measured by Micro-F1.}
\label{tab:cross-dataset-performance}
\resizebox{\textwidth}{!}{%
\begin{tabular}{lcccc}
\toprule
\textbf{Train $\backslash$ Test} & \textbf{MITBIH} & \textbf{Superventricular} & \textbf{INCART} & \textbf{VitalDB} \\
\midrule
MITBIH            & 0.9521 & 0.9272 & 0.5540 & 0.5514 \\
Superventricular  & 0.9351 & 0.9527 & 0.8711 & 0.6482 \\
INCART            & 0.9243 & 0.8942 & 0.9791 & 0.6735 \\
VitalDB           & 0.5835 & 0.0935 & 0.7214 & 0.9023 \\
\bottomrule
\end{tabular}%
}
\end{table*}

\begin{table*}[t]
\centering
\caption{Cross-dataset generalization performance of the 1D ResNet baseline measured by Micro-F1.}
\label{tab:cross-dataset-resnet}
\resizebox{\textwidth}{!}{%
\begin{tabular}{lcccc}
\toprule
\textbf{Train $\backslash$ Test} & \textbf{MITBIH} & \textbf{Superventricular} & \textbf{INCART} & \textbf{VitalDB} \\
\midrule
MITBIH            & 0.7596 & 0.8911 & 0.8635 & 0.6714 \\
Superventricular  & 0.0743 & 0.8135 & 0.4786 & 0.5827 \\
INCART            & 0.4199 & 0.6787 & 0.8693 & 0.5204 \\
VitalDB           & 0.1012 & 0.0759 & 0.1299 & 0.7859 \\
\bottomrule
\end{tabular}%
}
\end{table*}

\begin{table*}[t]
\centering
\caption{Cross-dataset generalization performance of the SVM baseline measured by Micro-F1.}
\label{tab:cross-dataset-svm}
\resizebox{\textwidth}{!}{%
\begin{tabular}{lcccc}
\toprule
\textbf{Train $\backslash$ Test} & \textbf{MITBIH} & \textbf{Superventricular} & \textbf{INCART} & \textbf{VitalDB} \\
\midrule
MITBIH            & 0.9366 & 0.8886 & 0.9384 & 0.6735 \\
Superventricular  & 0.8827 & 0.9374 & 0.8624 & 0.6735 \\
INCART            & 0.8825 & 0.8886 & 0.9845 & 0.6735 \\
VitalDB           & 0.8827 & 0.4505 & 0.8624 & 0.7501 \\
\bottomrule
\end{tabular}%
}
\end{table*}

\clearpage
\section{Knowledge Distillation for Morphology Analyzer}
\label{app:morph_kd}
To obtain a lightweight yet effective morphology analyzer, we adopt a knowledge distillation framework \citep{hsieh2023distilling}. Specifically, a teacher model is provided with the ground-truth beat label, ECG images, beat position, and related ECG features, and is prompted to generate morphology-focused explanations for each beat. During the inference,the student Morphology Analyzer is label-free. %
In our setup, Gemini 3.1 serves as the teacher model and Qwen3.5-4B serves as the student model. 
 For reproducibility, we provide the prompt format below. 

\begin{verbatim}
# This is a 10-second ECG segment from the MIT-BIH Arrhythmia Database
# (Record 100, Segment 0). The signal uses detailed AAMI beat labels.
# The following abnormal beats are annotated (time measured from segment start):

- t = 5.68 s -> label: A, peak amplitude: 1.255 mV,
  RR interval to previous beat: 0.656 s

## Normal/reference beats in the same segment:
- t = 0.21 s -> label: N, peak amplitude: 1.160 mV,
  RR interval to previous beat: N/A
- t = 1.03 s -> label: N, peak amplitude: 1.315 mV,
  RR interval to previous beat: 0.814 s
- t = 1.84 s -> label: N, peak amplitude: 1.350 mV,
  RR interval to previous beat: 0.814 s
- t = 2.63 s -> label: N, peak amplitude: 1.235 mV,
  RR interval to previous beat: 0.789 s
- t = 3.42 s -> label: N, peak amplitude: 1.210 mV,
  RR interval to previous beat: 0.789 s
- t = 4.21 s -> label: N, peak amplitude: 1.275 mV,
  RR interval to previous beat: 0.789 s
- t = 5.03 s -> label: N, peak amplitude: 1.310 mV,
  RR interval to previous beat: 0.817 s
- t = 6.67 s -> label: N, peak amplitude: 1.305 mV,
  RR interval to previous beat: 0.994 s
- t = 7.52 s -> label: N, peak amplitude: 1.285 mV,
  RR interval to previous beat: 0.842 s
- t = 8.33 s -> label: N, peak amplitude: 1.280 mV,
  RR interval to previous beat: 0.811 s
- t = 9.12 s -> label: N, peak amplitude: 1.200 mV,
  RR interval to previous beat: 0.792 s
- t = 9.89 s -> label: N, peak amplitude: 1.205 mV,
  RR interval to previous beat: 0.769 s

## Task Instructions:
For each annotated beat listed above, inspect the waveform at the given time
position and explain the specific morphological features that justify its label
(e.g., wide/narrow QRS, premature timing, abnormal P-wave, T-wave inversion,
compensatory pause, amplitude differences, etc.). Use the provided peak
amplitudes and RR intervals for both abnormal and normal beats as quantitative
support, especially when comparing premature or wide-complex beats against
nearby normal reference beats. Be concise, refer to exact time positions,
and avoid disclosing information that directly reveals the class label.
\end{verbatim}

\clearpage
\section{Example of Complete Pipeline for DeepArrhythmia}
\label{app:Example of Complete Pipeline for DeepArrhythmia}
To illustrate the end-to-end behavior of DeepArrhythmia, we provide a concise example interaction pipeline below.

\begin{Verbatim}[breaklines=true,breakanywhere=true,breaksymbolleft={},breaksymbolright={},fontsize=\scriptsize]
human: Given the ECG segment. Identify the arrhythmia class for each R peak.
Use the format [position:class] for each beat, separated by spaces.

LLM: Call_tools[Peak Detector]

Tool_output: R peaks are located at the following sample positions:
77 370 663 947 1231 1515 1809 2045 2403 2706 2998 3283 3560.

LLM: [77:N] [370:N] [663:N] [947:N] [1231:N] [1515:N] [1809:N] [2045:N] [2403:N] [2706:N] [2998:N] [3283:N] [3560:N]

LLM: Call_tools[Confidence Calculator]

Tool_output: The confidence is 0.989392. which is lower than Threshold 0.990529, Call Feature Extractor and Morphology Analyzer.

LLM: Call_tools[Feature Extractor]

Tool_output: Best ECG-classification features from config RR+norm_RR+HOS+myMorph (C=10): RR [pre, next, local, global] in samples describes beat timing, normalized RR [pre, next, local, global] shows timing relative to the record trend, peak amplitude (mV) captures R-peak height, HOS summarizes higher-order statistics of beat-shape intervals, and myMorph measures distances to representative morphology landmarks:

[77:RR=77,293,77,77;norm_RR=0.2693,1.0247,0.2694,0.2688;amp=1.1600;HOS=[1.1053,-0.3433,1.2196,0.0950,-0.2560,4.6480,2.4145,2.9714,2.7759,2.2308];myMorph=[0.9410,1.0038,0.9457,1.0186]]

[370:RR=293,293,185,77;norm_RR=1.0246,1.0247,0.6472,0.2688;amp=1.3150;HOS=[-0.4416,0.3670,0.9802,-0.1661,0.1701,2.0221,2.4089,2.4058,2.0685,2.0092];myMorph=[0.9454,1.0031,0.9396,1.0078]]

[663:RR=293,284,221,185;norm_RR=1.0246,0.9932,0.7731,0.6458;amp=1.3500;HOS=[-0.7455,1.1639,1.2957,-0.0901,-0.3184,2.8724,3.3559,3.1715,2.4122,1.8911];myMorph=[0.9140,1.0022,0.9523,1.0351]]

[947:RR=284,284,237,221;norm_RR=0.9931,0.9932,0.8282,0.7715;amp=1.2350;HOS=[-0.4047,-1.4507,1.2064,0.0547,0.0933,2.7067,6.1521,2.9994,2.0822,2.1874];myMorph=[0.9015,1.0039,0.9182,0.9532]]

[1231:RR=284,284,246,237;norm_RR=0.9931,0.9932,0.8613,0.8265;amp=1.2100;HOS=[-0.5737,-0.0519,1.3420,-0.4798,-0.7527,3.0223,2.1574,3.3234,2.4888,3.3500];myMorph=[0.9323,1.0023,0.9439,1.0004]]

[1515:RR=284,294,252,246;norm_RR=0.9931,1.0282,0.8833,0.8595;amp=1.2750;HOS=[0.2622,1.3791,1.2894,0.2687,0.2932,2.5440,5.3323,3.1648,1.9164,2.1889];myMorph=[0.9039,1.0038,0.9501,1.0355]]

[1809:RR=294,236,258,252;norm_RR=1.0281,0.8254,0.9041,0.8815;amp=1.3100;HOS=[-0.3015,0.7845,1.1879,0.2678,-0.0181,2.2311,3.6662,2.8801,2.3006,1.7732];myMorph=[0.9045,1.0019,0.9446,1.0376]]

[2045:RR=236,358,256,258;norm_RR=0.8253,1.2520,0.8943,0.9022;amp=1.2550;HOS=[-0.7270,-0.4906,1.1749,-0.5142,-0.3069,2.6356,2.8848,2.8712,3.0628,2.4644];myMorph=[0.8997,1.0033,0.9322,1.0002]]

[2403:RR=358,303,267,256;norm_RR=1.2519,1.0597,0.9341,0.8924;amp=1.3050;HOS=[-0.2710,0.8416,1.0584,0.2052,-0.3983,1.9386,3.6821,2.5817,2.2463,3.0717];myMorph=[0.9381,1.0025,0.9712,1.0192]]

[2706:RR=303,292,271,267;norm_RR=1.0596,1.0212,0.9466,0.9321;amp=1.2850;HOS=[-0.1878,0.8982,1.2062,-0.4905,-0.1132,1.8725,4.3606,2.9380,2.6907,2.3904];myMorph=[0.9376,1.0024,0.9540,1.0004]]

[2998:RR=292,285,292,271;norm_RR=1.0211,0.9967,1.0219,0.9446;amp=1.2800;HOS=[-0.9765,1.4327,1.3017,-0.4580,-0.2072,3.3586,4.3886,3.1866,2.7695,2.8285];myMorph=[0.9363,1.0026,0.9572,1.0056]]

[3283:RR=285,277,291,273;norm_RR=0.9966,0.9688,1.0191,0.9514;amp=1.2000;HOS=[-0.9414,0.8672,1.2883,-0.1023,0.1537,3.7850,3.5034,3.1642,1.8658,2.7679];myMorph=[0.9032,1.0026,0.9443,0.9939]]

[3560:RR=277,303,290,274;norm_RR=0.9687,1.0597,1.0135,0.9551;amp=1.2050;HOS=[-0.7466,1.2535,1.2960,0.1509,0.5796,3.2926,4.4371,3.1819,2.4824,2.8179];myMorph=[0.9191,1.0023,0.9366,1.0154]]

LLM: Call_tools[Morphology Analyzer]

Tool_output: Abnormal Beat: t = 5.68 s

Premature Timing: The RR interval of 0.656 s is significantly shorter than the preceding and subsequent baseline RR intervals, approximately 0.79 to 0.82 s. This beat occurs earlier than expected in the underlying rhythm.

Morphology: The QRS complex is narrow and similar in width to the surrounding beats. The preceding P wave is likely abnormal or hidden within the preceding T wave. The peak amplitude is 1.255 mV, which is within the range of the surrounding beats in this segment.

Reference Beats: t = 0.21 s, 1.03 s, 1.84 s, 2.63 s, 3.42 s, 4.21 s, 5.03 s

Consistent Morphology: These beats form a stable run. They exhibit narrow QRS complexes with consistent upright P waves preceding each QRS. Their RR intervals are relatively regular, clustering around 0.789 s to 0.817 s. Their amplitudes are also consistent, ranging from 1.160 mV to 1.350 mV.

t = 6.67 s

Compensatory Pause: This beat follows the premature beat at 5.68 s. The RR interval from the premature beat to this beat is 0.994 s, which is notably longer than the baseline rhythm. This represents a pause before the rhythm resumes its regular pattern. Morphology returns to the baseline pattern.

t = 7.52 s, 8.33 s, 9.12 s, 9.89 s

Return to Baseline: After the pause, the rhythm returns to the stable baseline pattern seen earlier. RR intervals normalize, ranging from 0.842 s to 0.769 s, and QRS morphology remains narrow and consistent, with stable amplitudes of approximately 1.200 mV to 1.285 mV.

Summary: The beat at 5.68 s is identified as abnormal because it occurs earlier than expected, has a narrow QRS complex, and is followed by a longer RR interval. The remaining beats show consistent narrow QRS morphology, associated P waves, and regular timing.

LLM: [77:N] [370:N] [663:N] [947:N] [1231:N] [1515:N] [1809:N] [2045:SVEB] [2403:N] [2706:N] [2998:N] [3283:N] [3560:N]

\end{Verbatim}
\clearpage

\section{Detail Result Analysis for DeepArrhythmia}
\label{app:detail analysis}
\begin{figure*}[t]
  \centering
  \begin{subfigure}{0.48\textwidth}
    \centering
    \includegraphics[width=\linewidth]{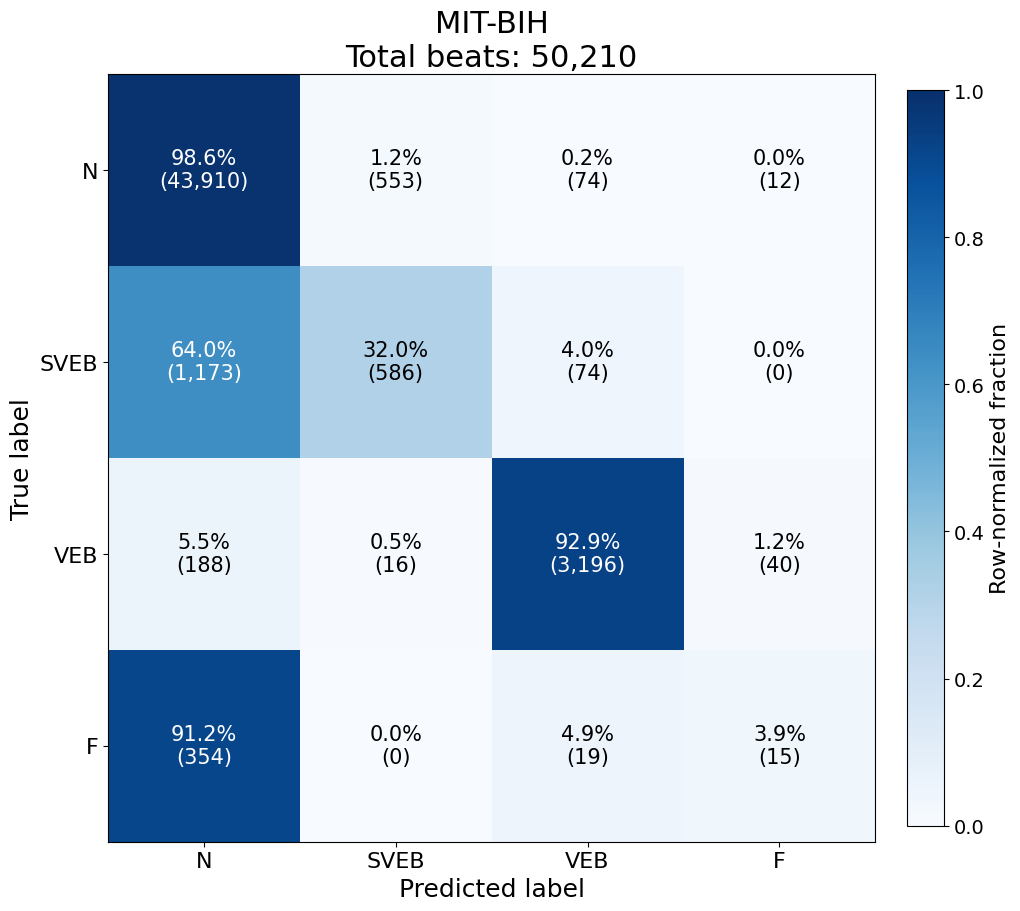}
    \caption{Confusion matrix on MIT-BIH Arrhythmia.}
    \label{fig:cm-mitbih}
  \end{subfigure}
  \hfill
  \begin{subfigure}{0.48\textwidth}
    \centering
    \includegraphics[width=\linewidth]{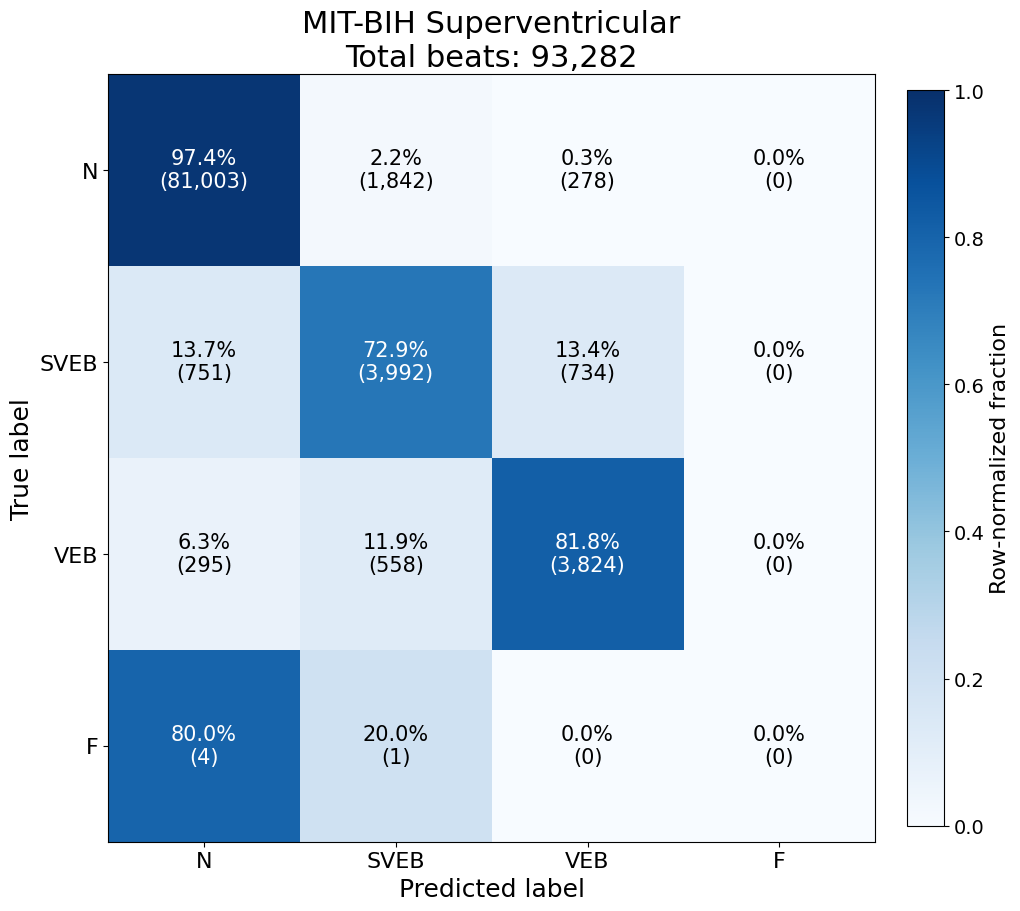}
    \caption{Confusion matrix on MIT-BIH Supraventricular.}
    \label{fig:cm-mitbih-s}
  \end{subfigure}

  \vspace{0.5em}

  \begin{subfigure}{0.48\textwidth}
    \centering
    \includegraphics[width=\linewidth]{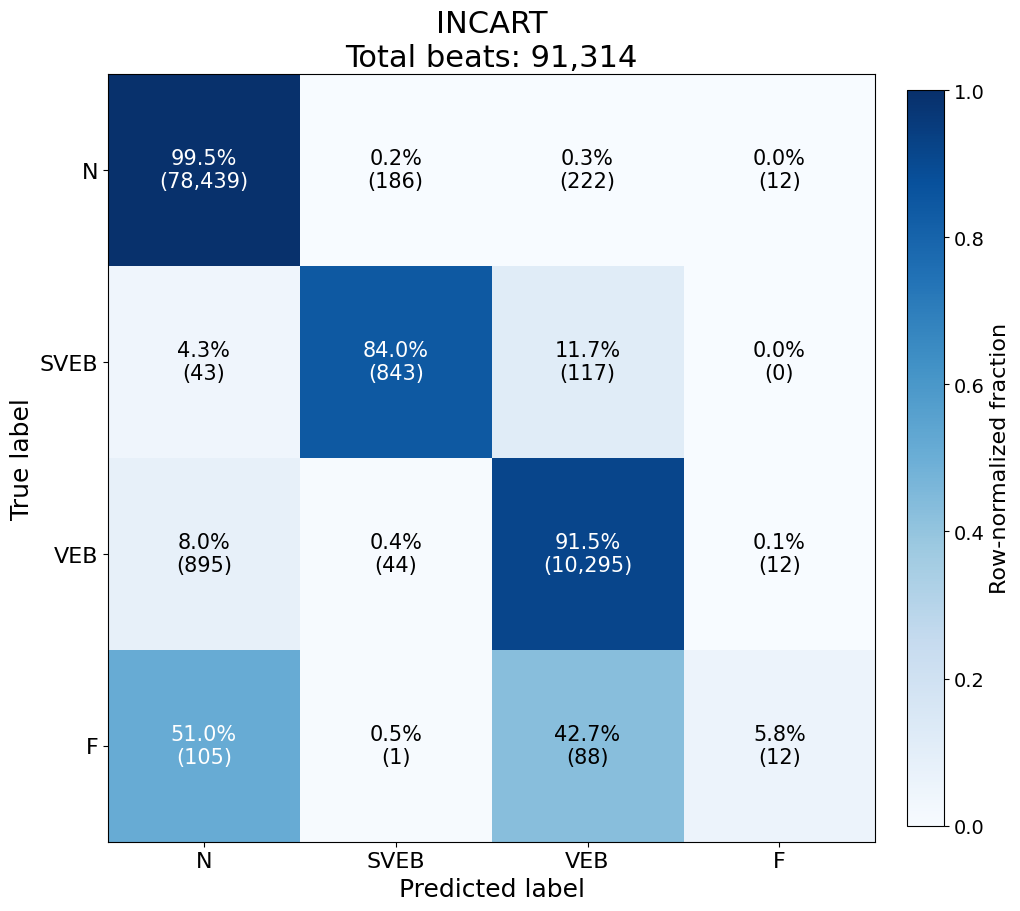}
    \caption{Confusion matrix on INCART.}
    \label{fig:cm-incart}
  \end{subfigure}
  \hfill
  \begin{subfigure}{0.48\textwidth}
    \centering
    \includegraphics[width=\linewidth]{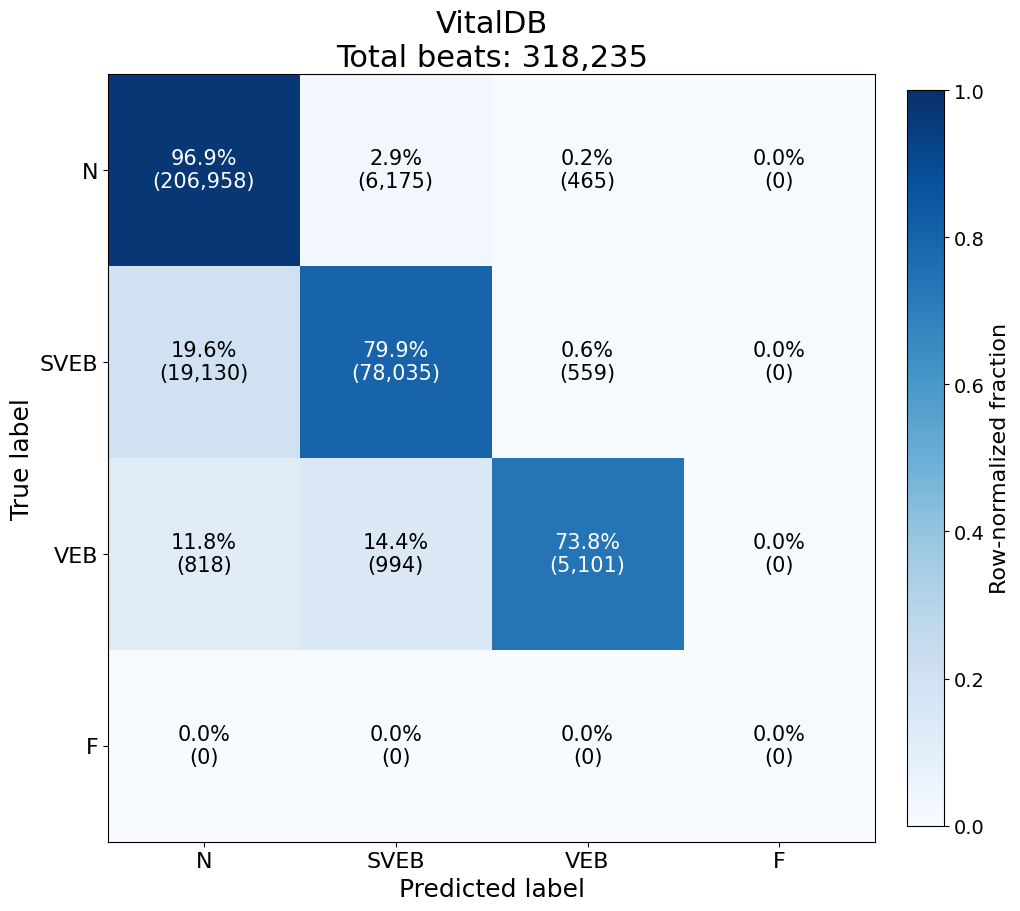}
    \caption{Confusion matrix on VitalDB.}
    \label{fig:cm-vitaldb}
  \end{subfigure}

  \caption{Confusion-matrix panels for DeepArrhythmia across four datasets.}
  \label{fig:cm-2x2-deeparrhythmia}
\end{figure*}

Table~\ref{tab:deeparrhythmia-prf} and Figure~\ref{fig:cm-2x2-deeparrhythmia} jointly characterize the class-wise behavior of the DeepArrhythmia threshold-induced routed variant across the four evaluation datasets, with the feature-based SVM and the raw-signal 1D ResNet included in Table~\ref{tab:deeparrhythmia-prf} as reference baselines. Table~\ref{tab:deeparrhythmia-prf} reports precision, recall, and F1 for each class, whereas Figure~\ref{fig:cm-2x2-deeparrhythmia} visualizes the corresponding confusion patterns for DeepArrhythmia. Across all datasets, DeepArrhythmia recognizes normal beats reliably, with F1 values ranging from 0.9396 to 0.9908, indicating that the model preserves high accuracy on the dominant class; the 1D ResNet is lower on this class (0.8499--0.9234), whereas the SVM is comparable (0.8224--0.9928). Ventricular beats are also classified robustly by DeepArrhythmia, with F1 scores above 0.78 on all four datasets, whereas the 1D ResNet is markedly weaker on this class (F1 between 0.4027 and 0.8019), underscoring that access to raw signal alone does not guarantee reliable ventricular discrimination. Performance on supraventricular beats is more variable: for DeepArrhythmia it is weakest on MIT-BIH Arrhythmia (F1 = 0.3922) but improves substantially on MIT-BIH Supraventricular, INCART, and VitalDB, where the corresponding F1 scores reach 0.6726, 0.8117, and 0.8532, respectively. On this class DeepArrhythmia is the strongest of the three methods on every dataset; the 1D ResNet is consistently the weakest (e.g., F1 = 0.0728 on MIT-BIH and 0.2174 on INCART), and the SVM lies in between.

The principal limitation concerns fusion beats. As shown in Table~\ref{tab:deeparrhythmia-prf}, absolute performance for class F remains poor for all three methods on the datasets where meaningful evaluation is possible. This weakness is expected given the extreme rarity and heterogeneity of fusion morphologies, but the comparison with the baselines is still informative: on MIT-BIH, DeepArrhythmia attains the highest F1 for class F (0.0659), ahead of both the 1D ResNet (0.0404) and the SVM (0.0253), and on INCART it reaches 0.0992 while the 1D ResNet attains only 0.0603 and the SVM remains at 0.0000. We do not report F for MIT-BIH Supraventricular or VitalDB because the corresponding evaluation splits contain fewer than five fusion-beat examples, making class-wise estimates too unstable for meaningful comparison. On MIT-BIH, all three methods are evaluated on the identical DS2 test split ($\approx$50k beats: $\approx$44k N, $\approx$2k S, $\approx$3.2k V, and only 388 F). Taken together, these results show that the difficulty of the minority classes (S and F) is shared across a feature-based classifier, a deep raw-signal network, and DeepArrhythmia, and is therefore intrinsic to the extreme class imbalance of beat-level ECG classification rather than specific to our framework; among the three methods, DeepArrhythmia consistently attains the strongest minority-class performance. The confusion matrices in Figure~\ref{fig:cm-2x2-deeparrhythmia} are consistent with these quantitative findings: residual errors are predominantly misses toward the normal class rather than confusion among abnormal classes. On MIT-BIH, 1{,}173 of 1{,}833 S beats (64\%) are predicted as N (32\% correct), because S beats preserve a near-normal QRS and are distinguished mainly by timing and P-wave cues; 354 of 388 F beats (91\%) are predicted as N (5\% as V), since a fusion beat is a blend of a normal and a ventricular complex. The diagonal corresponding to normal beats remains dominant across datasets.

To quantify statistical uncertainty, we compute 95\% bootstrap confidence intervals for MIT-BIH Macro-F1: DeepArrhythmia 0.593 [0.583, 0.603], SVM 0.523 [0.516, 0.530], and 1D ResNet 0.362 [0.356, 0.367]; the intervals do not overlap. Under class-balanced evaluation (balanced accuracy, i.e., mean per-class recall) on MIT-BIH, the ordering is preserved: DeepArrhythmia 0.568, SVM 0.516, and 1D ResNet 0.449.

The rendered waveform image is derived from the raw signal and therefore carries no independent information, yet removing it lowers MIT-BIH Macro-F1 from 0.5929 to 0.5394 (Table~\ref{tab:ablation-study}). A per-beat analysis shows that the beats recovered by the image are overwhelmingly abnormal beats that the model without the image labels as normal, chiefly S beats: 138 of the 143 recovered S beats were labelled N without the image. These are premature beats (normalized pre-RR $\approx$0.71) with a near-normal QRS, for which RR timing alone is insufficient; the image provides a complementary spatial representation of P-wave presence and subtle QRS shape that the scalar features compress away.

\begin{table*}[t]
  \centering
  \caption{Class-wise precision, recall, and F1 for DeepArrhythmia, SVM, and 1D ResNet across the four evaluation datasets. `--' indicates that the corresponding class is not reported because the evaluation split contains fewer than five beats.}
  \label{tab:deeparrhythmia-prf}
  \small
  \setlength{\tabcolsep}{3pt}
  \resizebox{\textwidth}{!}{%
  \begin{tabular}{llcccccccccccc}
    \toprule
    \multirow{2}{*}{\textbf{Model}} & \multirow{2}{*}{\textbf{Class}} & \multicolumn{3}{c}{\textbf{MIT-BIH}} & \multicolumn{3}{c}{\textbf{MIT-BIH-SUP}} & \multicolumn{3}{c}{\textbf{INCART}} & \multicolumn{3}{c}{\textbf{VitalDB}} \\
    \cmidrule(r){3-5} \cmidrule(r){6-8} \cmidrule(r){9-11} \cmidrule(r){12-14}
     &  & \textbf{Prec.} & \textbf{Rec.} & \textbf{F1} & \textbf{Prec.} & \textbf{Rec.} & \textbf{F1} & \textbf{Prec.} & \textbf{Rec.} & \textbf{F1} & \textbf{Prec.} & \textbf{Rec.} & \textbf{F1} \\
    \midrule
    \multirow{4}{*}{DeepArrhythmia} & N & 0.9624 & 0.9857 & 0.9739 & 0.9872 & 0.9745 & 0.9808 & 0.9869 & 0.9947 & 0.9908 & 0.9121 & 0.9689 & 0.9396 \\
     & S & 0.5074 & 0.3197 & 0.3922 & 0.6244 & 0.7289 & 0.6726 & 0.7849 & 0.8405 & 0.8117 & 0.9159 & 0.7985 & 0.8532 \\
     & V & 0.9503 & 0.9291 & 0.9396 & 0.7907 & 0.8176 & 0.8040 & 0.9602 & 0.9154 & 0.9373 & 0.8328 & 0.7379 & 0.7825 \\
     & F & 0.2239 & 0.0387 & 0.0659 & -- & -- & -- & 0.3333 & 0.0583 & 0.0992 & -- & -- & -- \\
    \midrule
    \multirow{4}{*}{SVM} & N & 0.9533 & 0.9802 & 0.9665 & 0.9735 & 0.9714 & 0.9724 & 0.9908 & 0.9949 & 0.9928 & 0.8340 & 0.8111 & 0.8224 \\
     & S & 0.3791 & 0.1415 & 0.2060 & 0.6096 & 0.5855 & 0.5973 & 0.8068 & 0.7410 & 0.7725 & 0.5965 & 0.6359 & 0.6155 \\
     & V & 0.8696 & 0.9199 & 0.8938 & 0.7017 & 0.7616 & 0.7304 & 0.9596 & 0.9562 & 0.9579 & 0.3794 & 0.3470 & 0.3625 \\
     & F & 0.0328 & 0.0206 & 0.0253 & -- & -- & -- & 0.0000 & 0.0000 & 0.0000 & -- & -- & -- \\
    \midrule
    \multirow{4}{*}{1D ResNet}
    & N & 0.9366 & 0.7919 & 0.8582 & 0.9872 & 0.8244 & 0.8985 & 0.9880 & 0.8667 & 0.9234 & 0.9177 & 0.7914 & 0.8499 \\
    & S & 0.0591 & 0.0946 & 0.0728 & 0.2319 & 0.7614 & 0.3555 & 0.1313 & 0.6334 & 0.2174 & 0.6534 & 0.7791 & 0.7107 \\
    & V & 0.3342 & 0.8230 & 0.4753 & 0.5542 & 0.6829 & 0.6119 & 0.7123 & 0.9174 & 0.8019 & 0.2809 & 0.7111 & 0.4027 \\
    & F & 0.0265 & 0.0851 & 0.0404 & -- & -- & -- & 0.0324 & 0.4320 & 0.0603 & -- & -- & -- \\
    \bottomrule
  \end{tabular}
  }
\end{table*}

\section{Faithfulness Evaluation}
\label{app:Faithfulness Evaluation}
To assess whether the generated explanations reflect evidence used by the model, we perform an evidence-deletion test on the MIT-BIH test set using 160 beats stratified by cue (QRS 53, P-wave 54, RR 53). For numerical and textual morphology analyses that cite QRS width, P-wave evidence, or RR-interval regularity, we apply targeted perturbations that selectively disrupt the cue explicitly referenced in the explanation and compare the resulting classification accuracy against matched control perturbations applied to unrelated regions. As shown in Table~\ref{tab:faithfulness_deletion}, targeted perturbations cause a dramatic drop in accuracy relative to control perturbations, both overall (1.88\% vs. 92.50\%) and within each cue category. These results suggest that the generated explanations are at least partially faithful, in the sense that the model's predictions depend strongly on the evidence explicitly cited in the explanation.

\begin{table}[t]
\centering
\caption{Faithfulness evaluation via evidence deletion. We report classification accuracy after targeted perturbations and matched control perturbations. Targeted perturbations disrupt the cue explicitly referenced in the explanation, while control perturbations are applied to unrelated regions. Results are shown overall and by cue type. Lower accuracy under targeted perturbations indicates that the model relies on the cited evidence.}
\label{tab:faithfulness_deletion}
\begin{tabular}{lccc}
\toprule
\textbf{Cue type} & \textbf{Targeted accuracy} & \textbf{Control accuracy} & \textbf{Sample size} \\
\midrule
Overall   & 3 / 160 (1.88\%)  & 148 / 160 (92.50\%) & 160 \\
QRS       & 1 / 53 (1.89\%)   & 51 / 53 (96.23\%)   & 53 \\
P\_WAVE   & 1 / 54 (1.85\%)   & 49 / 54 (90.74\%)   & 54 \\
RR        & 1 / 53 (1.89\%)   & 48 / 53 (90.57\%)   & 53 \\
\bottomrule
\end{tabular}
\end{table}

\begin{figure*}[t]
  \centering
  \begin{subfigure}{0.48\textwidth}
    \centering
    \includegraphics[width=\linewidth]{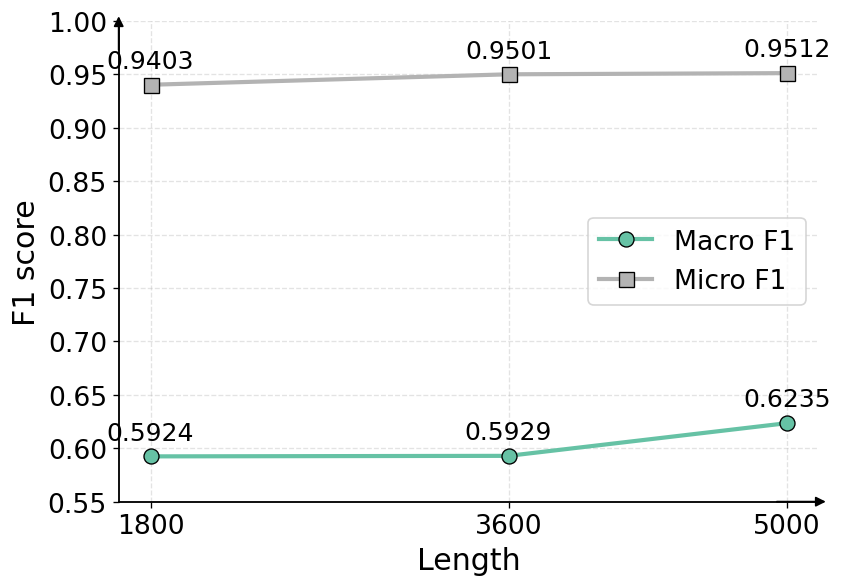}
    \caption{Impact of context length on DeepArrhythmia performance.}
    \label{fig:context-length-impact}
  \end{subfigure}
  \hfill
  \begin{subfigure}{0.48\textwidth}
    \centering
    \includegraphics[width=\linewidth]{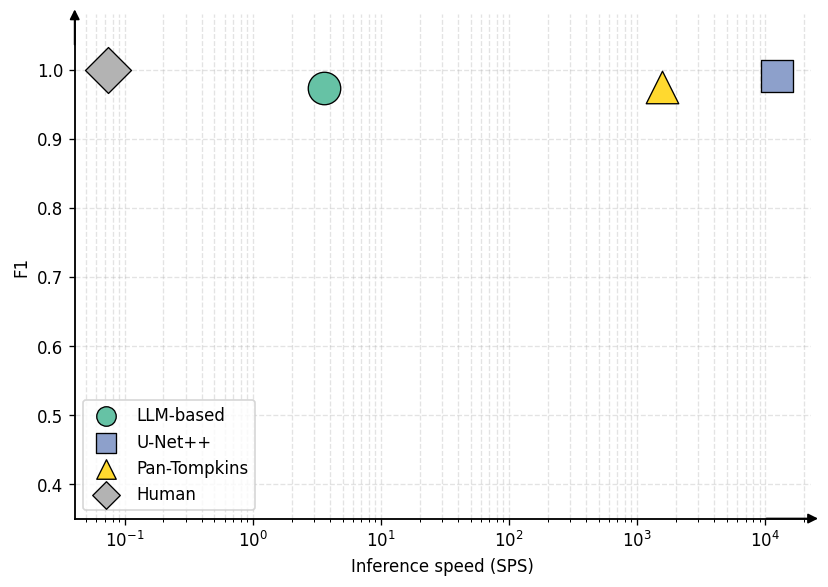}
    \caption{Peak detector trade-off between throughput (samples per second) and F1 score.}
    \label{fig:inference-peak-detector}
  \end{subfigure}
  \caption{Sensitivity analyses for context length and peak-detector selection.}
  \label{fig:context-and-peak-analysis}
\end{figure*}

\clearpage
\section{Impact of Context Window Length and Boundary Truncation}
\label{app:context_length}
Figure~\ref{fig:context-length-impact} evaluates the sensitivity of DeepArrhythmia with rich-evidence to input context length. Increasing the context window from 1,800 samples (5 s) to 3,600 samples (10 s) yields consistent gains, with Macro-F1 improving from 0.5924 to 0.5929 and Micro-F1 improving from 0.9403 to 0.9501. Extending the context further to 5,000 samples provides an additional improvement, increasing Macro-F1 to 0.6235 and Micro-F1 to 0.9512. These results indicate that longer temporal context strengthens rhythm-level reasoning and particularly benefits class-balanced performance.

To assess the effect of fixed-window boundary truncation, we compare beats at the segment
boundary (the first and last beat of each 10-second window) against interior beats on the
MIT-BIH test set (Table~\ref{tab:boundary}). Boundary beats are only marginally harder
overall---Micro-F1 is $0.9435$ versus $0.9521$ for interior beats (a $0.9$-point gap), and
the dominant normal class is essentially unaffected ($0.970$ vs.\ $0.975$). The residual gap
is instead concentrated on the abnormal classes that most depend on cross-beat context:
V-class F1 drops from $0.950$ to $0.901$ and the S and F classes decline similarly, because a
boundary beat lacks a complete preceding/following RR interval and neighbouring morphology.
This robustness is largely attributable to the shift-based re-anchoring augmentation, which
exposes each beat at many within-window positions during training.

\begin{table}[t]
\centering
\caption{Boundary versus interior beat classification on the MIT-BIH test set. Boundary beats (the first and last beat of each 10-second window) are only marginally harder overall, and the residual gap is concentrated on the context-dependent abnormal classes, while the normal class is essentially unaffected.}
\label{tab:boundary}
\begin{tabular}{lrcccccc}
\toprule
\textbf{Beat group} & \textbf{\#Beats} & \textbf{Micro-F1} & \textbf{Macro-F1} & \textbf{N} & \textbf{S} & \textbf{V} & \textbf{F} \\
\midrule
Boundary (first/last) & 8{,}066  & 0.9435 & 0.5673 & 0.970 & 0.360 & 0.901 & 0.038 \\
Interior              & 41{,}981 & 0.9521 & 0.5987 & 0.975 & 0.400 & 0.950 & 0.070 \\
\bottomrule
\end{tabular}
\end{table}

Because this degradation is confined to the abnormal classes---the normal class is already
unaffected by boundary position---we test the overlapping-window remedy precisely where it
could help: the abnormal (S/V/F) boundary beats. We therefore select every S, V, or F beat
that occurs first or last in a window and can be re-centered within the record, yielding
$867$ target beats (of which $3$ lie too close to a record edge to center and are excluded).
For each such beat we re-classify it inside a $10$-second window \emph{re-centered} on it,
emulating an overlapping (sliding-window) inference scheme in which every beat is evaluated
where it has full surrounding context; the $N$ class and interior beats are excluded because
they show no boundary deficit to recover and would only dilute the measurement. As shown in
Table~\ref{tab:boundary-sliding}, re-centering these boundary beats raises their accuracy
from $0.606$ to $0.694$ and Macro-F1 from $0.466$ to $0.584$, recovering the boundary gap and
matching interior-beat performance, with the largest gain on the context-dependent fusion
class (F1 $0.038 \rightarrow 0.254$); $131$ of the $867$ beats are corrected while only $54$
regress. This confirms that boundary degradation stems from missing context rather than a
model deficiency, and that a lightweight overlapping-window scheme resolves it. We use
non-overlapping windows in the main results for training and inference efficiency, since
overlap multiplies the compute per beat.

\begin{table}[t]
\centering
\caption{Effect of beat position within the fixed context window on abnormal (S/V/F) beat classification, MIT-BIH test set. Boundary beats are the weakest; interior beats are stronger; and re-classifying each boundary beat in a window re-centered on it (overlapping/sliding-window inference) recovers the gap and matches interior-beat performance. ``Boundary'' and ``Re-centered'' are the same $867$ beats.}
\label{tab:boundary-sliding}
\begin{tabular}{lrccccc}
\toprule
\textbf{Beat position} & \textbf{\#Beats} & \textbf{Acc.} & \textbf{Macro-F1} & \textbf{S} & \textbf{V} & \textbf{F} \\
\midrule
Boundary (original)   & 867     & 0.606 & 0.466 & 0.445 & 0.915 & 0.038 \\
Interior (reference)  & 4{,}680 & 0.683 & 0.507 & 0.490 & 0.961 & 0.072 \\
Re-centered (sliding) & 867     & \textbf{0.694} & \textbf{0.584} & \textbf{0.531} & \textbf{0.966} & \textbf{0.254} \\
\bottomrule
\end{tabular}
\end{table}
\section{Impact of Peak Detector}
\label{app:peak detector}
Peak detection is a critical upstream stage in beat-centered ECG analysis because errors in R-peak localization propagate directly to subsequent feature extraction, morphology reasoning, and beat-label alignment. To quantify this dependency, we compare representative detectors on 10-second MIT-BIH segments, including Pan--Tompkins \citep{pan2007real}, U-Net++ \citep{zhou20211d}, an LLM-based detector \citep{li2025peak}, and human annotations as a reference.

Figure~\ref{fig:inference-peak-detector} summarizes the trade-off between detection quality, measured by F1 score using a 30~ms peak-matching tolerance~\citep{li2025peak}, and runtime throughput, measured in samples per second. Throughput is measured on the same single-GPU A6000 inference setup described in Appendix~\ref{app:implementaiton}. Among the automated methods, U-Net++ achieves the highest F1 score (0.9913) while also providing the fastest inference speed (12{,}379 segments per second); only the human-annotation reference performs better in detection quality. This favorable accuracy--efficiency profile motivates our use of U-Net++ as the default peak detector in DeepArrhythmia.

\begin{table}[t]
\centering
\caption{Class-wise decrease in F1 score ($\Delta$F1 = F1$_\text{clean}$ - F1$_\text{stress}$) under controlled peak-detection stress conditions: masking (a true R-peak is removed), mislocalization (a peak is shifted by 6--30 samples), and extra (a spurious R-peak is inserted). ``--'' indicates the interfered beat has no ground-truth label (masked or spurious). Positive values denote degradation; negative values indicate improvement. Class S is the most sensitive to mislocalization and spurious peaks, whereas under masking the collateral effect is largest for the minority F and V classes.}
\label{tab:stress_f1_drop}
\resizebox{\columnwidth}{!}{%
\begin{tabular}{lcccccc}
\toprule
\textbf{Label} & \textbf{Mask} & \textbf{Mask} & \textbf{Mislocalize} & \textbf{Mislocalize} & \textbf{Extra} & \textbf{Extra} \\
 & \textbf{Interfered} & \textbf{Non-interfered} & \textbf{Interfered} & \textbf{Non-interfered} & \textbf{Interfered} & \textbf{Non-interfered} \\
\midrule
N & -- & 0.004 & 0.020 & 0.014 & -- & 0.028 \\
S & -- & 0.019 & 0.256 & 0.136 & -- & \textbf{0.302} \\
V & -- & 0.043 & 0.137 & 0.025 & -- & 0.072 \\
F & -- & 0.056 & 0.039 & 0.033 & -- & 0.050 \\
\bottomrule
\end{tabular}%
}
\end{table}

To further characterize the downstream consequences of detector errors, we evaluate three controlled perturbation settings on the rich-evidence variant of DeepArrhythmia, using a class-enriched 400-segment MIT-BIH subset (100 all-normal segments and 100 segments each containing S, V, or F beats): \emph{masking}, which removes a true R peak; \emph{mislocalization}, which shifts a detected peak to the left or right by 6 to 30 samples; and \emph{extra}, which inserts a spurious R peak into a wide gap between two true beats. These perturbations isolate three common failure modes of practical peak detectors---missed detections, temporal offsets, and false detections---and allow us to examine how each propagates to beat-level classification.

Table~\ref{tab:stress_f1_drop} reveals three main patterns. First, for non-interfered collateral beats, mislocalization causes larger degradation than masking; directly masked beats are omitted from prediction and are therefore not directly comparable. The largest degradation occurs for supraventricular beats ($\Delta$F1 = 0.256), followed by ventricular beats ($\Delta$F1 = 0.137), whereas the effect is smaller for normal and fusion beats ($\Delta$F1 values of 0.020 and 0.039, respectively). This suggests that precise beat alignment is especially important for classes whose discrimination depends on localized morphology and rhythm context, particularly class S. Second, the collateral effect on non-interfered beats remains modest under masking ($\Delta$F1 between 0.004 and 0.056), but becomes larger under mislocalization, again most notably for class S ($\Delta$F1 = 0.136). A plausible interpretation is that a missed beat primarily removes one decision anchor, whereas a mislocalized beat can actively distort downstream feature extraction and morphology reasoning. Third, inserting a spurious peak produces the same qualitative signature but the strongest collateral effect on class S of all three settings ($\Delta$F1 = 0.302), while remaining modest overall (Micro-F1 $\Delta$ = 0.043); a false anchor injects an inconsistent RR pattern and morphology descriptor that most affects the timing- and context-dependent S class. Examining the spurious anchors themselves, only 34.6\% are labelled normal, whereas 63.4\% are assigned an arrhythmic class (52.0\% V, 11.4\% S). Because prediction is anchored to detected peaks, spurious detections are thus not robustly rejected---a limitation shared with detector-dependent beat-level pipelines that motivates tighter detector--classifier coupling in future work.

\section{Impact of Noise}
\label{app: Impact of Noise}

\begin{figure}[t]
\centering
\includegraphics[width=\columnwidth]{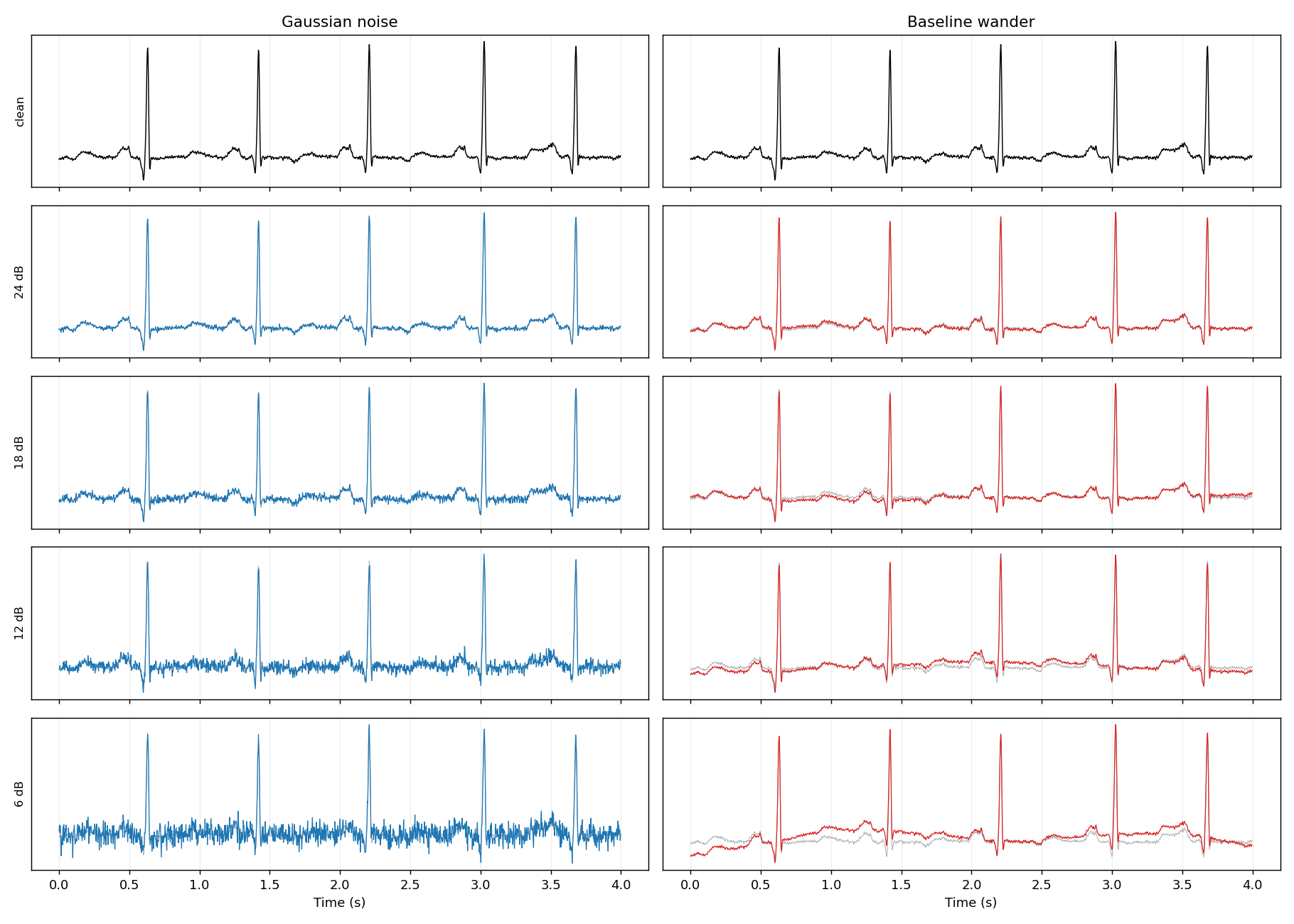}
\caption{Illustration of the two noise types at each SNR on a representative $4$-second
MIT-BIH segment (lead MLII). The clean trace (top) is overlaid in grey on each corrupted
signal for reference. Gaussian noise (left, blue) adds broadband high-frequency
fluctuations, while baseline wander (right, red) adds low-frequency drift.}
\label{fig:noise-illustration}
\end{figure}

A natural concern for any morphology-aware classifier is its resilience to signal
corruption. We therefore quantify how beat-level accuracy responds to two canonical ECG
disturbances---\emph{additive Gaussian noise} and \emph{baseline wander}---applied to each
$10$-second segment at signal-to-noise ratios (SNR) of $24$, $18$, $12$, and $6$\,dB.
Figure~\ref{fig:noise-illustration} illustrates the two disturbances at each level: at
$24$\,dB the corruption is barely perceptible, whereas at $6$\,dB Gaussian noise obscures
the P and T waves while leaving the QRS complex identifiable, and baseline wander induces a
pronounced low-frequency drift of the isoelectric line.

\begin{figure}[t]
\centering
\begin{subfigure}[t]{0.49\columnwidth}
  \includegraphics[width=\textwidth]{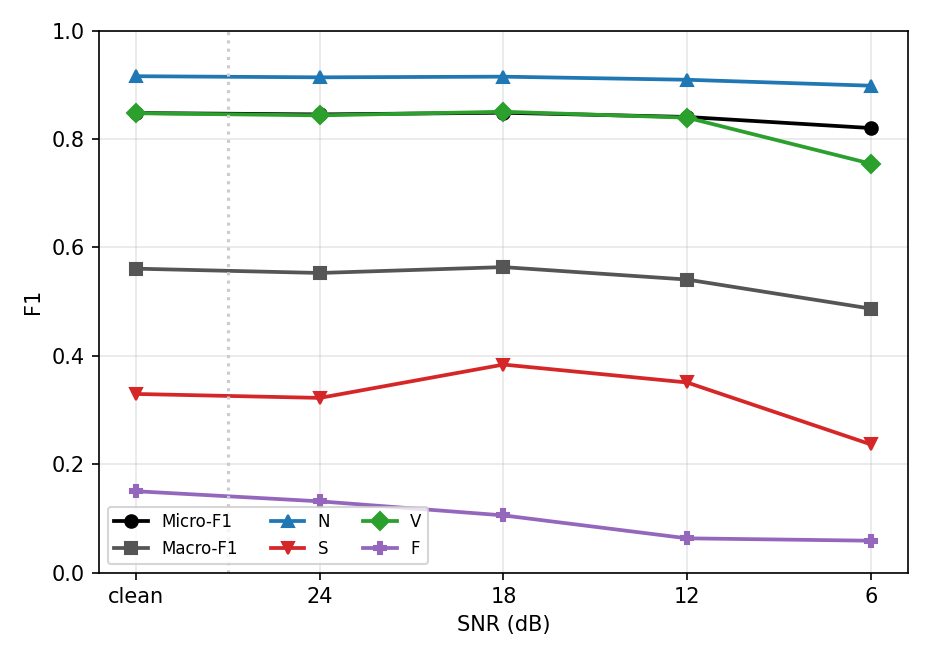}
  \caption{Additive Gaussian noise}
  \label{fig:noise-gaussian}
\end{subfigure}\hfill
\begin{subfigure}[t]{0.49\columnwidth}
  \includegraphics[width=\textwidth]{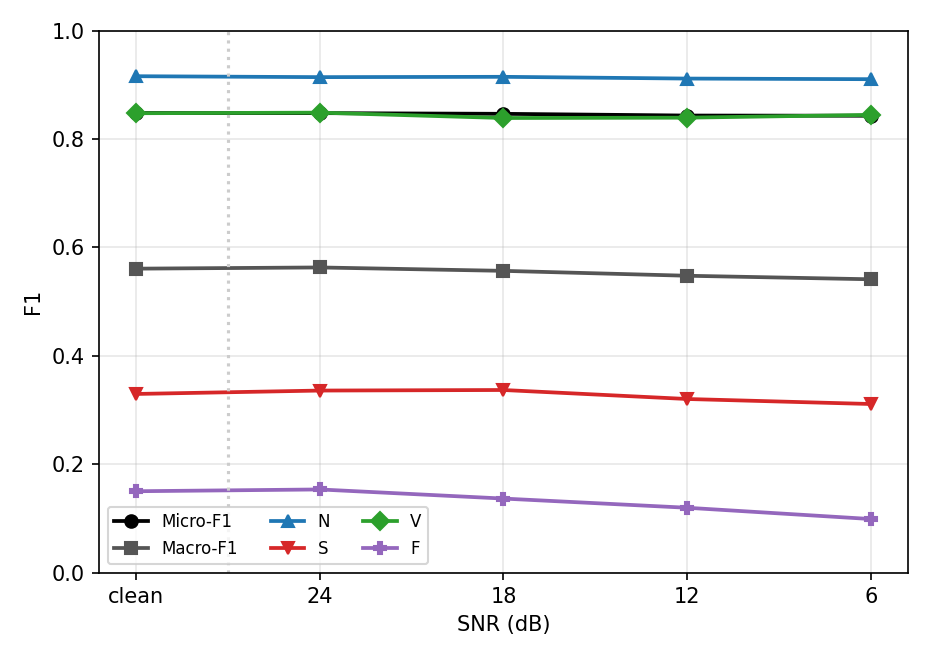}
  \caption{Baseline wander}
  \label{fig:noise-bwander}
\end{subfigure}
\caption{Beat-level F1 (Micro, Macro, and per-class N/S/V/F) as a function of SNR under
(a) additive Gaussian noise and (b) baseline wander, with the clean condition as the
leftmost reference. Performance degrades gracefully and never collapses: baseline wander
is almost fully absorbed, while Gaussian noise leaves accuracy essentially unchanged down
to $12$\,dB and induces only a localized decline at $6$\,dB, concentrated on the
morphology-dependent V class and the minority S and F classes.}
\label{fig:signal-noise}
\end{figure}

Crucially, each perturbation is propagated through the entire input pipeline rather than a
single modality: the corrupted waveform is re-encoded by the ECG-CoCa encoder, the ECG
image is re-rendered from it, and the numerical morphology descriptors (peak amplitude,
HOS, and myMorph) are re-extracted from it. We hold the R-peak anchors and RR intervals at
their clean values so as to isolate the effect of \emph{signal quality} from that of
\emph{peak-detection error}, which is examined separately in
Appendix~\ref{app:peak detector}. All conditions use the rich-evidence variant and are evaluated on the same
400-segment MIT-BIH subset as the peak-perturbation study ($5{,}374$ beats),
ensuring that differences reflect noise sensitivity alone rather than sampling variation. Because the subset is class-enriched, the clean
reference (Micro-F1 $0.8482$) sits below the $0.9501$ of Table~\ref{tab:result-performance} for the same model, while Macro-F1 ($0.5609$ vs.\ $0.5929$) is close. SNR is defined as $10\log_{10}$(signal power / noise power) per segment and lead. Figure~\ref{fig:signal-noise} visualizes the corresponding degradation curves.

Across every tested condition, DeepArrhythmia degrades gracefully rather than collapsing.
Baseline wander is almost entirely absorbed: from the clean reference to the most severe
$6$\,dB level, Micro-F1 declines by only $0.5$ points ($0.848\!\rightarrow\!0.843$) and
Macro-F1 by $2.0$ points ($0.561\!\rightarrow\!0.541$). This is expected, as the
median-filter baseline removal in preprocessing suppresses low-frequency drift while
preserving beat morphology. Additive Gaussian noise is more disruptive yet remains well
tolerated over the practical range: accuracy is statistically flat down to $12$\,dB,
staying within one Micro-F1 point of the clean baseline, and a pronounced decline emerges
only at the extreme $6$\,dB setting (Micro-F1 $-2.8$, Macro-F1 $-7.4$ points).

The structure of this degradation is consistent with the design of the framework. As
Figure~\ref{fig:signal-noise} shows, the loss is concentrated on classes whose
discrimination depends on fine morphology: under Gaussian noise at $6$\,dB the V-class F1
falls from $0.848$ to $0.755$ and the minority S and F classes deteriorate, whereas the
rhythm-anchored N class remains essentially stable ($0.916\!\rightarrow\!0.899$).
Intuitively, noise corrupts the morphological evidence---the rendered waveform image and
the HOS/myMorph descriptors---but not the timing-based RR intervals and peak anchors, so
decisions grounded in rhythm context stay reliable. Taken together, these results
demonstrate that DeepArrhythmia is robust to realistic levels of signal noise, exhibiting
only graceful, morphology-localized degradation even at severe SNR (Table~\ref{tab:noise}). Noise also lowers
confidence, so a fixed threshold routes degraded segments to rich evidence more often; the routing mechanism therefore fails safe.

\begin{table}[t]
\centering
\caption{Beat-level Micro-F1 and Macro-F1 under additive Gaussian noise and baseline wander (rich-evidence variant, 400-segment MIT-BIH subset, $5{,}374$ beats).}
\label{tab:noise}
\small
\begin{tabular}{lcccc}
\toprule
 & \multicolumn{2}{c}{Gaussian noise} & \multicolumn{2}{c}{Baseline wander} \\
\cmidrule(r){2-3} \cmidrule(r){4-5}
SNR & Micro-F1 & Macro-F1 & Micro-F1 & Macro-F1 \\
\midrule
Clean & 0.8482 & 0.5609 & 0.8482 & 0.5609 \\
24 dB & 0.8456 & 0.5529 & 0.8480 & 0.5632 \\
18 dB & 0.8487 & 0.5638 & 0.8467 & 0.5569 \\
12 dB & 0.8409 & 0.5408 & 0.8435 & 0.5478 \\
6 dB  & 0.8204 & 0.4872 & 0.8431 & 0.5413 \\
\bottomrule
\end{tabular}
\end{table}

\section{Impact of Backbone Scale}
\label{app:model_scale}

\begin{figure}[t]
  \centering
  \begin{subfigure}[t]{0.48\linewidth}
  
    \includegraphics[width=\linewidth]{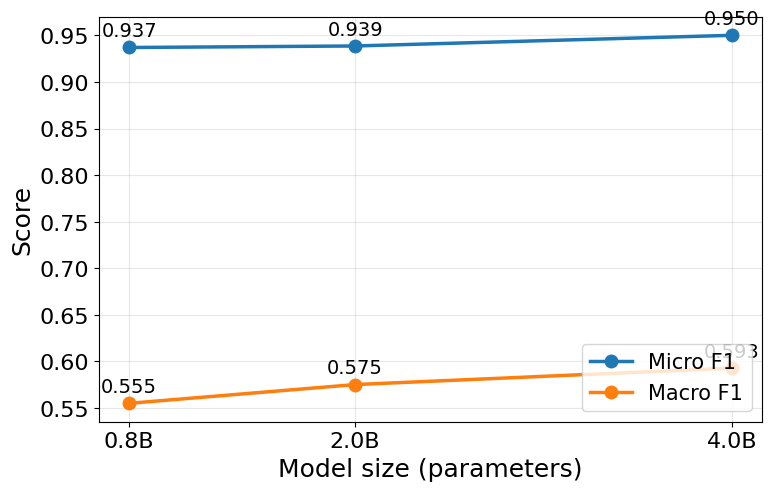}%
    
    \caption{Aggregate beat-level Micro-F1 and Macro-F1.}
    \label{fig:model_scale_aggregate}
  \end{subfigure}
  \hfill
  \begin{subfigure}[t]{0.48\linewidth}
    \IfFileExists{images/model_scale_per_class.png}{%
      \includegraphics[width=\linewidth]{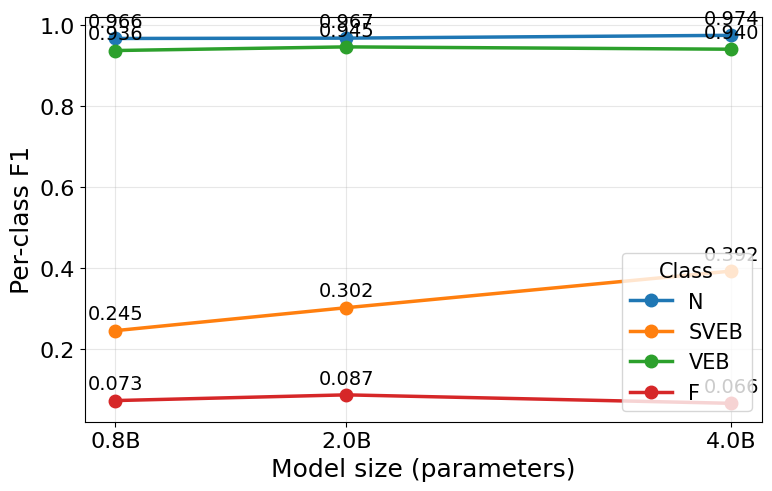}%
    }{%
      \fbox{\parbox[c][3.0cm][c]{0.95\linewidth}{\centering
      Placeholder for per-class model-scale plot.\\
      Add \texttt{images/model\_scale\_perclass.png}.}}%
    }
    \caption{Per-class beat-level F1.}
    \label{fig:model_scale_perclass}
  \end{subfigure}
  \caption{Effect of Qwen3.5 backbone scale on DeepArrhythmia beat classification
    on MIT-BIH Arrhythmia. Performance increases with backbone size, primarily
    through improved \emph{S}-class recognition.}
  \label{fig:model_scale}
\end{figure}

We assess backbone scale by training three DeepArrhythmia variants with rich-evidence that differ only in the Qwen3.5 backbone (0.8B, 2B, and 4B), while holding the remaining architecture and training settings fixed. All models are evaluated on the same held-out MIT-BIH split using beat-level Micro-F1, Macro-F1, and per-class F1.

Figure~\ref{fig:model_scale} indicates a clear positive scaling trend with increasing backbone size. At the aggregate level, beat-level Micro-F1 improves from 0.937 (0.8B) to 0.950 (4B), while Macro-F1 increases from 0.555 to 0.593. At the class level, the gains for N and VEB are relatively modest, rising from 0.966 to 0.974 and from 0.936 to 0.940, respectively, which is expected given that both classes already operate near ceiling performance. In contrast, SVEB benefits most substantially from increased model scale, with F1 improving from 0.245 to 0.392, suggesting that larger backbones primarily enhance recognition of the minority class. Fusion beats remain challenging across all model sizes, with F1 remaining close to zero throughout.

\section{Interpretation Analysis}
\label{app:Interpretation Analysis}
\begin{figure*}[t]
  \centering
  \begin{subfigure}{0.48\textwidth}
    \centering
    \includegraphics[width=\linewidth]{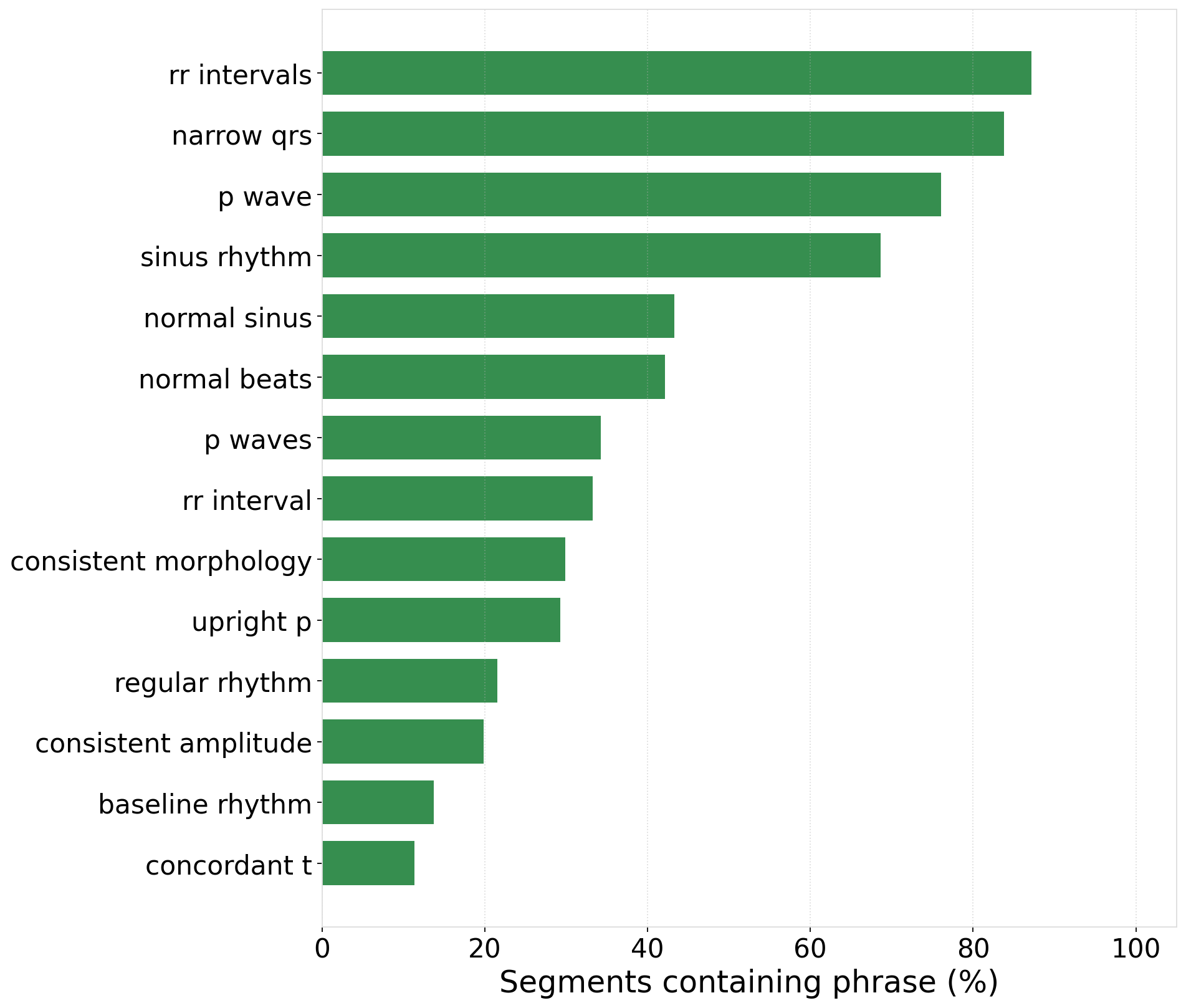}
    \caption{Bigram distribution for class N.}
    \label{subfig:n}
  \end{subfigure}
  \hfill
  \begin{subfigure}{0.48\textwidth}
    \centering
    \includegraphics[width=\linewidth]{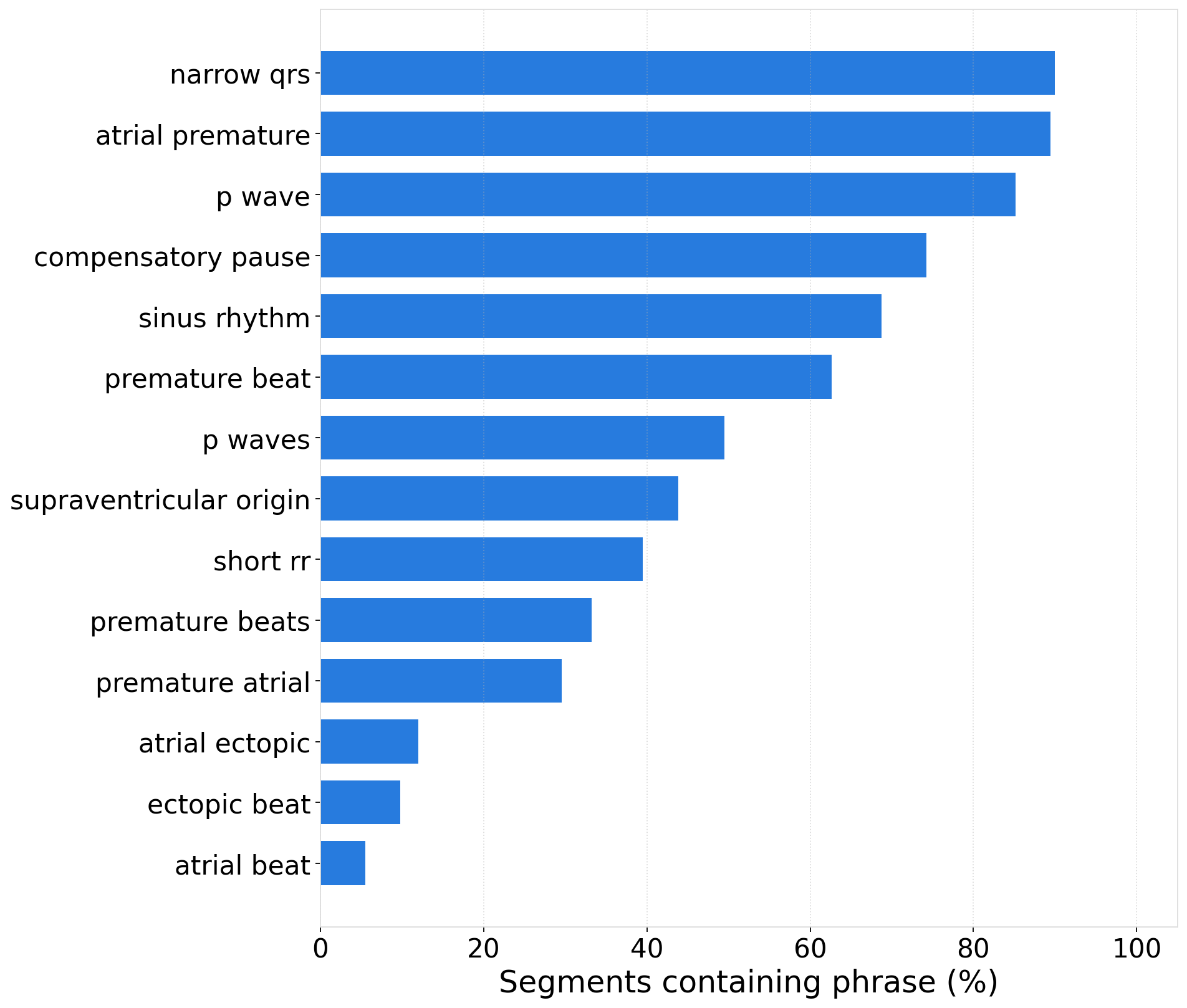}
    \caption{Bigram distribution for class S.}
    \label{subfig:s}
  \end{subfigure}

  \vspace{0.5em}

  \begin{subfigure}{0.48\textwidth}
    \centering
    \includegraphics[width=\linewidth]{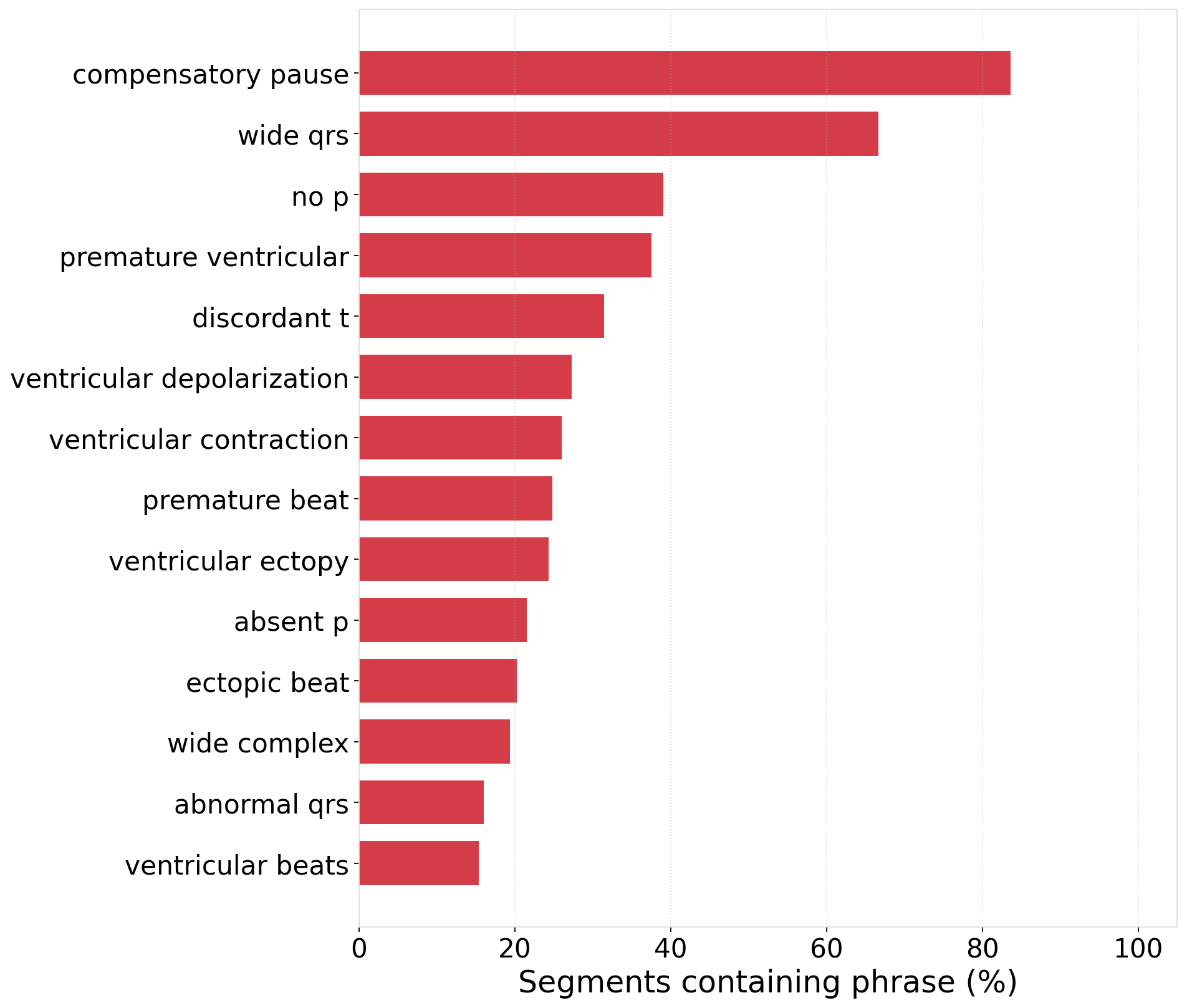}
    \caption{Bigram distribution for class V.}
    \label{subfig:v}
  \end{subfigure}
  \hfill
  \begin{subfigure}{0.48\textwidth}
    \centering
    \includegraphics[width=\linewidth]{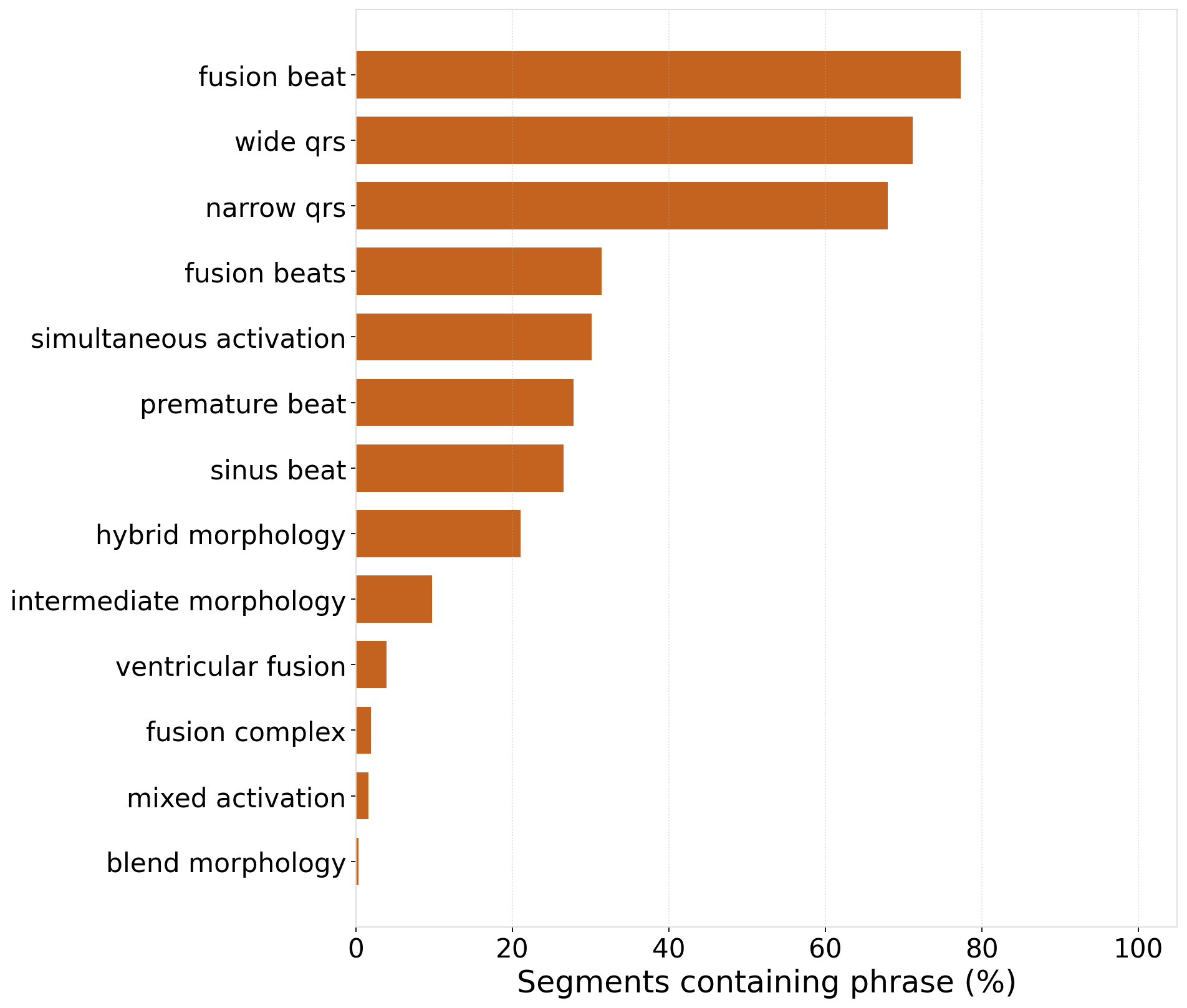}
    \caption{Bigram distribution for class F.}
    \label{subfig:f}
  \end{subfigure}

  \caption{Class-wise bigram statistics extracted from interpretation captions across beat classes N, S, V, and F.}
  \label{fig:interpretation-bigram-2x2}
\end{figure*}

DeepArrhythmia provides an additional advantage beyond classification accuracy: improved interpretability. The morphology analyzer generates clinically grounded textual rationales that align model predictions with recognizable ECG patterns, thereby supporting transparent decision-making. Figure~\ref{fig:interpretation-bigram-2x2} shows the top bigram distributions of generated interpretations for each class on the MIT-BIH dataset. Three recurring diagnostic cues emerge.

\textbf{QRS range}
QRS-width descriptors are among the most frequent bigrams across all four classes. In particular, ``narrow QRS'' is more prevalent in normal and supraventricular beats, whereas ``wide QRS'' appears more frequently in ventricular beats. For fusion beats, narrow- and wide-QRS descriptors appear at comparable rates, which is consistent with their mixed morphology. This pattern is clinically plausible: wide-complex tachycardia (QRS $\geq$ 120 ms) is commonly associated with ventricular-origin rhythms, including monomorphic ventricular tachycardia, polymorphic ventricular tachycardia, and ventricular fibrillation \citep{desai2023arrhythmias}.

\textbf{P wave}
P-wave-related bigrams are highly ranked for normal and supraventricular classes, but are less prominent for ventricular beats. This trend is also physiologically meaningful: ventricular depolarization can dominate surface ECG morphology and obscure atrial activity, making P-wave visibility less reliable in ventricular rhythms \citep{goldberger2023goldberger}. Because P-wave detection remains challenging in noisy clinical recordings \citep{marvsanova2019advanced}, multimodal visual-language reasoning may provide complementary support for identifying subtle atrial signatures.

\textbf{RR interval}
RR-interval regularity remains a central cue for rhythm discrimination. Normal beats are associated with regular intervals (e.g., Figure~\ref{subfig:n}), while premature atrial and ventricular events are more often linked to irregular interval patterns and compensatory pauses. Importantly, RR-interval evidence is inherently contextual and cannot be inferred reliably from isolated beats. This observation supports segment-level modeling, where temporal context is explicitly incorporated into classification.

\section{Interpretation Examples}
\label{app:interpretation_examples}
\begin{figure}[t]
  \centering
  \includegraphics[width=0.8\linewidth]{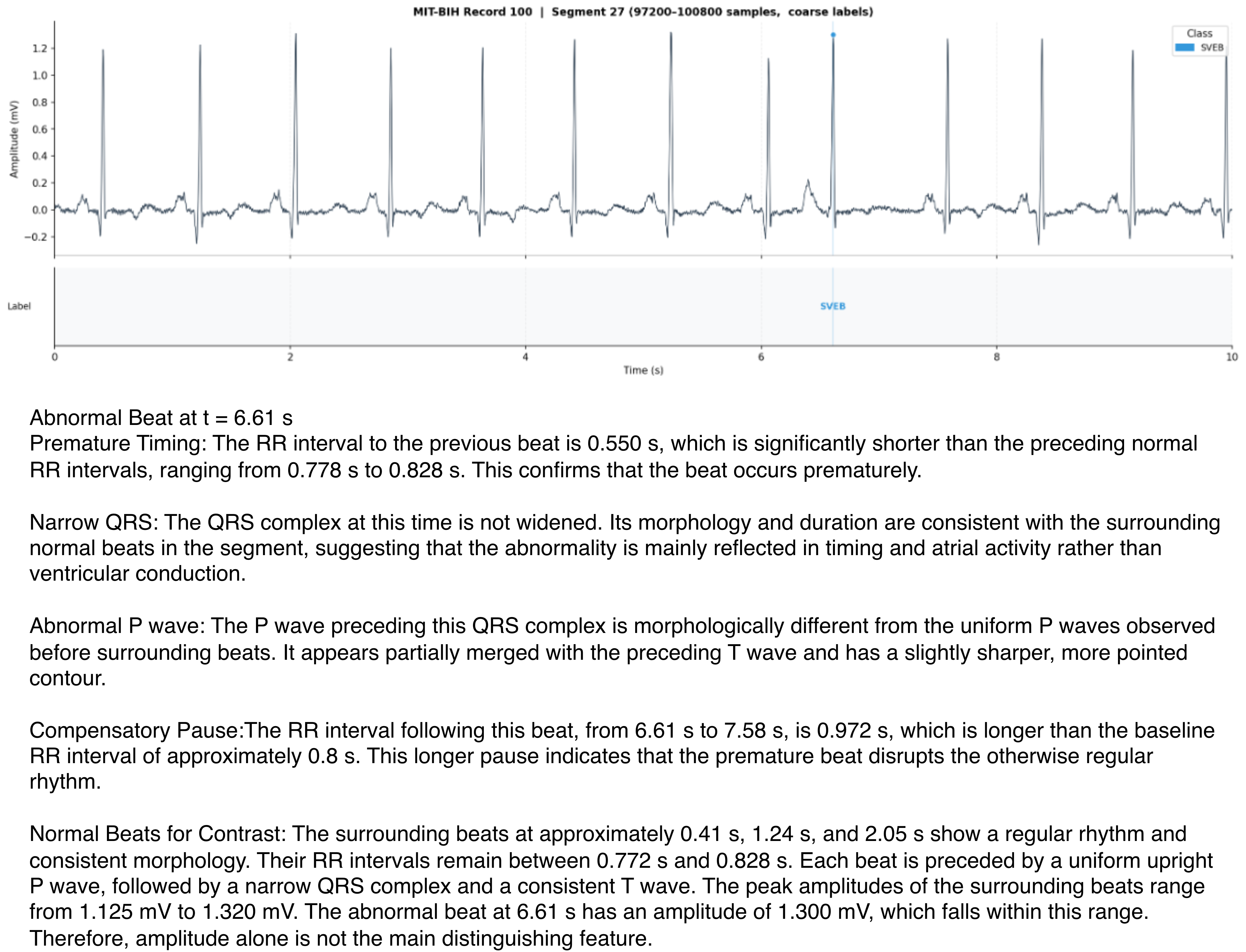}
  \caption{Interpretation example from MIT-BIH Record 100 (segment 27). Morphology analyzer identifies a supraventricular beat using premature timing, narrow QRS morphology, and a compensatory pause relative to surrounding normal beats.}
  \label{fig:ecg-100-seg27-coarse}
\end{figure}

We present representative ECG segments with analyses generated by the morphology analyzer to analyze the interpretability of DeepArrhythmia predictions. In Figure \ref{fig:ecg-100-seg27-coarse}, the model identifies a supraventricular beat using clinically relevant evidence, including premature timing, a narrow QRS complex, an abnormal or partially merged P wave, and a compensatory pause. The explanation further contrasts the target beat with neighboring normal beats through RR-interval and P-wave comparisons, providing transparent justification for the assigned class.

\begin{figure}[t]
  \centering
  \includegraphics[width=0.8\linewidth]{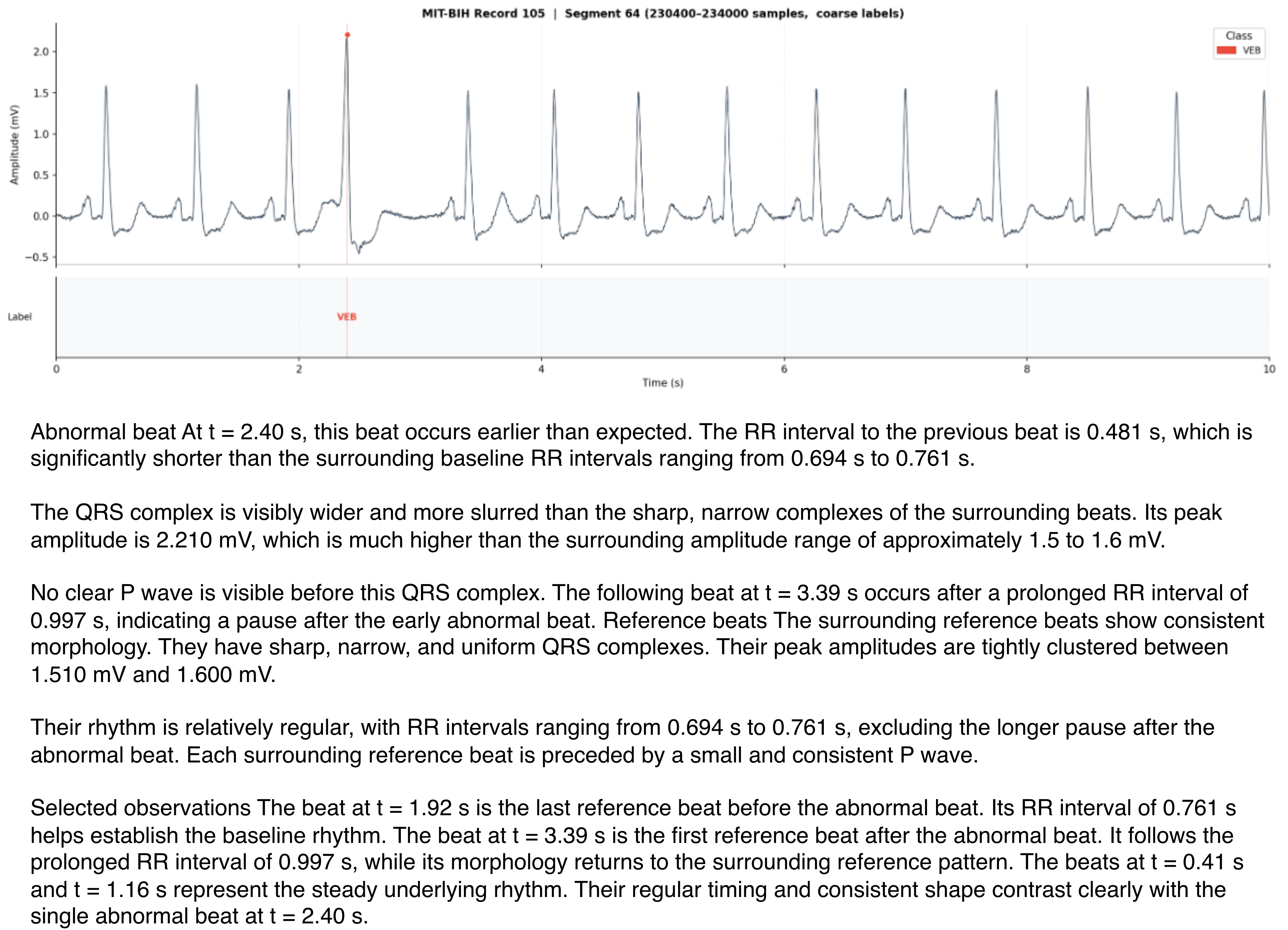}
  \caption{Second interpretation example from MIT-BIH Record 105 (segment 64). The model highlights altered P-wave morphology and elevated amplitude as evidence for its prediction.}
  \label{fig:ecg-105-seg64-coarse}
\end{figure}

Figure \ref{fig:ecg-105-seg64-coarse} provides a second positive example with a similar reasoning pattern. In this case, the model emphasizes altered atrial activity (e.g., absent or poorly resolved P-wave morphology) together with elevated peak amplitude relative to nearby normal beats, supporting its class decision with both morphological and quantitative cues.

\begin{figure}[t]
  \centering
  \includegraphics[width=0.8\linewidth]{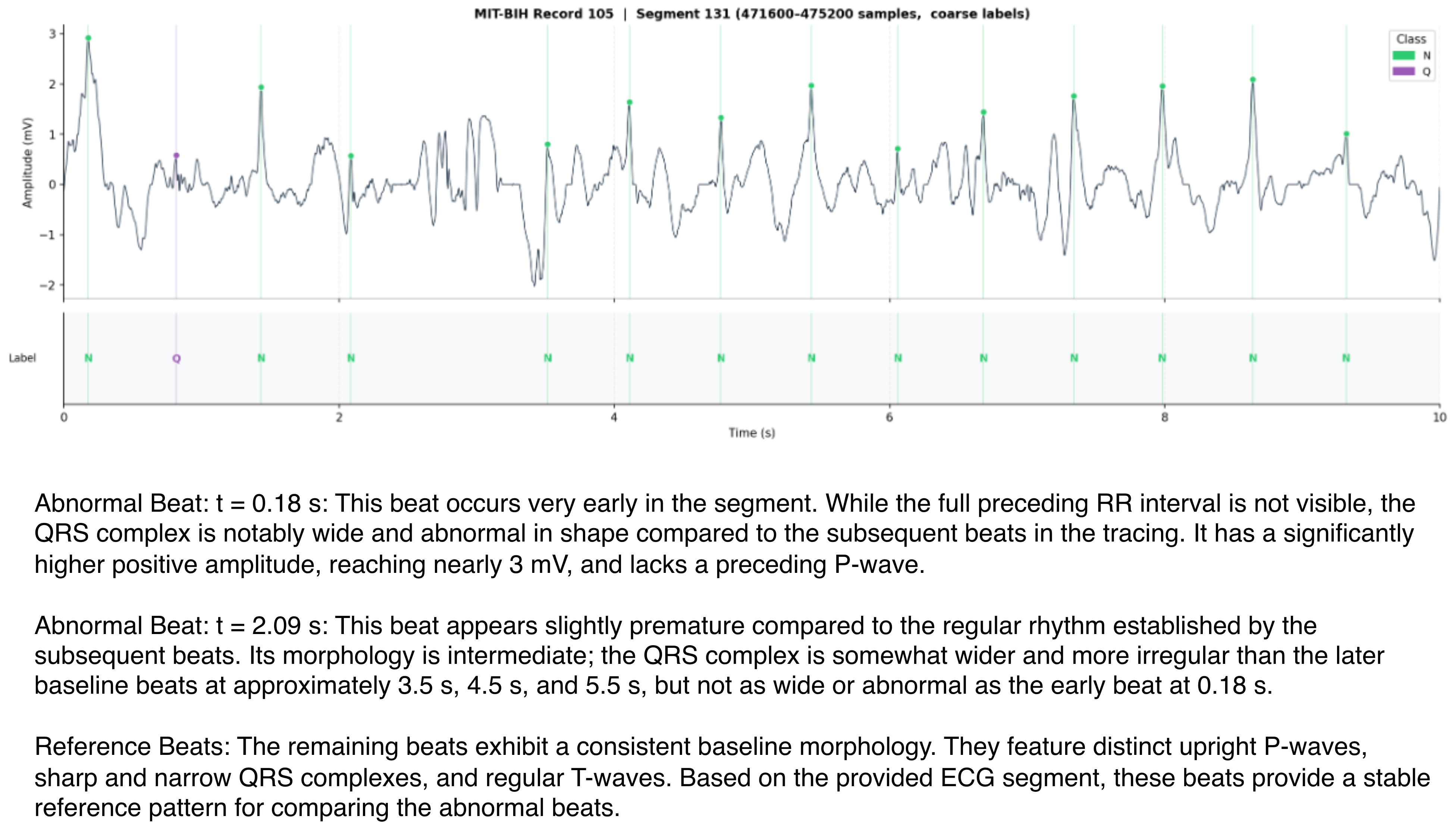}
  \caption{Failure-case example from MIT-BIH Record 105 (segment 131). Misclassifications are associated with incomplete temporal context for an early beat and morphology distortion under localized noise.}
  \label{fig:ecg-105-471600-475200}
\end{figure}

Figure \ref{fig:ecg-105-471600-475200} illustrates a failure case that clarifies current limitations. The first error occurs for an early beat where incomplete contextual information (e.g., limited visibility of the preceding RR interval and P-wave morphology) leads to a ventricular prediction based on partial evidence. The second error occurs in a noisy region, where signal corruption distorts morphology and RR-interval estimates, resulting in an incorrect classification. These examples indicate that performance remains sensitive to context truncation and local noise artifacts.

\section{Effect of the Teacher Model in the Morphology Analyzer}
\label{app:teacher_model_effect}
\begin{figure}[t]
  \centering
  \includegraphics[width=0.8\linewidth]{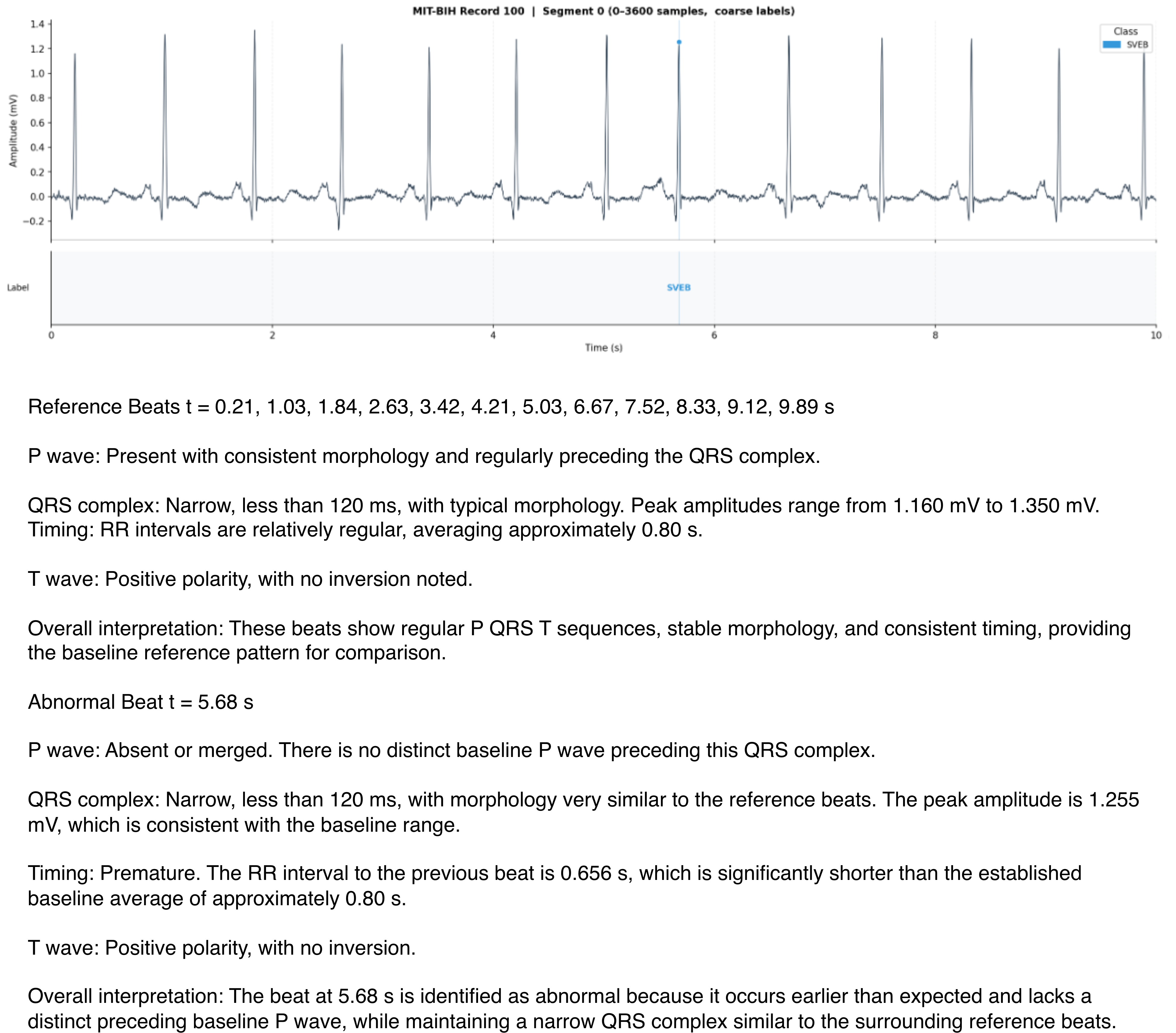}
  \caption{Illustrative interpretation produced by the morphology-analyzer student distilled from Gemma 4 on MIT-BIH Record 100 (segment 0).}
  \label{fig:gemma4_teacher_example}
\end{figure}

Because the current teacher used for morphology-rationale distillation is Gemini 3.1~\citep{team2023gemini}, which is closed-source, we additionally compare it with an open-source alternative, Gemma 4-31B~\citep{team2024gemma}. This ablation isolates teacher choice while keeping the student architecture and downstream classification pipeline fixed. We report the Macro-F1 and Micro-F1 performance of the rich evidence variant of DeepArrhythmia on the MIT-BIH dataset in Table~\ref{tab:teacher_model_effect}. The student distilled from Gemini 3.1 achieves higher Macro-F1 and Micro-F1 than the student distilled from Gemma 4 (0.593 vs.~0.572 and 0.950 vs.~0.944, respectively), indicating that teacher quality has a measurable effect on the usefulness of the learned morphology evidence. Figure~\ref{fig:gemma4_teacher_example} provides an illustrative output from the Gemma-4-distilled student on MIT-BIH Record 100; the rationale is relatively ambiguous and less clinically specific, for example describing the P wave only as absent or merged.

\begin{table}[t]
\centering
\caption{Effect of teacher choice on downstream beat-classification performance of the morphology-analyzer student. Higher values are better.}
\label{tab:teacher_model_effect}
\small
\begin{tabular}{lcc}
\toprule
Teacher model & Macro-F1 & Micro-F1 \\
\midrule
Gemini 3.1 & 0.593 & 0.950 \\
Gemma 4 & 0.572 & 0.944 \\
\bottomrule
\end{tabular}
\end{table}

\section{Broader Impacts and Limitations}
\label{app:broader_impacts_limitations}
DeepArrhythmia illustrates the potential of tool-grounded agentic systems to improve both predictive performance and interpretability in ECG analysis. By integrating peak detection, feature extraction, and morphology-aware reasoning, the framework produces evidence-linked predictions rather than purely opaque outputs. This design may support clinical workflows by enabling faster beat-level screening, prioritizing suspicious segments for expert review, and providing traceable rationales aligned with established electrophysiological cues, including RR-interval dynamics, QRS morphology, and P-wave characteristics. Such support may be particularly valuable in resource-constrained settings where specialist availability is limited.

Beyond near-term clinical assistance, our results highlight a broader methodological opportunity: using a VLM as a central agent to orchestrate specialized physiological domain models. Rather than replacing decades of progress in physiological signal processing, this paradigm operationalizes that historical advancement by integrating established detectors, feature extractors, and morphology analyzers into a unified reasoning loop. The resulting architecture is inherently extensible: new domain tools can be added as modules while the central agent maintains cross-modal coordination and decision synthesis. In this sense, DeepArrhythmia represents an expandable AI-centered framework that is strengthened by domain knowledge, with potential applicability beyond ECG to other physiological monitoring tasks. The peak detector, feature extractor, and morphology analyzer operate on fiducial points and their local context, so they transfer to other quasi-periodic signals (e.g., systolic peaks in PPG, J-peaks in BCG, pulse onsets in arterial blood pressure); only the ECG-CoCa encoder is modality-specific.

Several limitations, however, remain. First, performance is still heterogeneous across classes, and fusion beats remain difficult to detect because of extreme class imbalance and substantial morphological variability. Second, as shown in the interpretation examples, prediction quality degrades under incomplete temporal context and low signal quality, indicating sensitivity to context truncation and noise contamination. Third, although results are strong across four benchmark datasets, the present study is retrospective; therefore, prospective utility across institutions, devices, and patient populations is not yet established. Fourth, LLM-based components may reflect biases in training data and can generate plausible but imperfect explanations, so outputs should be used to assist---not replace---clinical judgment. Finally, real-world deployment requires additional safeguards, including uncertainty calibration, robust out-of-distribution detection, privacy-preserving data governance, and human-in-the-loop oversight to reduce automation bias.

Overall, DeepArrhythmia should be regarded as an assistive decision-support framework rather than an autonomous diagnostic system. It targets offline and retrospective analysis, such as Holter review and second reading of flagged segments, rather than hard real-time on-device alarming. We evaluate on beat-annotated datasets because the task requires per-beat AAMI labels; large resources such as PTB-XL and MIMIC-IV-ECG provide record-level 12-lead diagnostic statements and are therefore outside the beat-level scope of this paper. Future work should emphasize multicenter external validation, improved robustness for rare classes and noisy recordings (e.g., generative synthesis of realistic fusion morphology), label-free or conformal selection of the routing threshold, and prospective workflow-level evaluation of effects on diagnostic quality, patient safety, and clinician workload.

\section{Human Evaluation Protocol}
\label{app:human_eval_protocol}

We evaluate generated morphology-analysis outputs from three sources: Gemini 3.1 Pro with label conditioning, Gemini 3.1 Pro without label conditioning, and DeepArrhythmia. For each source, 200 samples are randomly selected, resulting in 600 total evaluated samples. All generated outputs are anonymized before review and presented in blinded, randomized order to three experts so that the evaluators assess the text itself rather than model identity. Each sample is graded independently according to the seven criteria summarized in Table~\ref{tab:expert_eval_revised}.

The evaluator instructions were as follows: for each anonymized generated output, read the morphology analysis and assign a score from 1 to 5 for each criterion in Table~\ref{tab:expert_eval_revised}; score each sample independently; and base the judgment on the clinical quality of the generated output, including whether it provides class-consistent diagnostic support, cites observable ECG evidence, remains physiologically plausible, uses segment context when needed, covers the key evidence, expresses uncertainty appropriately, and would be useful in expert review.

\begin{table*}[t]
\centering
\caption{Description of the seven criteria used in the expert evaluation of morphology analyses.}
\label{tab:expert_eval_revised}
\small
\setlength{\tabcolsep}{5pt}
\begin{tabularx}{\textwidth}{p{3.4cm} X}
\toprule
\textbf{Criterion} & \textbf{Description and scoring guidance} \\
\midrule

\textbf{Diagnostic Support} &
Does the explanation provide convincing evidence for the predicted arrhythmia label (\textit{N}, \textit{S}, \textit{V}, or \textit{F}) rather than merely restating the answer?
A score of 5 indicates that the explanation strongly supports the predicted class with class-consistent ECG evidence; a score of 1 indicates that the explanation does not support the prediction or supports the wrong class. \\

\textbf{ECG Evidence Grounding} &
Are the stated findings explicitly grounded in observable ECG evidence or tool-derived measurements, such as RR-interval pattern, QRS width/shape, P-wave morphology, compensatory pause, or amplitude differences?
A score of 5 requires specific, verifiable evidence; a score of 1 indicates vague, generic, or unsupported statements. \\

\textbf{Physiological Correctness} &
Are the described ECG findings medically plausible and internally consistent?
A score of 5 indicates no clear physiological errors; a score of 1 indicates major factual errors, contradictions, or clinically implausible reasoning. \\

\textbf{Use of Segment Context} &
Does the explanation appropriately use surrounding beats and temporal context when needed, especially for rhythm-related discrimination?
A score of 5 indicates effective use of neighboring beats and context; a score of 1 indicates isolated-beat reasoning when context is essential. \\

\textbf{Evidence Completeness} &
Does the explanation cover the main evidence needed for this case, including both morphology and rhythm cues when relevant?
A score of 5 indicates that the important cues are covered with no major omissions; a score of 1 indicates that most key evidence is missing. \\

\textbf{Uncertainty and Safety} &
Does the output avoid overclaiming in noisy, ambiguous, or low-context cases, and does it appropriately acknowledge limited evidence?
A score of 5 indicates well-calibrated and safe phrasing; a score of 1 indicates unjustified certainty or unsafe overinterpretation. \\

\textbf{Clinical Review Utility} &
Would this output help an expert review the case more efficiently, verify the prediction, or understand possible failure modes?
A score of 5 indicates strong practical usefulness for expert review; a score of 1 indicates little or no value for clinical assessment. \\

\bottomrule
\end{tabularx}
\end{table*}

\end{document}